\documentclass[10pt,twocolumn,letterpaper]{article}

\usepackage{cvpr}              
\usepackage{times}
\usepackage{epsfig}
\usepackage{graphicx}
\usepackage{caption}
\usepackage[super]{nth}
\usepackage{amsmath}
\usepackage{amssymb}
\usepackage{balance}

\usepackage{pifont}
\usepackage{url}
\usepackage{multirow} 
\usepackage{float} 
\usepackage{enumitem}
\usepackage{paralist}
\usepackage{booktabs}
\usepackage[T1]{fontenc}
\usepackage[table]{xcolor} 
\definecolor{deeppink}{rgb}{1.0, 0.08, 0.58}
\definecolor{blue_new}{rgb}{0.06, 0.75, 0.99}
\usepackage{array}

\usepackage[pagebackref=true,breaklinks=true,letterpaper=true,colorlinks,citecolor=cvprblue, linkcolor=cvprblue,urlcolor=deeppink,bookmarks=false]{hyperref}

\definecolor{cvprblue}{rgb}{0.21,0.49,0.74}

\definecolor{mygray}{gray}{0.4}
\newenvironment{mytiny}{\color{mygray}\small}{}

\title{Text-Driven Image Editing via Learnable Regions}

\author{
  Yuanze Lin$^\spadesuit$ ~~  Yi-Wen Chen$^\clubsuit$ ~~ Yi-Hsuan Tsai$^\bigstar$ ~~ Lu Jiang$^\bigstar$ ~~ Ming-Hsuan Yang$^{\clubsuit\bigstar}$ \vspace{.3em}\\ 
  $^\spadesuit$ University of Oxford ~~ $^\clubsuit$ UC Merced ~~
    $^\bigstar$ Google \\
}

\begin{document}

\twocolumn[{%
\renewcommand\twocolumn[1][]{#1}%
\maketitle

\begin{center}
\begin{minipage}{0.16\textwidth}
\captionof*{figure}{\small{\textcolor{black}{Input Image}}}
\vspace{-3mm}
\includegraphics[width=\linewidth]{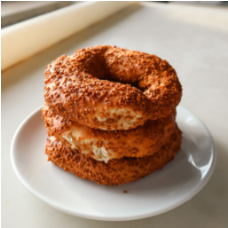}
\end{minipage}\hfill
\begin{minipage}{0.16\textwidth}
\captionof*{figure}{\small{\textcolor{black}{Edited Image}}}
\vspace{-3mm}
\includegraphics[width=\linewidth]{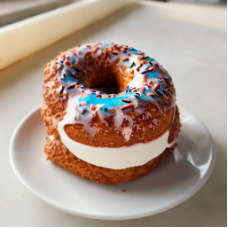}   
\end{minipage}\hfill
\begin{minipage}{0.16\textwidth}
\captionof*{figure}{\small{\textcolor{black}{Input Image}}}
\vspace{-3mm}
\includegraphics[width=\linewidth]{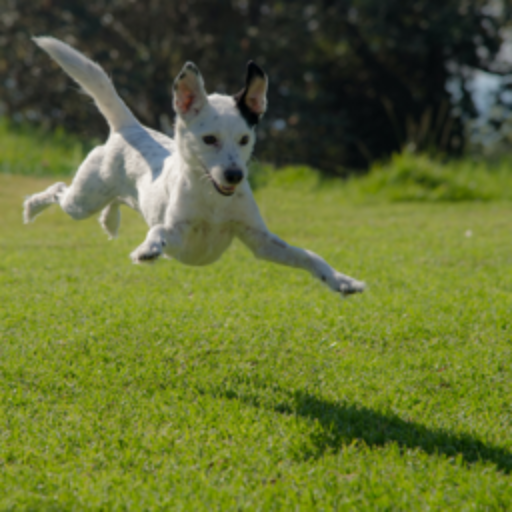} 
\end{minipage}\hfill
\begin{minipage}{0.16\textwidth}
\captionof*{figure}{\small{\textcolor{black}{Edited Image}}}
\vspace{-3mm}
\includegraphics[width=\linewidth]{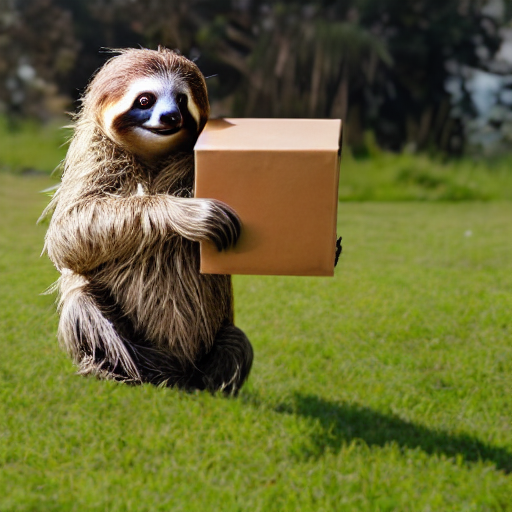} 
\end{minipage}\hfill
\begin{minipage}{0.16\textwidth}
\captionof*{figure}{\small{\textcolor{black}{Input Image}}}
\vspace{-3mm}
\includegraphics[width=\linewidth]{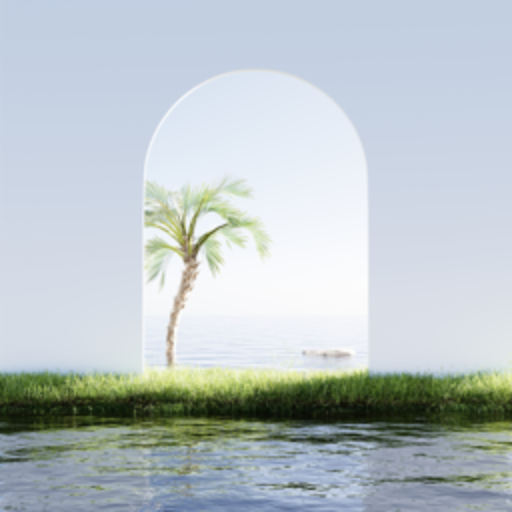} 
\end{minipage}\hfill
\begin{minipage}{0.16\textwidth}
\captionof*{figure}{\small{\textcolor{black}{Edited Image}}}
\vspace{-3mm}
\includegraphics[width=\linewidth]{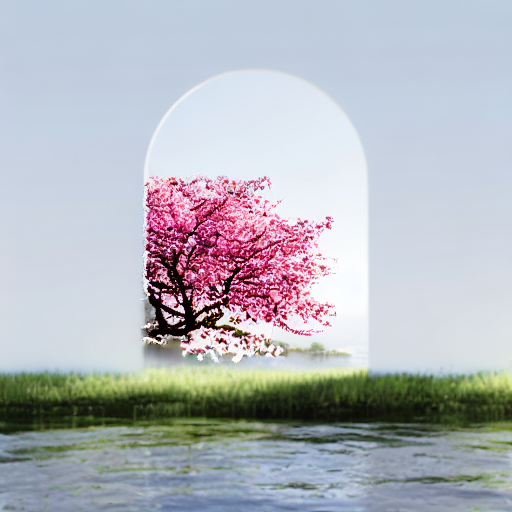} 
\end{minipage}
\\
\vspace{2mm}
\textcolor{black}{\hspace{1mm} \small{\textit{donuts with ice cream}}  \hspace{24mm}  \small{\textit{a cute sloth holds a box}}   \hspace{26mm}  \small{\textit{a flowering cherry tree}}}
\vspace{3mm}

\begin{minipage}{0.16\linewidth}
\vspace{-1mm}
\includegraphics[width=\linewidth]{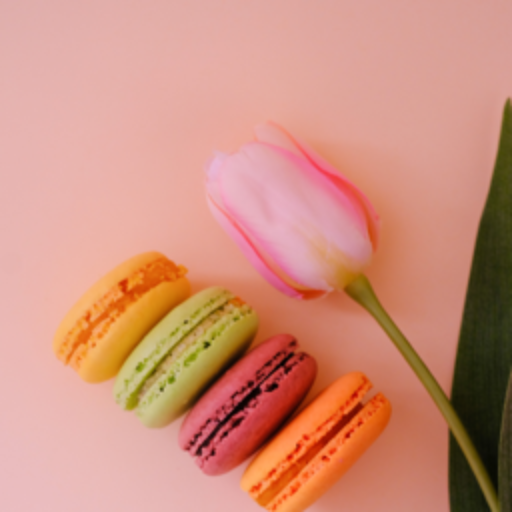}
\end{minipage}\hfill
\begin{minipage}{0.16\textwidth}
\vspace{-1mm}
\includegraphics[width=\linewidth]{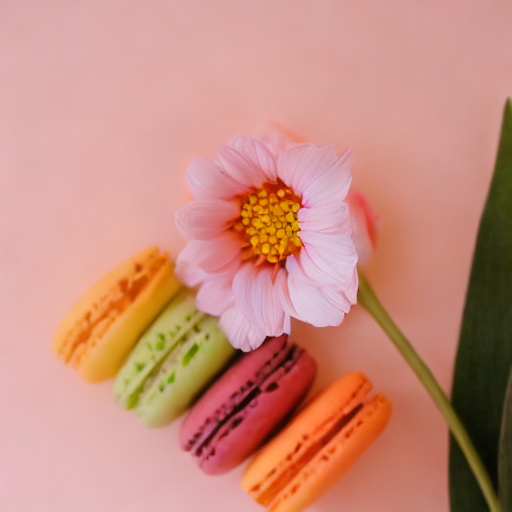}   
\end{minipage}\hfill
\begin{minipage}{0.16\textwidth}
\vspace{-1mm}
\includegraphics[width=\linewidth]{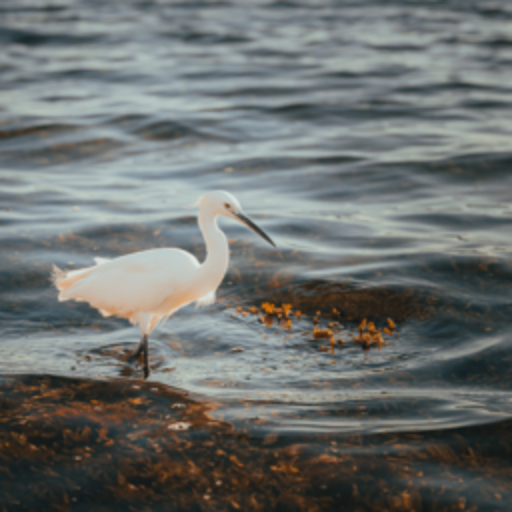} 
\end{minipage}\hfill
\begin{minipage}{0.16\textwidth}
\vspace{-1mm}
\includegraphics[width=\linewidth]{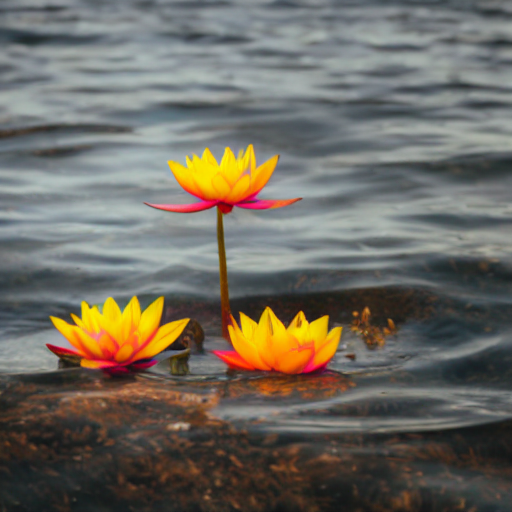} 
\end{minipage}\hfill
\begin{minipage}{0.16\textwidth}
\vspace{-1mm}
\includegraphics[width=\linewidth]{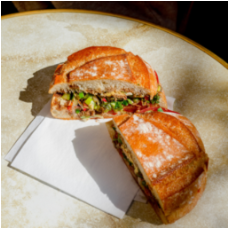} 
\end{minipage}\hfill
\begin{minipage}{0.16\textwidth}
\vspace{-1mm}
\includegraphics[width=\linewidth]{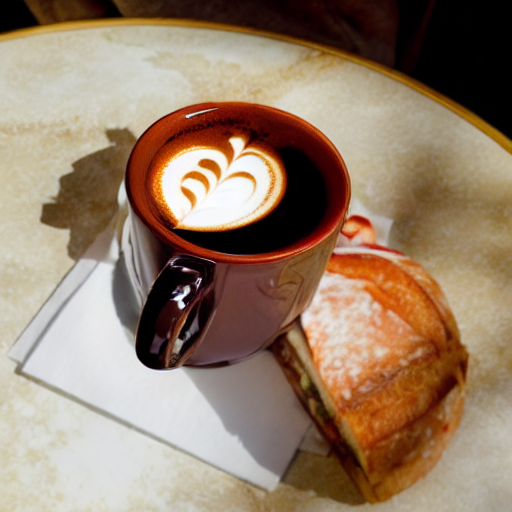} 
\end{minipage}
\\
\vspace{3mm}
\textcolor{black}{\textit{a blooming flower and dessert}   \hspace{5mm} \textit{several lotus flowers growing in the water} \hspace{5mm}  \textit{a cup of coffee next to the bread}}
\captionof{figure}{\textbf{Overview.} Given an input image and a language description for editing, our method can generate realistic and relevant images without the need for user-specified regions for editing. It performs local image editing while preserving the image context.}
\label{figure1}
\end{center}
\vspace{3mm}
}]

\begin{abstract}
\vspace{-2mm}
Language has emerged as a natural interface for image editing. In this paper, we introduce a method for region-based image editing driven by textual prompts, without the need for user-provided masks or sketches. Specifically, our approach leverages an existing pre-trained text-to-image model and introduces a bounding box generator to identify the editing regions that are aligned with the textual prompts. We show that this simple approach enables flexible editing that is compatible with current image generation models, and is able to handle complex prompts featuring multiple objects, complex sentences, or lengthy paragraphs. We conduct an extensive user study to compare our method against state-of-the-art methods. The experiments demonstrate the competitive performance of our method in manipulating images with high fidelity and realism that correspond to the provided language descriptions. Our project webpage  can be found at: \url{https://yuanze-lin.me/LearnableRegions_page}.
\end{abstract}    
\section{Introduction}
\label{sec:intro}

\begin{figure*}[htp]
\begin{minipage}{0.139\linewidth}
\captionof*{figure}{\small{\textcolor{black}{Input Image}}}
\vspace{-3mm}
\includegraphics[width=\linewidth]{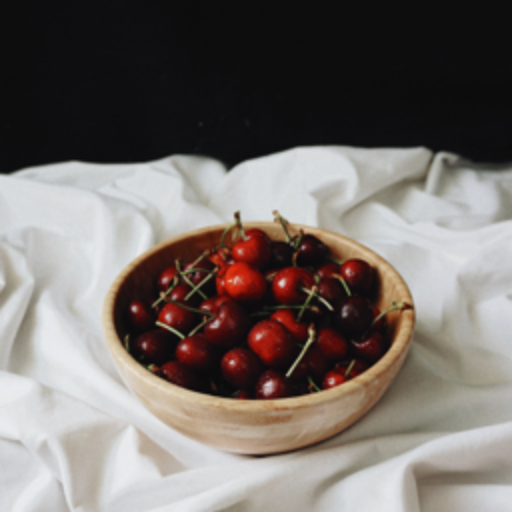}
\end{minipage}\hfill
\begin{minipage}{0.139\textwidth}
\captionof*{figure}{\small{\textcolor{black}{Region 1}}}
\vspace{-3mm}
\includegraphics[width=\linewidth]{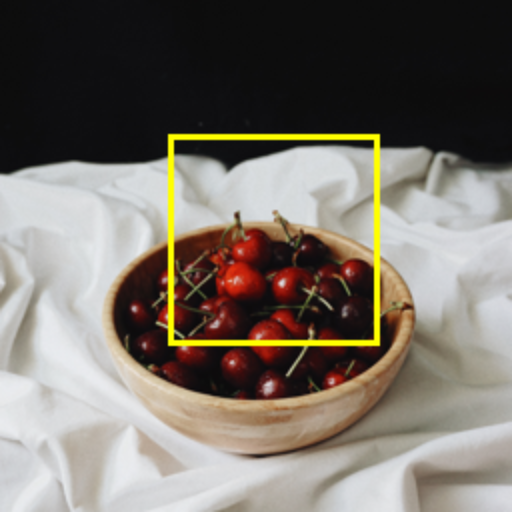}   
\end{minipage}\hfill
\begin{minipage}{0.139\textwidth}
\captionof*{figure}{\small{\textcolor{black}{Output 1}}}
\vspace{-3mm}
\includegraphics[width=\linewidth]{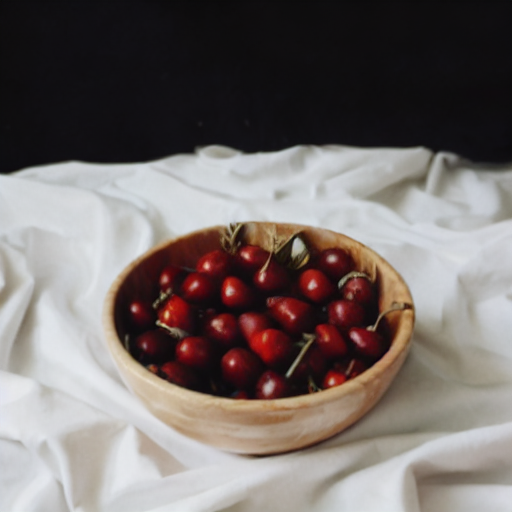} 
\end{minipage}\hfill
\begin{minipage}{0.139\textwidth}
\captionof*{figure}{\small{\textcolor{black}{Region 2}}}
\vspace{-3mm}
\includegraphics[width=\linewidth]{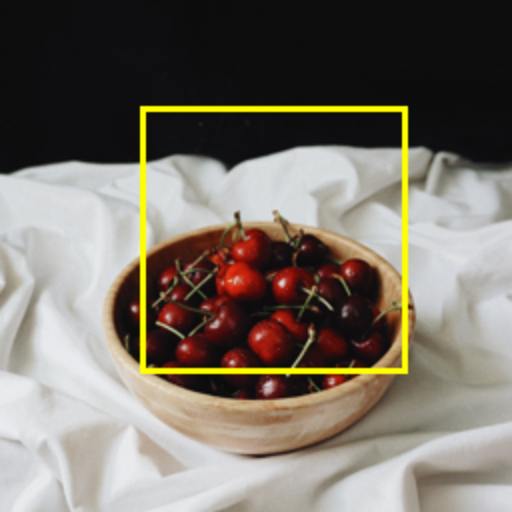} 
\end{minipage}\hfill
\begin{minipage}{0.139\textwidth}
\captionof*{figure}{\small{\textcolor{black}{Output 2}}}
\vspace{-3mm}
\includegraphics[width=\linewidth]{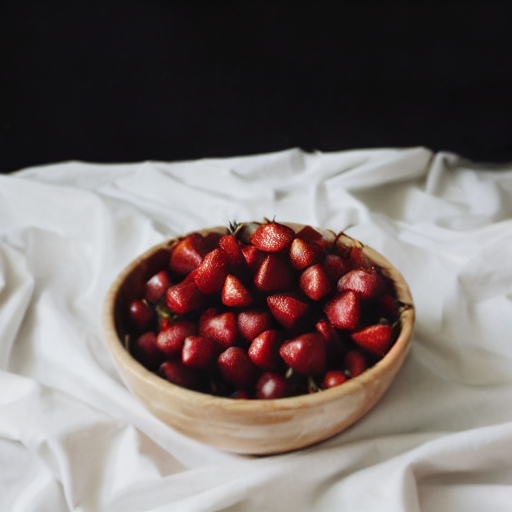} 
\end{minipage}\hfill
\begin{minipage}{0.139\textwidth}
\captionof*{figure}{\small{\textcolor{black}{Region (Ours)}}}
\vspace{-3mm}
\includegraphics[width=\linewidth]{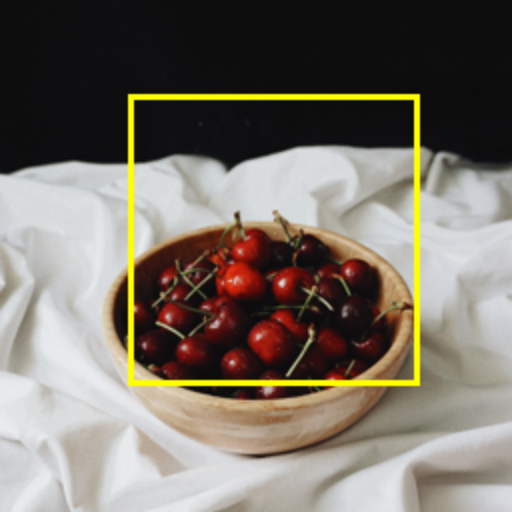} 
\end{minipage}\hfill
\begin{minipage}{0.139\textwidth}
\captionof*{figure}{\small{\textcolor{black}{Output (Ours)}}}
\vspace{-3mm}
\includegraphics[width=\linewidth]{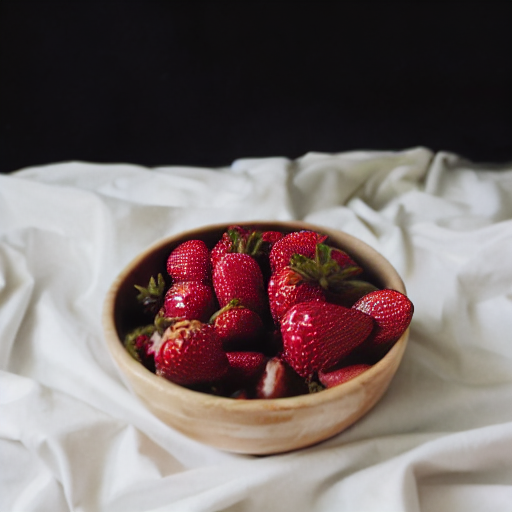} 
\end{minipage}
\vspace{1mm}
\\
\hspace{-2mm} \small{\textcolor{black}{\textsl{Editing Text:}}}   \hspace{54mm} \textcolor{black}{\textit{a bowl of strawberries}}
\vspace{1mm}
\\
\begin{minipage}{0.139\linewidth}
\includegraphics[width=\linewidth]{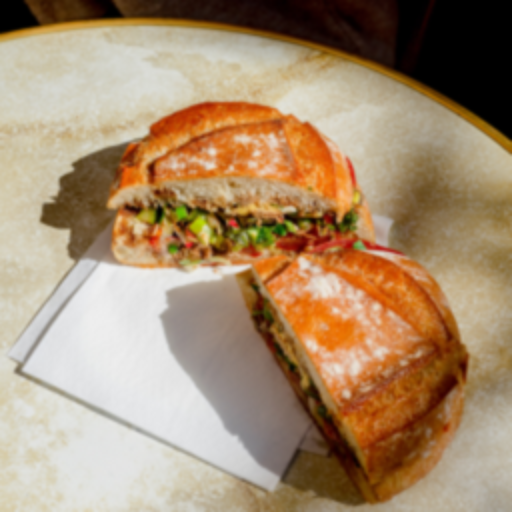}
\end{minipage}\hfill
\begin{minipage}{0.139\textwidth}
\includegraphics[width=\linewidth]{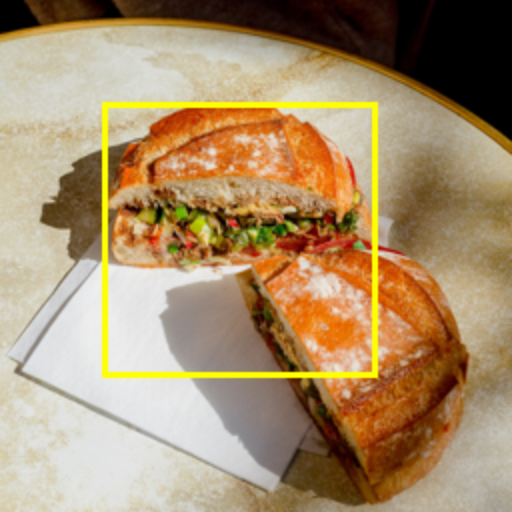}   
\end{minipage}\hfill
\begin{minipage}{0.139\textwidth}
\includegraphics[width=\linewidth]{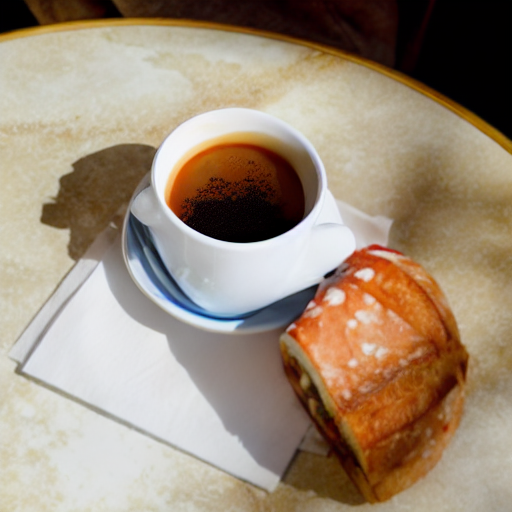} 
\end{minipage}\hfill
\begin{minipage}{0.139\textwidth}
\includegraphics[width=\linewidth]{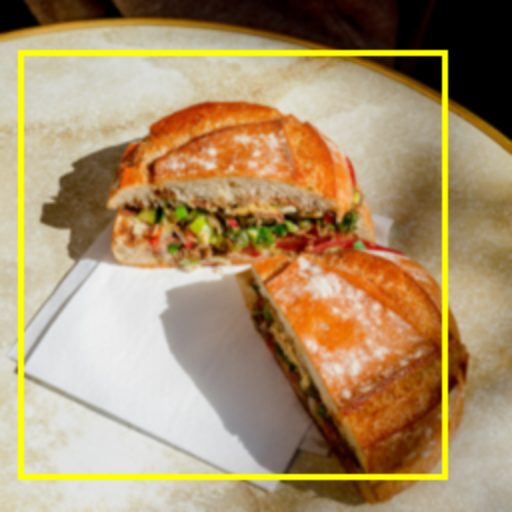} 
\end{minipage}\hfill
\begin{minipage}{0.139\textwidth}
\includegraphics[width=\linewidth]{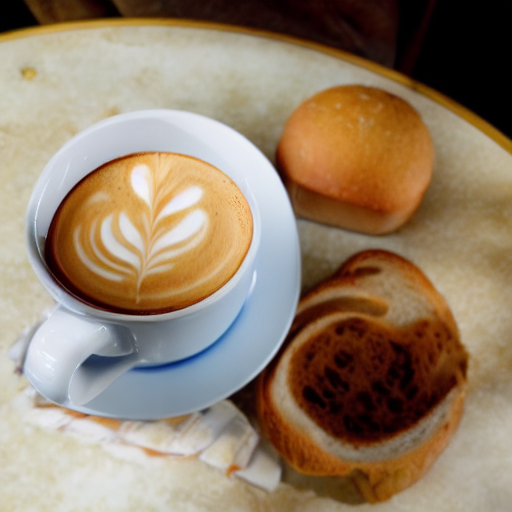} 
\end{minipage}\hfill
\begin{minipage}{0.139\textwidth}
\includegraphics[width=\linewidth]{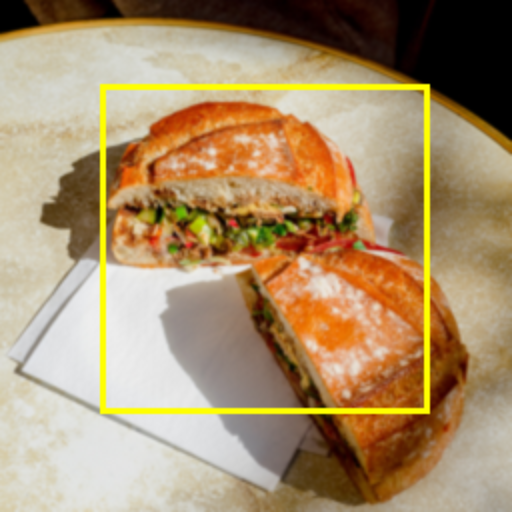} 
\end{minipage}\hfill
\begin{minipage}{0.139\textwidth}
\includegraphics[width=\linewidth]{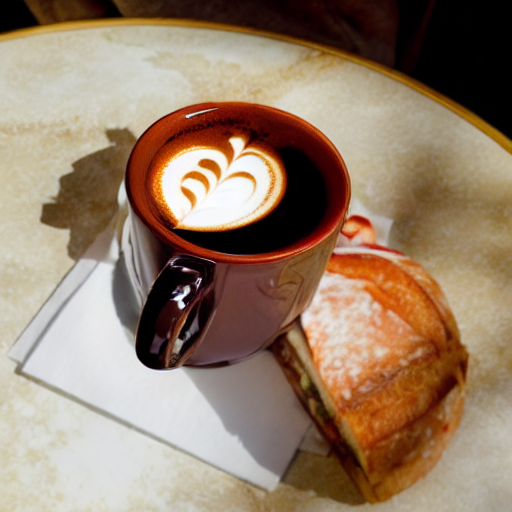} 
\end{minipage}
\vspace{1mm}
\\
\hspace{-2mm} \small{\textcolor{black}{\textsl{Editing Text:}}}   \hspace{47mm} \textcolor{black}{\textit{a cup of coffee next to the bread}}
\vspace{-1mm}

\captionof{figure}{\textbf{Effects of variations in editing regions on generated image quality.} \textit{Region 1} and \textit{Region 2} are two prior regions drawn from the self-attention map of DINO~\cite{caron2021emerging}. \textit{Region (ours)}, shown in the second-to-last column, represents the regions produced by our model which have the best overall quality.}
\label{fig:region_editing}
\end{figure*}

With the availability of a massive amount of text-image paired data and large-scale vision-language models, recent text-driven image synthesis models \cite{zhang2021text,lin2021self,lin2022revive,lin2023smaug, nichol2021glide, su2022dual, ramesh2021zero,radford2021learning,chang2022maskgit,ye2022improving} have enabled people to create and manipulate specific visual contents of realistic images using natural language descriptions in an interactive fashion.

Recent text-driven image editing methods \cite{nichol2021glide,kwon2022clipstyler,avrahami2022blended,bar2022text2live, tumanyan2023plug, hertz2023prompt, mokady2023null} have shown impressive capabilities in editing realistic images based on natural descriptions, with approaches typically falling into two paradigms: \emph{mask-based} or \emph{mask-free} methods. Mask-based editing approaches~\cite{nichol2021glide,avrahami2022blended} are perceived intuitively for local image editing because they allow users to specify precisely which areas of an image to modify. However, these methods can be laborious, as they demand users to manually create masks that are sometimes unnecessary, limiting their user experience in many applications.

In contrast, mask-free editing approaches~\cite{chang2023muse,hertz2023prompt,mokady2023null,tumanyan2023plug,cao2023masactrl} do not require masks and can directly modify the appearance or texture of the input image. These methods are trained to create fine-grained pixel masks, which can be applied to either the RGB space or the latent embedding space within the latent diffusion model framework. While there has been significant advancement in mask-free editing, the precision of editing in current methods relies heavily on the accuracy of detailed masks at the pixel level. Current methods encounter difficulties with local modifications, particularly when dealing with less accurate masks. 

Current mask-free image editing approaches have predominantly concentrated on pixel masks~\cite{couairon2022diffedit,cao2023masactrl}. The use of bounding boxes as an intermediate representation for editing images has not been thoroughly explored. Bounding boxes can provide an intuitive and user-friendly input for image editing. They facilitate a smoother interactive editing process by being quicker and easier for users to adjust the box, unlike pixel masks that typically require more time and precision to draw pixels accurately. Moreover, some generative transformer models such as Muse~\cite{chang2023muse} may support only box-like masks as opposed to pixel-level masking for image editing.

This paper explores the feasibility of employing bounding boxes as an intermediate representation within a mask-free editing framework. Our objective is not to propose a new image editing model, but to introduce a component that enables an existing pretrained mask-based editing model to perform mask-free editing via the learnable regions. To this end, we propose a region-based editing network that is trained to generate editing regions utilizing a text-driven editing loss with CLIP guidance~\cite{radford2021learning}. Our method can be integrated with different image editing models. To demonstrate its versatility, we apply it to two distinct image synthesis models: non-autoregressive transformers as used in MaskGIT~\cite{chang2022maskgit} and Muse~\cite{chang2023muse}, as well as Stable Diffusion \cite{rombach2022high}. It is worth highlighting that the latent spaces in transformer models (MaskGIT and Muse) are only compatible with box-like masks and lack the precision for pixel-level masks in image editing.

Our experimental results demonstrate that the proposed method can generate realistic images that match the context of the provided language descriptions. Furthermore, we conduct a user study to validate that our method outperforms five state-of-the-art baseline methods. The results indicate that our method edits images with greater fidelity and realism, following the changes specified in the language descriptions. The contributions of this work are as follows:
\begin{compactitem}
\item Our approach enables mask-based text-to-image models to perform local image editing without needing masks or other user-provided guidance. It can be integrated with existing text-guided editing models to improve their quality and relevance.
\item We introduce a novel region generator model that employs a new CLIP-guidance loss to learn to find regions for image editing. We demonstrate its applicability by integrating it with two popular and distinct text-guided editing models, MaskGIT \cite{chang2022maskgit} and Stable Diffusion \cite{rombach2022high}.
\item Experiments show the high quality and realism of our generated results. The user study further validates that our method outperforms state-of-the-art image editing baselines in producing favorable editing results.
\end{compactitem}

\begin{figure*}[htp]
\centering
\includegraphics[width=1\linewidth]{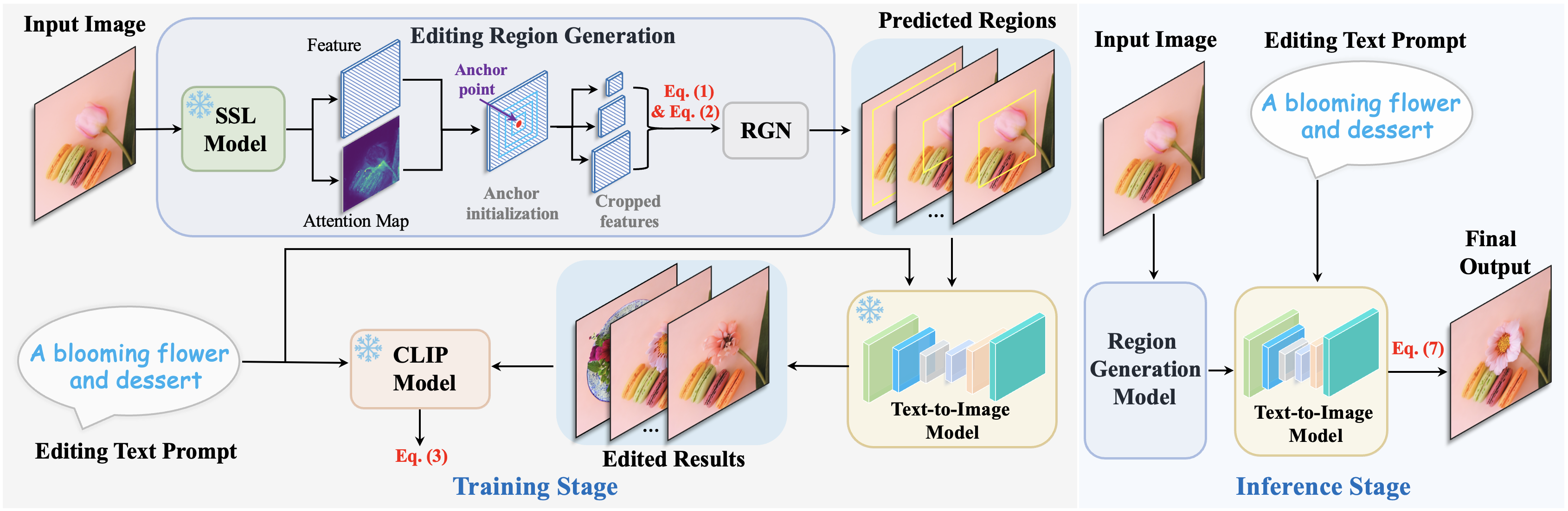}
\vspace{-5mm}
    \caption{\textbf{Framework of the proposed method.} We first feed the input image into the self-supervised learning (SSL) model, \eg, DINO~\cite{caron2021emerging}, to obtain the attention map and feature, which are used for anchor initialization. The region generation model initializes several region proposals (\eg, 3 proposals in this figure) around each anchor point, and learns to select the most suitable ones among them with the region generation network (RGN). The predicted region and the text descriptions are then fed into a pre-trained text-to-image model for image editing. We utilize the CLIP model for learning the score to measure the similarity between the given text description and the edited result, forming a training signal to learn our region generation model.
    }
\label{fig:architecture}
\end{figure*}
\section{Related Work}
\label{sec:related_work}

\paragraph{Text-to-Image Synthesis.}

In recent years, significant progress has been made in text-to-image synthesis. While early contributions are mainly based on Generative Adversarial Network (GAN) approaches \cite{zhang2017stackgan,zhang2018stackgan++,xu2018attngan,li2020manigan}, the latest models are mostly built on diffusion models~\cite{ho2020denoising,song2020denoising,nichol2021improved,rombach2022high,saharia2022imagen,Ramesh2022dalle2,gu2022vector,zhang2023adding,lin2024dreampolisher} or transformer models~\cite{Esser21vqgan,ramesh2021zero,ding2022cogview2,Yu2022parti}. For example, DALL·E 2 \cite{Ramesh2022dalle2} and Imagen \cite{ho2022imagen} propose to condition textual prompts to diffusion models, while Muse \cite{chang2023muse} leverages masked generative transformers to generate images from texts. Other approaches \cite{katherine1,crowson2022vqgan, Ramesh2022dalle2} leverage pre-trained CLIP models \cite{radford2021learning} to guide image generation based on textual descriptions.

More recently, Stable Diffusion~\cite{rombach2022high}, trained on large image-text pairs \cite{schuhmann2021laion}, has been made publicly available and has served as the foundation for numerous image generation and manipulation works. ControlNet~\cite{zhang2023adding} proposes to control Stable Diffusion with spatially localized conditions for image synthesis. Different from these works, we aim to introduce a component that can enable pre-trained text-to-image models for mask-free local image editing.

\vspace{-6mm}
\paragraph{Text-driven Image Manipulation.}

Several recent works have utilized pre-trained generator models and CLIP \cite{radford2021learning} for text-driven image manipulation \cite{bau2021paint,gal2022stylegan,liu2021fusedream,patashnik2021styleclip,kwon2022clipstyler,bar2022text2live}. StyleCLIP \cite{patashnik2021styleclip} combines the generative ability of StyleGAN \cite{karras2020analyzing} with CLIP to control latent codes, enabling a wide range of image manipulations. VQGAN-CLIP \cite{crowson2022vqgan} uses CLIP \cite{radford2021learning} to guide VQ-GAN \cite{Esser21vqgan} for high-quality image generation and editing.

There are several approaches \cite{nichol2021glide, su2022dual, avrahami2022blended, meng2021sdedit, kim2021diffusionclip, kawar2022imagic, gal2023textual, ruiz2022dreambooth} that use diffusion models for text-driven image manipulation. Imagic~\cite{kawar2022imagic} can generate textual embeddings aligning with the input images and editing prompts, and fine-tune the diffusion model to perform edits. InstructPix2Pix \cite{brooks2023instructpix2pix} combines GPT-3 \cite{brown2020language} and Stable Diffusion \cite{rombach2022high} to edit images with human instructions. Our work is related to the state-of-the-art methods DiffEdit \cite{couairon2022diffedit} and MasaCtrl \cite{cao2023masactrl}.
DiffEdit \cite{couairon2022diffedit} leverages DDIM inversion~\cite{dhariwal2021diffusion,song2020denoising} with the automatically produced masks for local image editing. MasaCtrl \cite{cao2023masactrl} proposes mutual self-attention and learns the editing masks from the cross-attention maps of the diffusion models. Motivated by the aforementioned works, we also utilize diffusion models and CLIP guidance for text-driven image manipulation. In contrast to DiffEdit \cite{couairon2022diffedit} and MasaCtrl \cite{cao2023masactrl}, whose editing is much more sensitive to the generated mask regions, our proposed method focuses on learning bounding boxes for local editing, which can be more flexible in accommodating diverse text prompts.
\section{Proposed Method}
Text-driven image editing manipulates the visual content of input images to align with the contexts or modifications specified in the text. 
Our goal is to enable text-to-image models to perform mask-free local image editing.
To this end, we propose a region generation network that can produce promising regions for image editing.

Figure~\ref{fig:architecture} shows the overall pipeline of our proposed method for text-driven image editing.

\subsection{Edit-Region Generation}
\label{generator}

Given the input image as $X\in \mathbb{R}^{3\times H \times W }$ and text with $p$ words as $T \in \mathbb{Z}^p$, we first use a pre-trained visual transformer model, ViT-B/16 \cite{dosovitskiy2020image}, for feature extraction. This model is pre-trained using the DINO self-supervised learning objective \cite{caron2021emerging}. The feature $F\in \mathbb{R}^{d \times h \times w}$ from the last layer has been shown to contain semantic segmentation of objects \cite{caron2021emerging,simeoni2021localizing}, which can serve as a prior in our problem.

Then we initialize $K$ anchor points ${\{C_{i}\}}_{i=1}^{K}$ located at the top-$K$ scoring patches of the self-attention map from the \texttt{[CLS]} token query of the DINO pre-trained transformer as shown in \cite{caron2021emerging}, where the \texttt{[CLS]} token carries guidance to locate the semantically informative parts of the objects.

Following this, we define a set of bounding box proposals ${\mathcal{B}_{i}=\{B_{j}\}}_{j=1}^{M}$ for each anchor point $C_{i}$, where each bounding box is centered at their corresponding anchor point. For simplicity, we parameterize the bounding box with a single parameter such that each $B_{j}$ is a square box with shape $j \times j$.

Subsequently, we train a region generation network to explicitly consider all unique bounding boxes derived from the same anchor point. For a given anchor point $C_{i}$, we then have: 
\begin{align}
    f_j & = \text{ROI-pool}(F, B_j), \label{roi} \\
   S([f_1, \cdots, f_M]) & = [\pi_1, \cdots, \pi_M], \label{select}
\end{align}
where $[\cdot]$ concatenates features along the channel dimension, and the \text{ROI-pool} operation~\cite{girshick2015fast} is used to perform pooling for the feature $F\in\mathbb{R}^{d \times h \times w}$ with respect to the box $B_{j}$, resulting in a feature tensor $f_{j}\in\mathbb{R}^{d \times l \times l}$. In our experiments, we set $l$ as 7. $S$ is the proposed region generation network consisting of two convolutional layers and two linear layers, with a ReLU activation layer between consecutive layers. The output from the final linear layer, denoted as $\pi_j$ in \cref{select} as the logits for the bounding box with size $j$, is fed into a softmax function to predict the scores for the bounding box proposal, \ie, $\text{Softmax}([\pi_1, \cdots, \pi_M])$. 

To learn the parameters of the region generation network, we use the Gumbel-Softmax trick~\cite{jang2016categorical}. 
We re-parameterize $\pi_j$ by adding a small Gumbel noise $g_j=-\log(-\log(u_j))$ where $u_j \sim \text{Uniform}(0,1)$. During training, we apply straight-through gradient estimation, in which backward propagation uses the differentiable variable (\ie, softmax) while the forward pass still takes the argmax, treating $\pi$ as the categorical variable. For each anchor point, once we obtain the editing region with the highest softmax score, we first generate a corresponding box-like mask image, and then feed the mask image, input image, and editing prompt into the text-to-image model to obtain the edited image. Thus, we can get $K$ edited images considering all anchor points, in Section \ref{inference}, we explain how to produce the final edited image as inference output.

\begin{figure*}[!th]
\begin{minipage}{0.16\textwidth}
\captionof*{figure}{\small{Input Image}}
\vspace{-3mm}
\includegraphics[width=\linewidth]{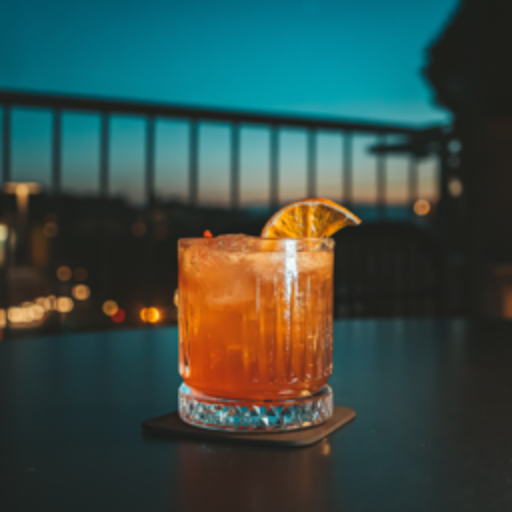}
\end{minipage}\hfill
\begin{minipage}{0.16\textwidth}
\captionof*{figure}{\small{\textcolor{black}{Edited Image}}}
\vspace{-3mm}
\includegraphics[width=\linewidth]{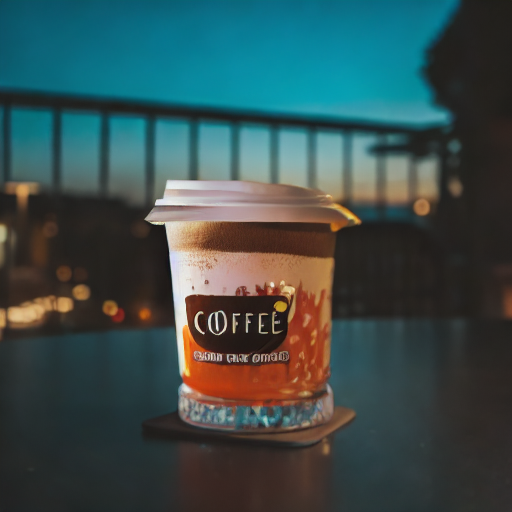}   
\end{minipage}\hfill
\begin{minipage}{0.16\textwidth}
\captionof*{figure}{\small{\textcolor{black}{Input Image}}}
\vspace{-3mm}
\includegraphics[width=\linewidth]{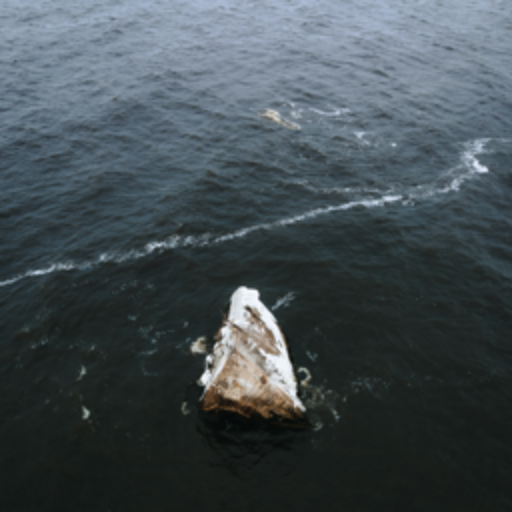} 
\end{minipage}\hfill
\begin{minipage}{0.16\textwidth}
\captionof*{figure}{\small{\textcolor{black}{Edited Image}}}
\vspace{-3mm}
\includegraphics[width=\linewidth]{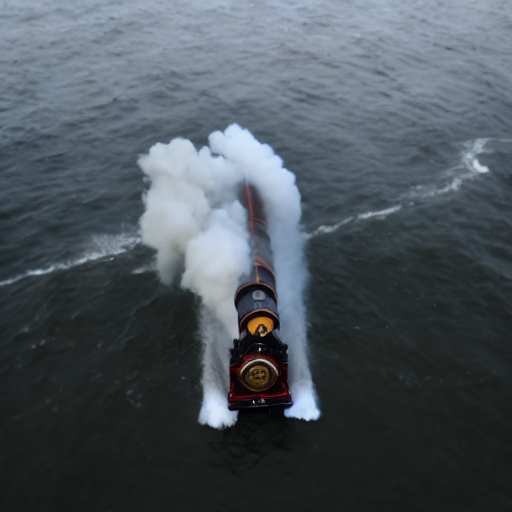} 
\end{minipage}\hfill
\begin{minipage}{0.16\textwidth}
\captionof*{figure}{\small{\textcolor{black}{Input Image}}}
\vspace{-3mm}
\includegraphics[width=\linewidth]{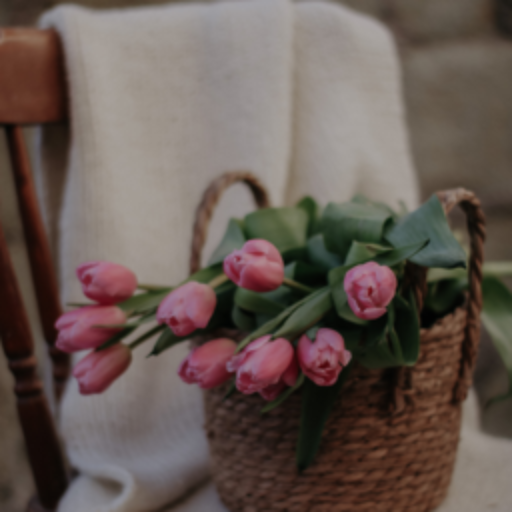} 
\end{minipage}\hfill
\begin{minipage}{0.16\textwidth}
\captionof*{figure}{\small{\textcolor{black}{Edited Image}}}
\vspace{-3mm}
\includegraphics[width=\linewidth]{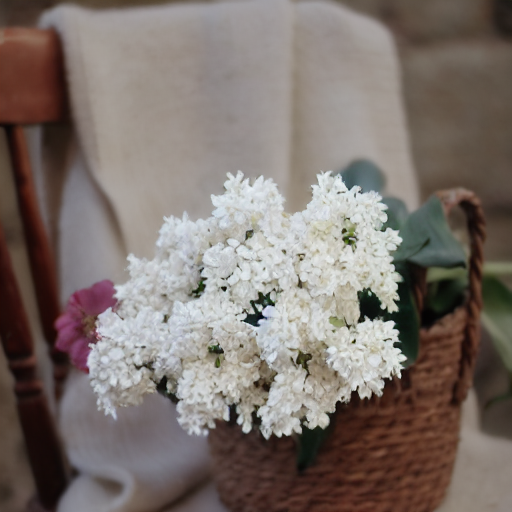} 
\end{minipage}
\vspace{2mm}
\\
\textcolor{black}{\textit{a cup whose logo is named as ``coffee"} \hspace{6mm} \textit{a steam train running on the sea} \hspace{2mm} \textit{many blooming jasmine flowers in the blanket}}
\vspace{5mm}
\begin{minipage}{0.16\textwidth}
\vspace{2mm}
\includegraphics[width=\linewidth]{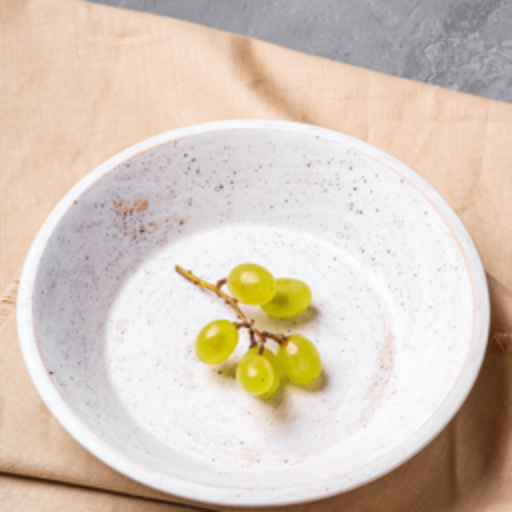}
\end{minipage}\hfill
\begin{minipage}{0.16\textwidth}
\vspace{2mm}
\includegraphics[width=\linewidth]{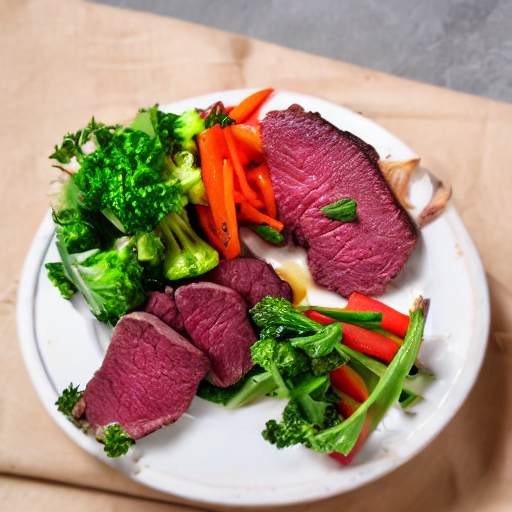}   
\end{minipage}\hfill
\begin{minipage}{0.16\textwidth}
\vspace{2mm}
\includegraphics[width=\linewidth]{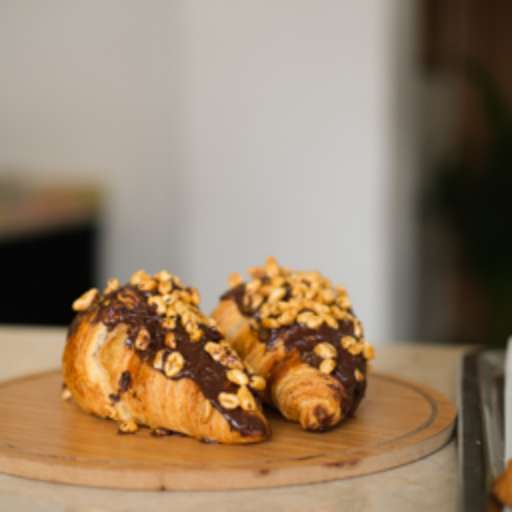} 
\end{minipage}\hfill
\begin{minipage}{0.16\textwidth}
\vspace{2mm}
\includegraphics[width=\linewidth]{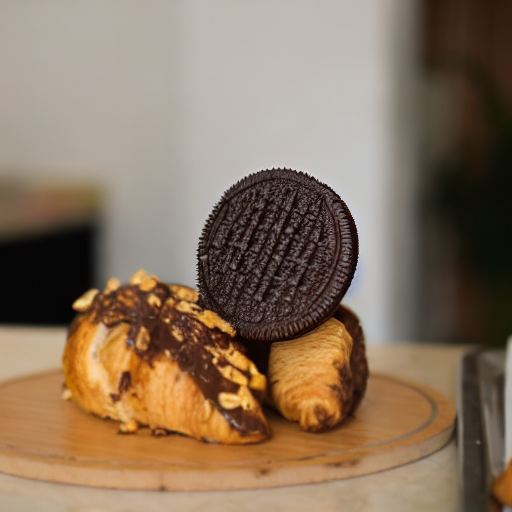} 
\end{minipage}\hfill
\begin{minipage}{0.16\textwidth}
\vspace{2mm}
\includegraphics[width=\linewidth]{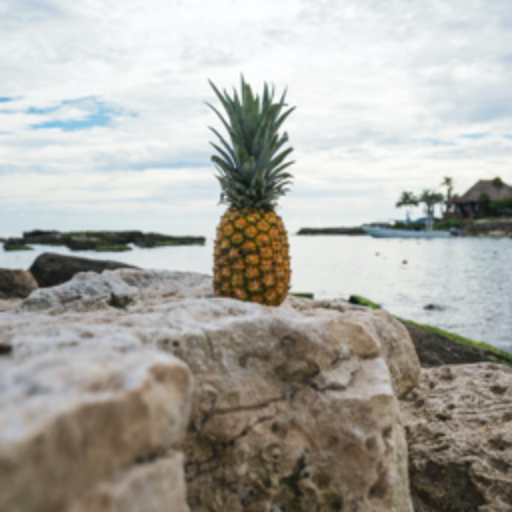} 
\end{minipage}\hfill
\begin{minipage}{0.16\textwidth}
\vspace{2mm}
\includegraphics[width=\linewidth]{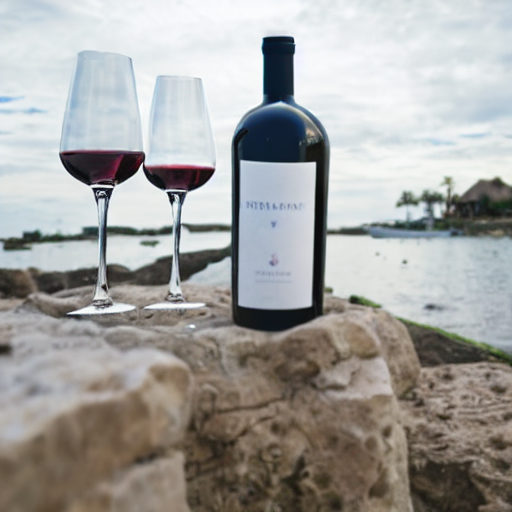} 
\end{minipage}
\vspace{-3mm}
\\
\hspace{1mm} \textcolor{black}{\textit{a plate of beefsteak and vegetable} \hspace{13mm} \textit{a piece of Oreo cookie and bread} \hspace{8mm} \textit{a bottle of wine and several wine cups}}
\\
\\
\vspace{5mm}
\hspace{-2mm}
\begin{minipage}{0.16\textwidth}
\vspace{-1mm}
\includegraphics[width=\linewidth]{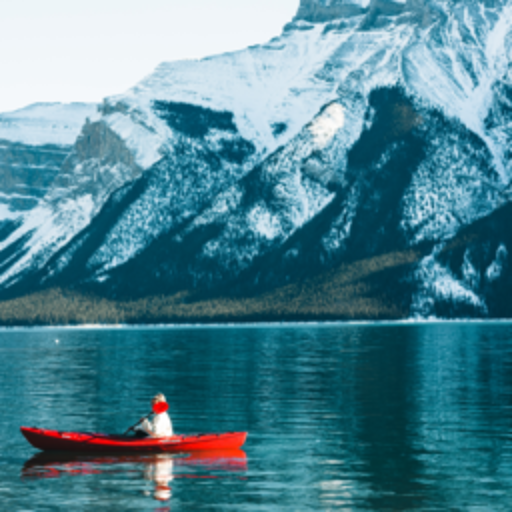}
\end{minipage}\hfill
\begin{minipage}{0.16\textwidth}
\vspace{-1mm}
\includegraphics[width=\linewidth]{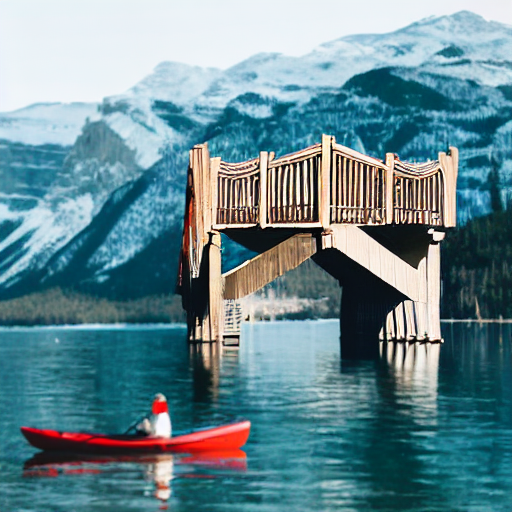}   
\end{minipage}\hfill
\begin{minipage}{0.16\textwidth}
\vspace{-1mm}
\includegraphics[width=\linewidth]{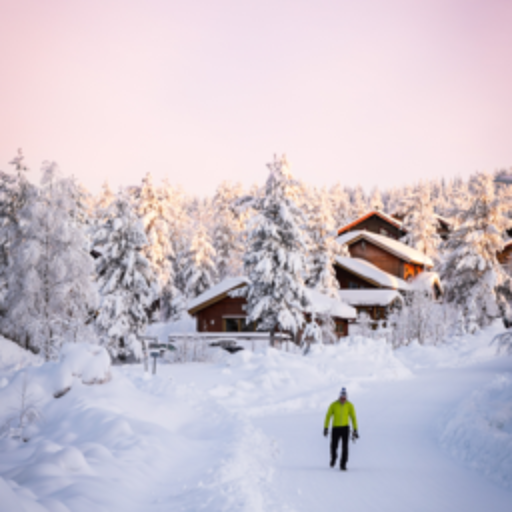} 
\end{minipage}\hfill
\begin{minipage}{0.16\textwidth}
\vspace{-1mm}
\includegraphics[width=\linewidth]{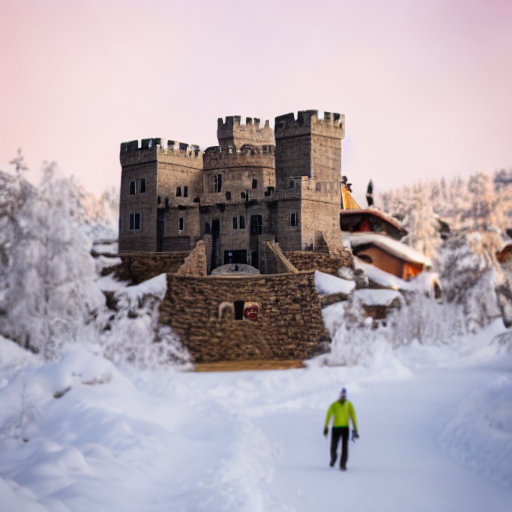} 
\end{minipage}\hfill
\begin{minipage}{0.16\textwidth}
\vspace{-1mm}
\includegraphics[width=\linewidth]{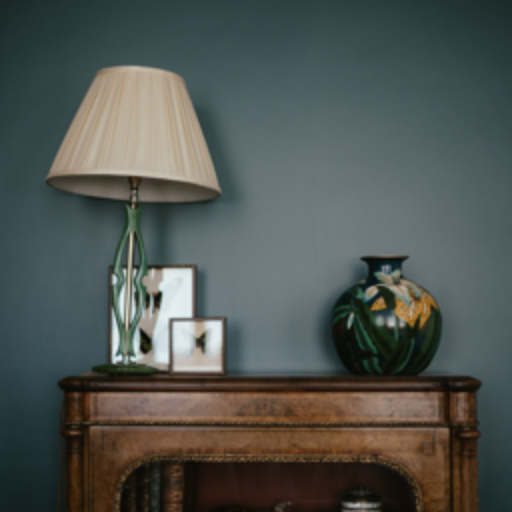} 
\end{minipage}\hfill
\begin{minipage}{0.16\textwidth}
\vspace{-1mm}
\includegraphics[width=\linewidth]{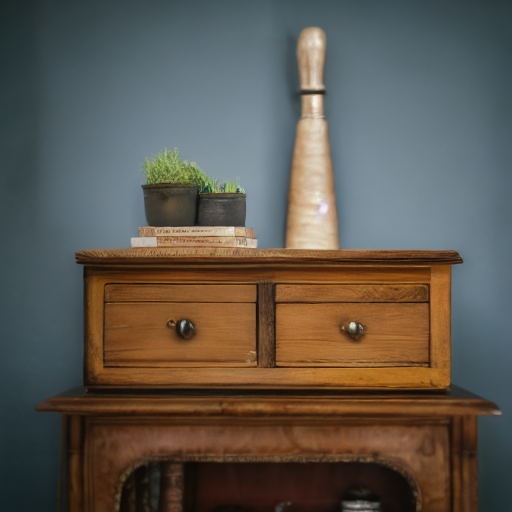} 
\end{minipage}
\vspace{-3mm}
\\
\textcolor{black}{\textit{a wooden bridge in front of the mountain} \hspace{1mm} \textit{a huge castle in the back of the person} \hspace{3mm} \textit{a wooden cabinet on top of the table}}
\\
\vspace{5mm}
\hspace{-2mm}
\begin{minipage}{0.16\textwidth}
\vspace{2mm}
\includegraphics[width=\linewidth]{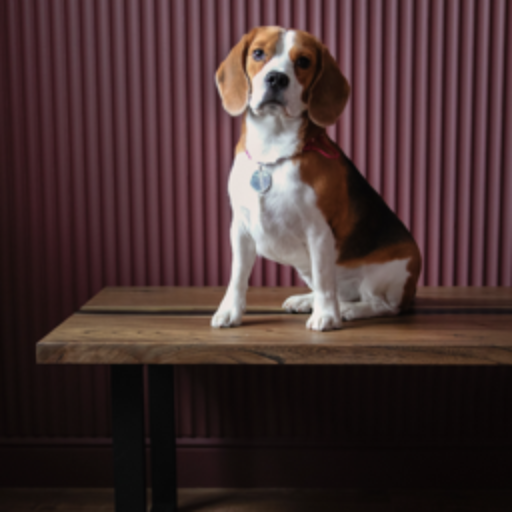}
\end{minipage}\hfill
\begin{minipage}{0.16\textwidth}
\vspace{2mm}
\includegraphics[width=\linewidth]{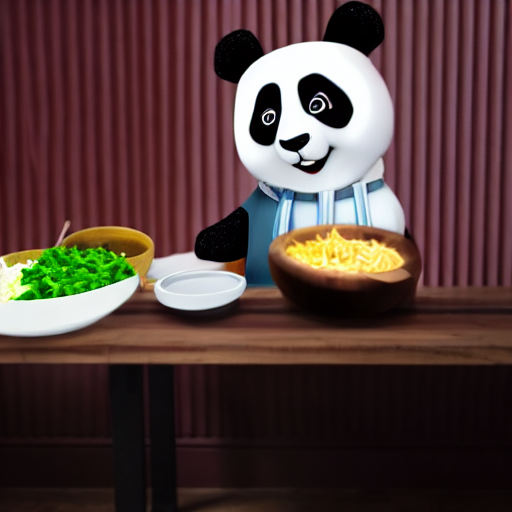}   
\end{minipage}\hfill
\begin{minipage}{0.16\textwidth}
\vspace{2mm}
\includegraphics[width=\linewidth]{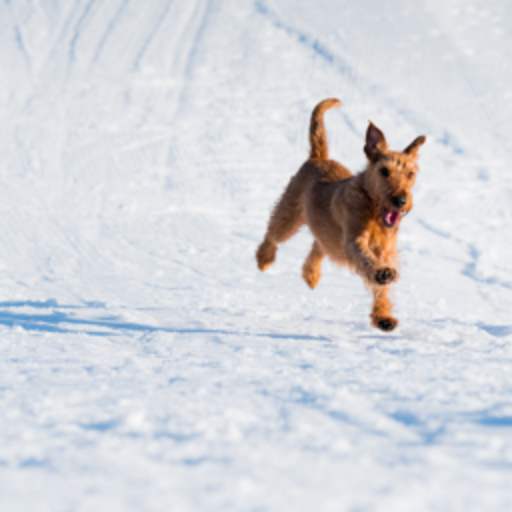} 
\end{minipage}\hfill
\begin{minipage}{0.16\textwidth}
\vspace{2mm}
\includegraphics[width=\linewidth]{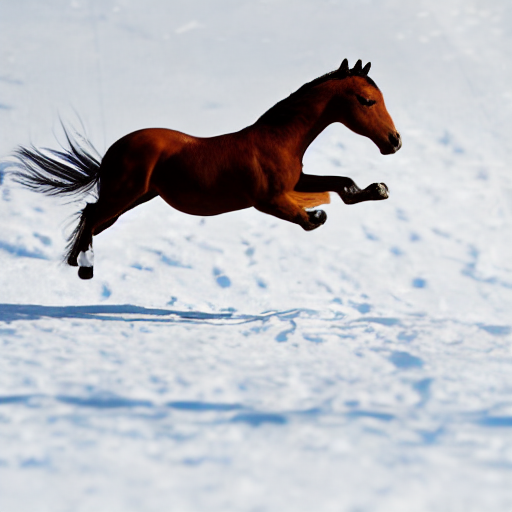} 
\end{minipage}\hfill
\begin{minipage}{0.16\textwidth}
\vspace{2mm}
\includegraphics[width=\linewidth]{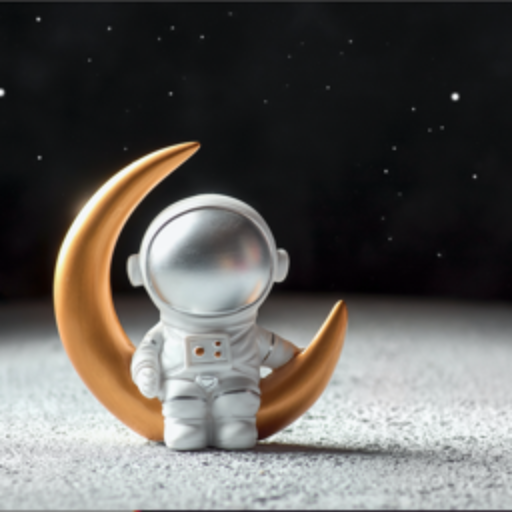} 
\end{minipage}\hfill
\begin{minipage}{0.16\textwidth}
\vspace{2mm}
\includegraphics[width=\linewidth]{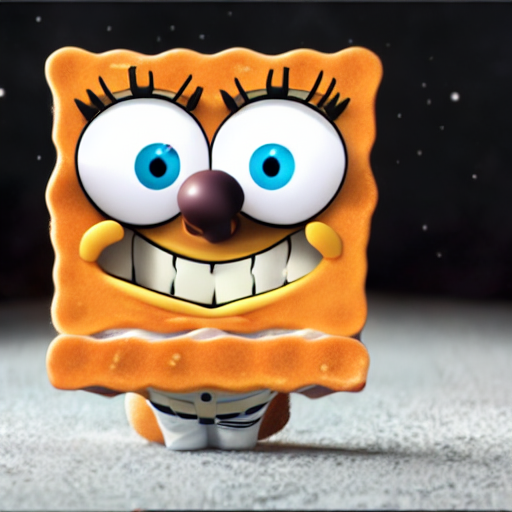} 
\end{minipage}
\vspace{-3mm}
\\
\textcolor{black}{
\small{\textit{A cartoon panda is preparing food. It wears}  \hspace{2mm} \textit{A little horse is jumping from the left side to} \hspace{1mm}  \textit{The cartoon character is smiling. It looks}} \\
\hspace{7mm} \textit{cloth which has blue and white colors and} \hspace{4mm} \textit{the right side. It jumps fast since its jumping \hspace{1mm} funny. The shape of its face is square, and}  \\
\hspace{7mm} \textit{there are several plates of food on the table}\hspace{2.5mm} \textit{stride is large, and it has red skin \hspace{16.5mm} its eyes and mouth are very large}} \\
\vspace{-4mm}
\captionof{figure}{\textbf{Image editing results with simple and complex prompts.} Given the input images and prompts, our method edits the image without requiring masks from the users. The learned region is omitted for better visualization. The \nth{1} row contains diverse prompts for one kind of object. The \nth{2} row displays prompts featuring multiple objects. The \nth{3} row shows prompts with geometric relations, and the last row presents prompts with extended length.
}
\label{fig:paragraph}
\end{figure*}

\subsection{Training Objectives}
\label{loss}
As the CLIP model \cite{radford2021learning} can 
estimate the similarity between images and texts, we employ it to guide our image editing based on user-specified prompts.

To train our models, we propose a composite editing loss that consists of three components: 1) the CLIP guidance loss $\mathcal{L}_{\text{Clip}}$ stands for the cosine distance between features extracted from generated images and texts, specifically derived from the last layers of CLIP's encoders. 2) the directional loss $\mathcal{L}_{\text{Dir}}$ \cite{patashnik2021styleclip} controls the direction of the applied edit within the CLIP space \cite{gal2022stylegan,patashnik2021styleclip}, and 3) the structural loss $\mathcal{L}_{\text{Str}}$ takes into account the self-similarity \cite{kolkin2019style,shechtman2007matching} of features between source and generated images, which facilitates editing in texture and appearance while preserving the original spatial layout of objects in the source images. The total loss $\mathcal{L}$ and each loss term are: 
\begin{equation}
\label{loss1}
 \mathcal{L} = \lambda_{C}\mathcal{L}_{\text{Clip}} +\lambda_{S}\mathcal{L}_{\text{Str}} +\lambda_{D}\mathcal{L}_{\text{Dir}},
\end{equation}
 \begin{equation}
\label{loss2}
 \hspace{-5mm}\mathcal{L}_{\text{Clip}} = \mathcal{D}_{\text{cos}}(E_{\text{v}}(X_{o}), E_{\text{t}}(T)),
 \end{equation}
\begin{equation}
\label{loss4}
\hspace{-4mm} \mathcal{L}_{\text{Str}} = \vert\vert Q(f_{X_{o}}) - Q(f_{X}) \vert\vert_{2},
 \end{equation}
  \begin{equation}
\label{loss3}
\mathcal{L}_{\text{Dir}} = \mathcal{D}_{\text{cos}}(E_{\text{v}}(X_{o})-E_{\text{v}}(X), E_{\text{t}}(T)-E_{\text{t}}(T_{\text{ROI}})),
 \end{equation}
 where $E_{\text{v}}$ and $E_{\text{t}}$ are the visual and textual encoder of the CLIP model. We empirically set the weights $\lambda_{C}=1$, $\lambda_{D}=1$, and $\lambda_{S}=1$ for our composite editing loss. Here, $X$, $T$, and $X_o$ denote the input image, text prompt, and the edited image by the proposed region, respectively.  $f_{X_{o}}$ and $f_{X}$ indicate the visual features of $X_{o}$ and $X$ from the last layer of CLIP's visual encoder, while $Q(f_{X_{o}})$ and $Q(f_{X})$ denote the similarity matrix of $f_{X_{o}}$ and $f_{X}$ respectively. For simplicity, we use the cosine distance $\mathcal{D}_{\text{cos}}$ to measure the similarity between images and texts. Note that $T_{\text{ROI}}$ represents a given region-of-interest of the source image for editing (\eg, in Figure \ref{figure5}, when $T$ is ``a big tree with many flowers in the center'', then $T_{\text{ROI}}$ could be ``tree'').

 During training, our loss functions encourage the region generator to produce appropriate regions for editing by taking into account the similarity between the edited images and the given text descriptions. 
 
\subsection{Inference}
\label{inference}
During the inference process, we define a quality score to rank the edited images generated from different anchor points and select the image with the highest score for presentation to the user.

While there exist more advanced methods, we use a simple weighted average to compute the quality score:
\begin{equation}
    \label{score}
    S = \alpha \cdot S_{\text{t2i}} + \beta \cdot S_{\text{i2i}},
\end{equation}
where $S_{\text{t2i}}$ estimates the cosine similarity scores between the given text descriptions and the edited images, $S_{\text{i2i}}$ measures the cosine similarity scores between the source images and the edited images, and $\alpha$ and $\beta$ are the coefficients to control the influences of $S_{\text{t2i}}$ and $S_{\text{i2i}}$. We adopt the features extracted from the last layer of CLIP's encoders for similarity calculation.

In our experiments, we set $\alpha$ and $\beta$ as 2 and 1 respectively, since a higher value for $\alpha$ can encourage the model to place more weight on the faithfulness of text-conditioned image editing.
The edited image with the highest quality score $S$ is chosen as the final edited image.

\subsection{Compatibility with Pretrained Editing Models}
Our proposed region generator can be integrated with various image editing models \cite{chang2022maskgit,rombach2022high,avrahami2022blended,nichol2021glide} for modifying the content of source images conditioning on the prompts, and to demonstrate its versatility, we apply it to two distinct image synthesis models: non-autoregressive transformers as used in MaskGIT~\cite{chang2022maskgit} or Muse~\cite{chang2023muse}, as well as diffusion U-Nets \cite{ronneberger2015u} as used in Stable Diffusion \cite{rombach2022high}. 

The transformer and diffusion models represent distinct base editing models to verify the applicability of the proposed method. It is worth noting that MaskGIT and Muse are transformers that operate over discrete tokens created by a VQ autoencoder \cite{van2017neural}, unlike diffusion models \cite{ho2020denoising,song2020denoising,rombach2022high} operating within the continuous space. As a result, the latent spaces in MaskGIT and Muse are only compatible with box-like masks and lack the precision for pixel-level masks in image editing. 

For our experiments, we use the official MaskGIT model instead of the Muse model~\cite{chang2023muse}, which is not publicly available. We also limit the text prompt to the class vocabulary that the model is trained on.

\begin{figure*}[htp]
\begin{minipage}{0.139\linewidth}
\captionof*{figure}{\small{\textcolor{black}{Input Image}}}
\vspace{-3mm}
\includegraphics[width=\linewidth,height=\linewidth]{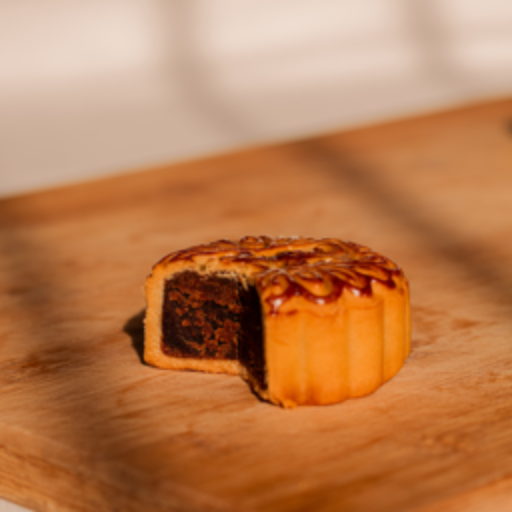}
\end{minipage}\hfill
\begin{minipage}{0.139\textwidth}
\captionof*{figure}{\small{\textcolor{black}{Plug-and-Play}}}
\vspace{-3mm}
\includegraphics[width=\linewidth,height=\linewidth]{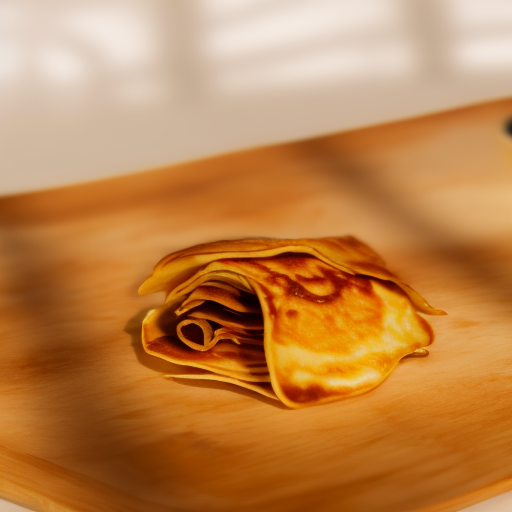}   
\end{minipage}\hfill
\begin{minipage}{0.139\textwidth}
\captionof*{figure}{\small{\textcolor{black}{InstructPix2Pix}}}
\vspace{-3mm}
\includegraphics[width=\linewidth,height=\linewidth]{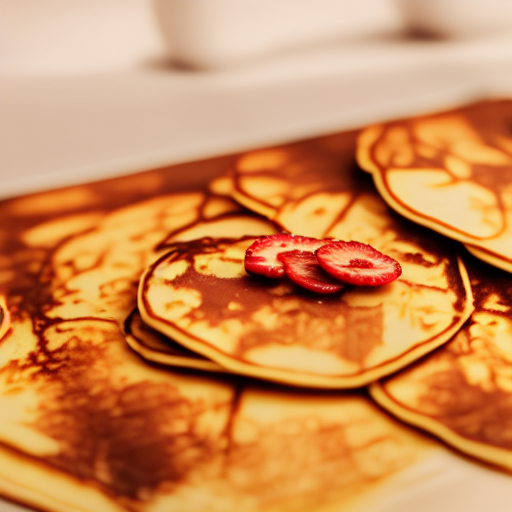}   
\end{minipage}\hfill
\begin{minipage}{0.139\textwidth}
\captionof*{figure}{\small{\textcolor{black}{Null-text}}}
\vspace{-3mm}
\includegraphics[width=\linewidth,height=\linewidth]{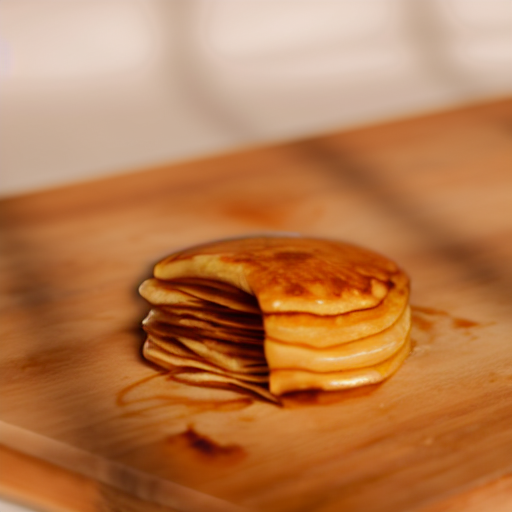} 
\end{minipage}\hfill
\begin{minipage}{0.139\textwidth}
\captionof*{figure}{\small{\textcolor{black}{DiffEdit}}}
\vspace{-3mm}
\includegraphics[width=\linewidth,height=\linewidth]{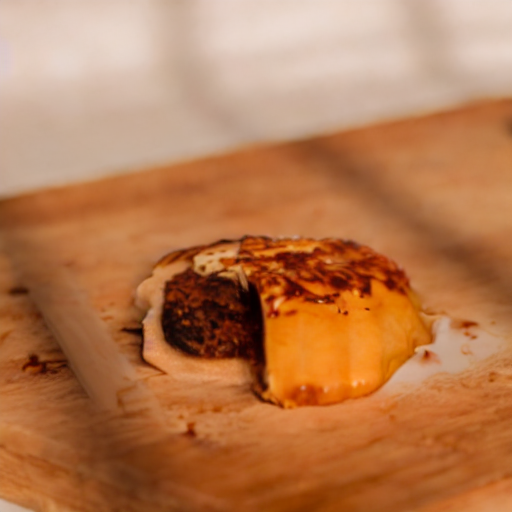} 
\end{minipage}\hfill
\begin{minipage}{0.139\textwidth}
\captionof*{figure}{\small{\textcolor{black}{MasaCtrl}}}
\vspace{-3mm}
\includegraphics[width=\linewidth,height=\linewidth]{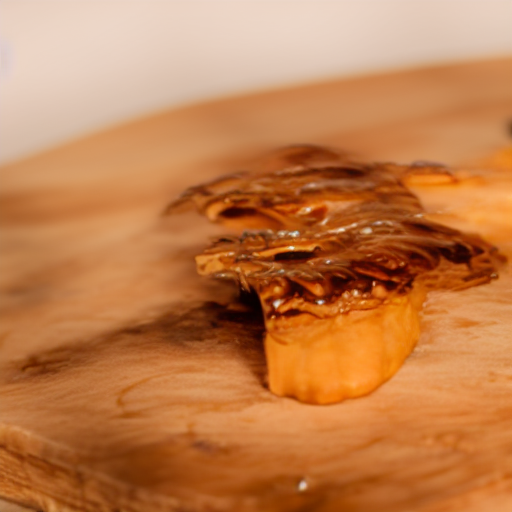} 
\end{minipage}\hfill
\begin{minipage}{0.139\textwidth}
\captionof*{figure}{\small{\textcolor{black}{Ours}}}
\vspace{-3mm}
\includegraphics[width=\linewidth,height=\linewidth]{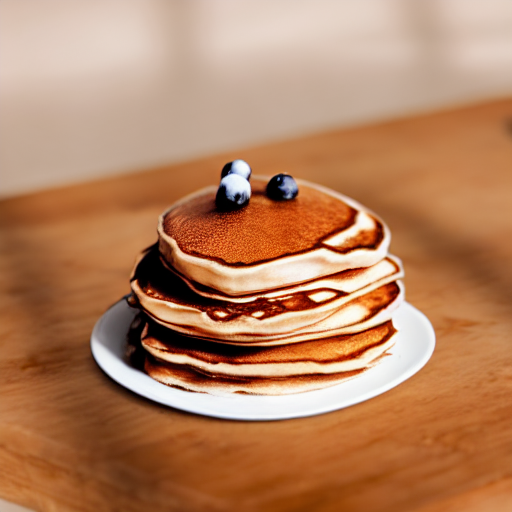} 
\end{minipage}
\vspace{1mm}
\\
\hspace{-2mm} \small{\textcolor{black}{\textsl{Editing Text:}}}   \hspace{57mm} \textcolor{black}{\textit{a dish of pancake}}
\vspace{4mm}
\\
\begin{minipage}{0.139\linewidth}
\vspace{-3mm}
\includegraphics[width=\linewidth,height=\linewidth]{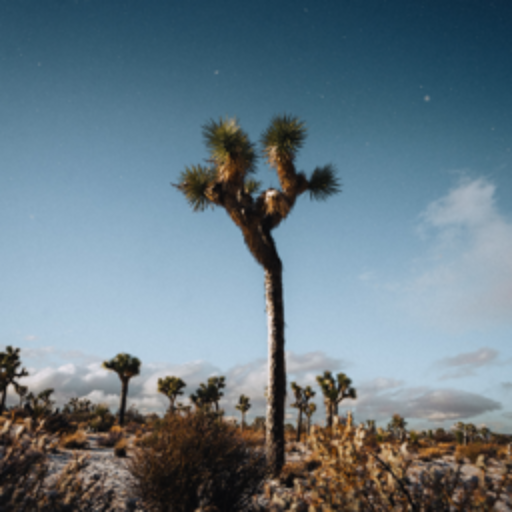}
\end{minipage}\hfill
\begin{minipage}{0.139\linewidth}
\vspace{-3mm}
\includegraphics[width=\linewidth,height=\linewidth]{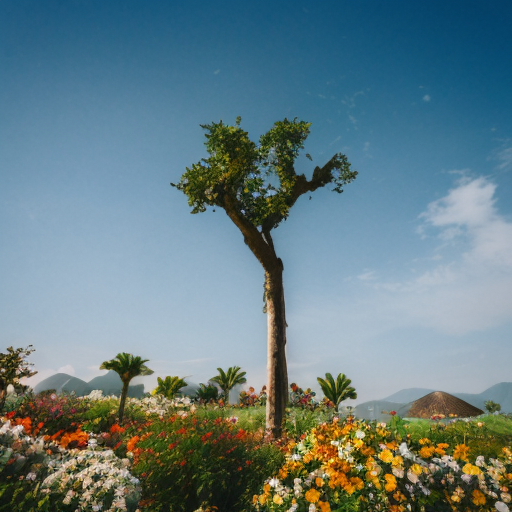}
\end{minipage}\hfill
\begin{minipage}{0.139\textwidth}
\vspace{-3mm}
\includegraphics[width=\linewidth,height=\linewidth]{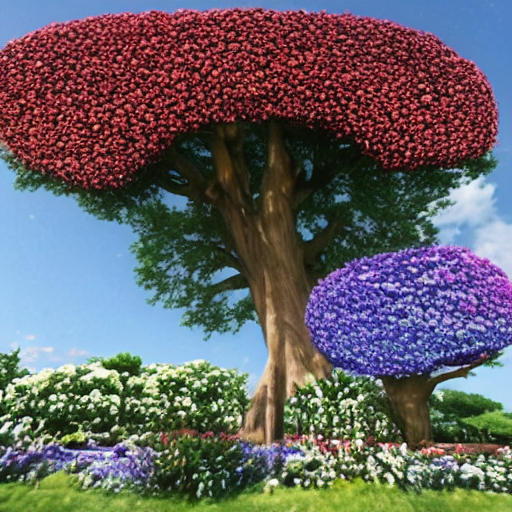}   
\end{minipage}\hfill
\begin{minipage}{0.139\textwidth}
\vspace{-3mm}
\includegraphics[width=\linewidth,height=\linewidth]{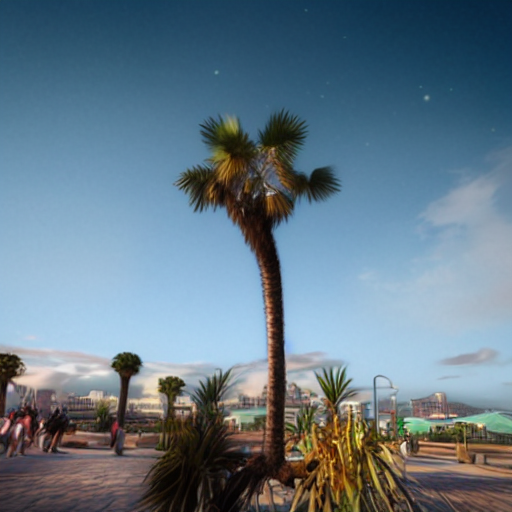} 
\end{minipage}\hfill
\begin{minipage}{0.139\textwidth}
\vspace{-3mm}
\includegraphics[width=\linewidth,height=\linewidth]{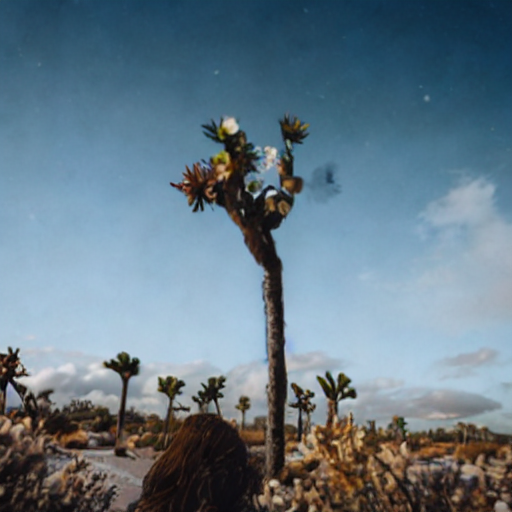} 
\end{minipage}\hfill
\begin{minipage}{0.139\textwidth}
\vspace{-3mm}
\includegraphics[width=\linewidth,height=\linewidth]{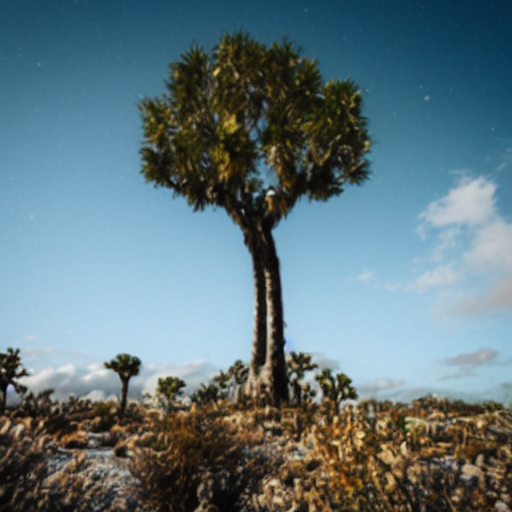} 
\end{minipage}\hfill
\begin{minipage}{0.139\textwidth}
\vspace{-3mm}
\includegraphics[width=\linewidth,height=\linewidth]{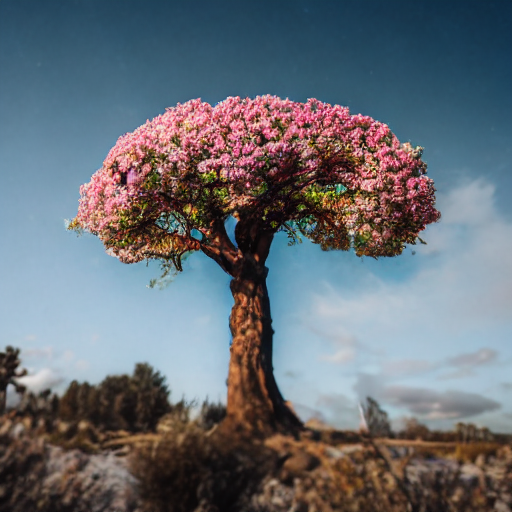} 
\end{minipage}
\vspace{1mm}
\\
\hspace{-2mm} \small{\textcolor{black}{\textsl{Editing Text:}}}   \hspace{41mm} \textcolor{black}{\textit{a big tree with many flowers in the center}}
\vspace{4mm}
\\
\begin{minipage}{0.139\linewidth}
\vspace{-3mm}
\includegraphics[width=\linewidth,height=\linewidth]{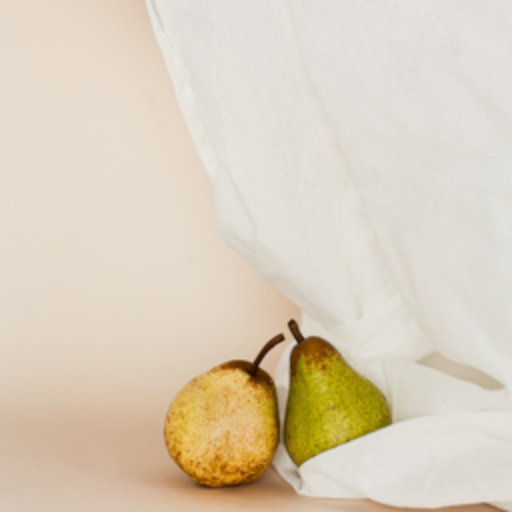}
\end{minipage}\hfill
\begin{minipage}{0.139\linewidth}
\vspace{-3mm}
\includegraphics[width=\linewidth,height=\linewidth]{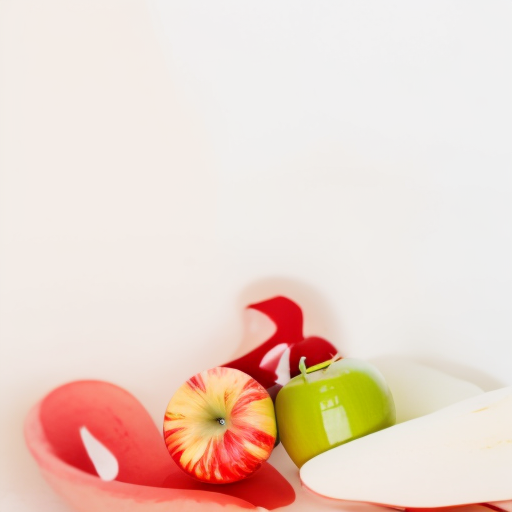}
\end{minipage}\hfill
\begin{minipage}{0.139\textwidth}
\vspace{-3mm}
\includegraphics[width=\linewidth,height=\linewidth]{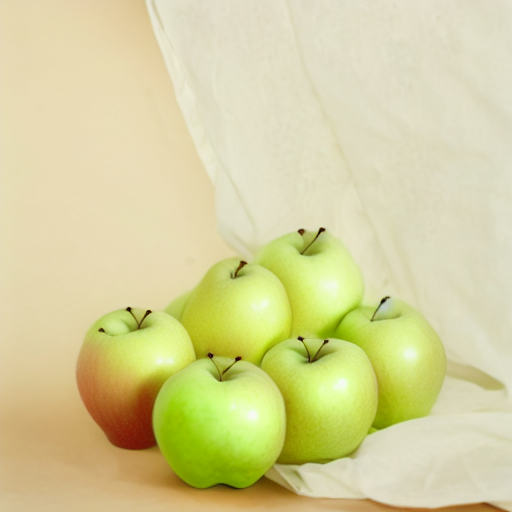}   
\end{minipage}\hfill
\begin{minipage}{0.139\textwidth}
\vspace{-3mm}
\includegraphics[width=\linewidth,height=\linewidth]{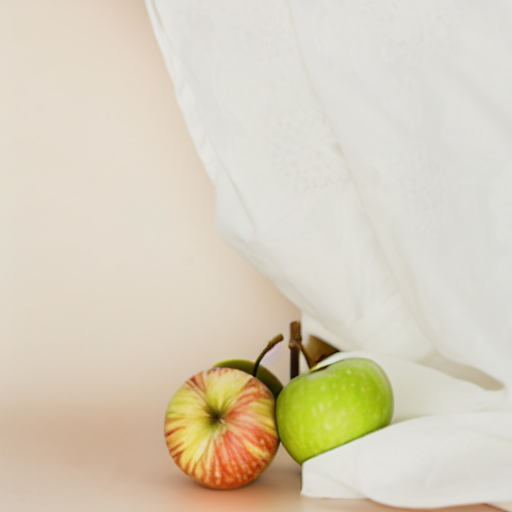} 
\end{minipage}\hfill
\begin{minipage}{0.139\textwidth}
\vspace{-3mm}
\includegraphics[width=\linewidth,height=\linewidth]{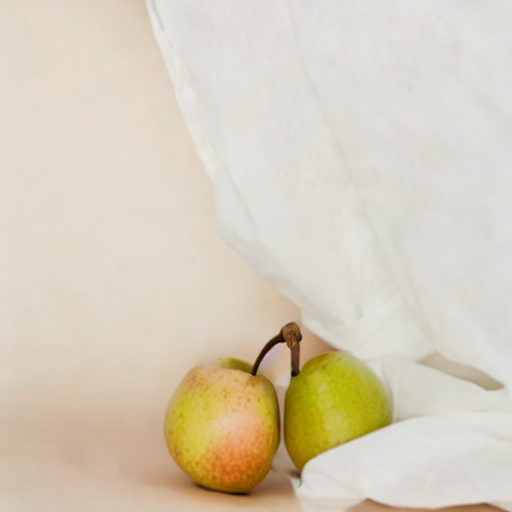} 
\end{minipage}\hfill
\begin{minipage}{0.139\textwidth}
\vspace{-3mm}
\includegraphics[width=\linewidth,height=\linewidth]{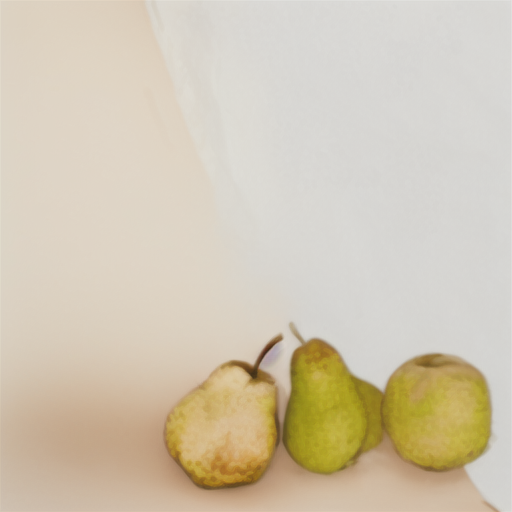} 
\end{minipage}\hfill
\begin{minipage}{0.139\textwidth}
\vspace{-3mm}
\includegraphics[width=\linewidth,height=\linewidth]{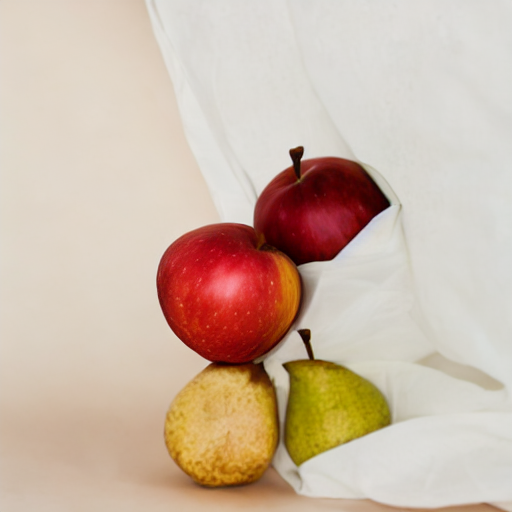} 
\end{minipage}
\vspace{1mm}

\hspace{-1mm} \small{\textcolor{black}{\textsl{Editing Text:}}}   \hspace{53mm} \textcolor{black}{\textit{several apples and pears}}
\vspace{4mm}
\\
\begin{minipage}{0.139\linewidth}
\vspace{-3mm}
\includegraphics[width=\linewidth,height=\linewidth]{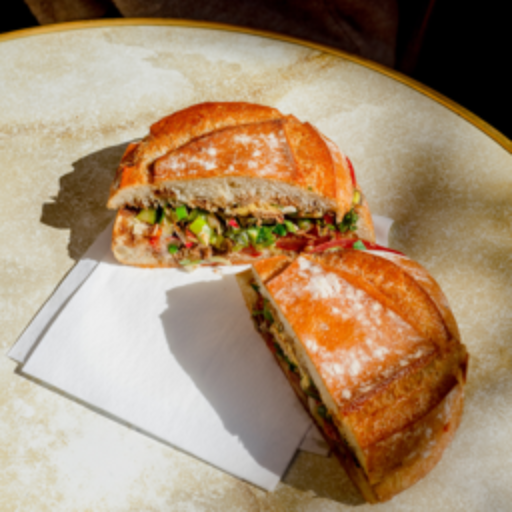}
\end{minipage}\hfill
\begin{minipage}{0.139\linewidth}
\vspace{-3mm}
\includegraphics[width=\linewidth,height=\linewidth]{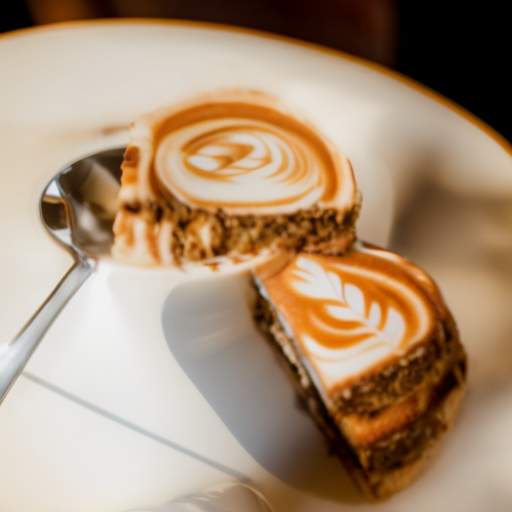}
\end{minipage}\hfill
\begin{minipage}{0.139\textwidth}
\vspace{-3mm}
\includegraphics[width=\linewidth,height=\linewidth]{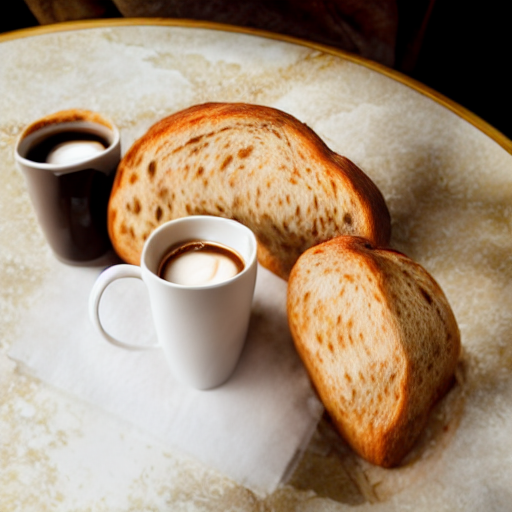}   
\end{minipage}\hfill
\begin{minipage}{0.139\textwidth}
\vspace{-3mm}
\includegraphics[width=\linewidth,height=\linewidth]{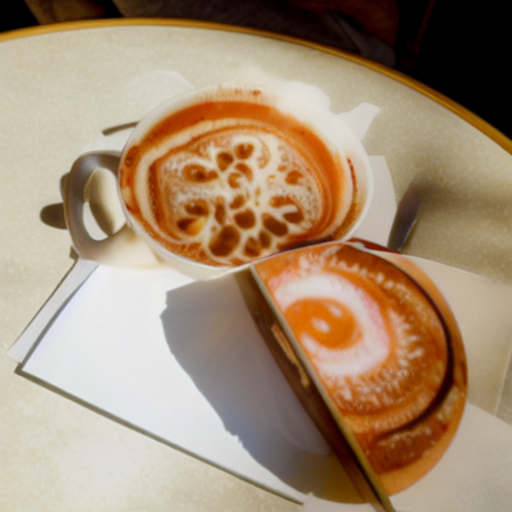} 
\end{minipage}\hfill
\begin{minipage}{0.139\textwidth}
\vspace{-3mm}
\includegraphics[width=\linewidth,height=\linewidth]{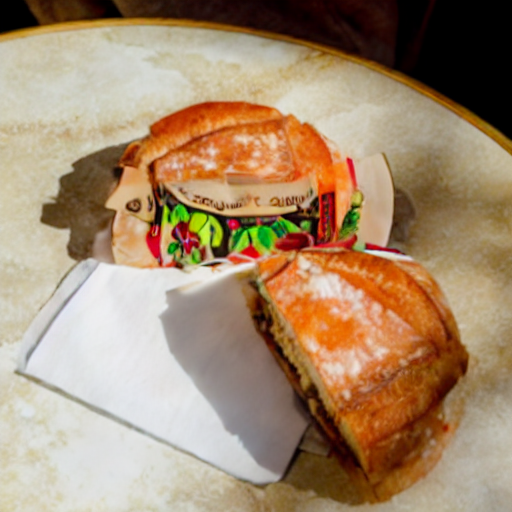} 
\end{minipage}\hfill
\begin{minipage}{0.139\textwidth}
\vspace{-3mm}
\includegraphics[width=\linewidth,height=\linewidth]{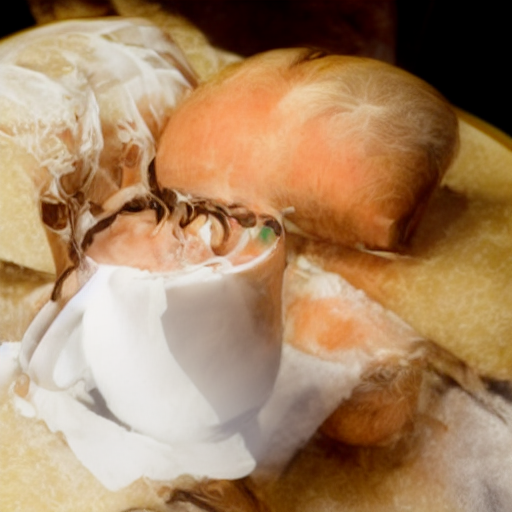} 
\end{minipage}\hfill
\begin{minipage}{0.139\textwidth}
\vspace{-3mm}
\includegraphics[width=\linewidth,height=\linewidth]{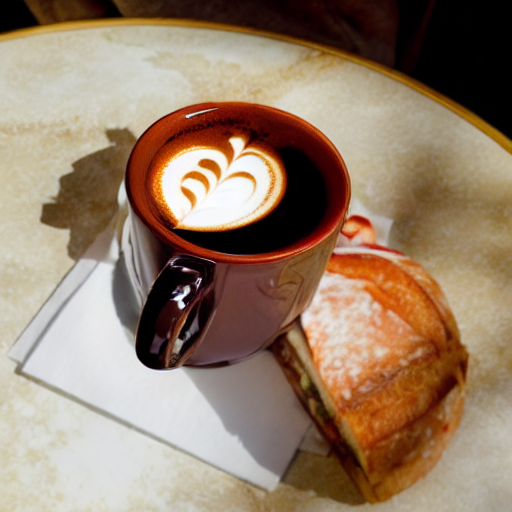} 
\end{minipage}
\vspace{1mm}
\\
\hspace{-2mm} \small{\textcolor{black}{\textsl{Editing Text:}}}   \hspace{47mm} \textcolor{black}{\textit{a cup of coffee next to the bread}}
\vspace{4mm}
\\
\begin{minipage}{0.139\linewidth}
\vspace{-3mm}
\includegraphics[width=\linewidth,height=\linewidth]{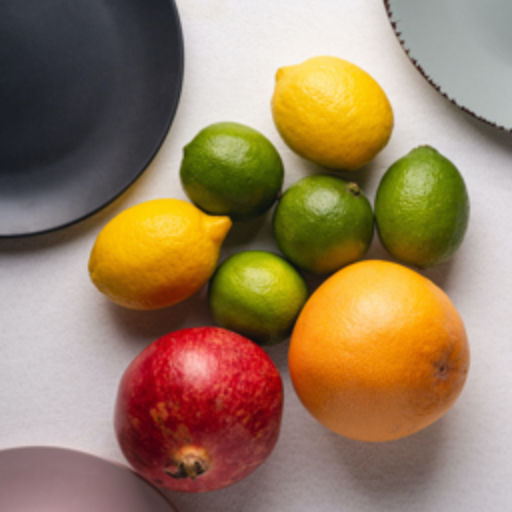}
\end{minipage}\hfill
\begin{minipage}{0.139\linewidth}
\vspace{-3mm}
\includegraphics[width=\linewidth,height=\linewidth]{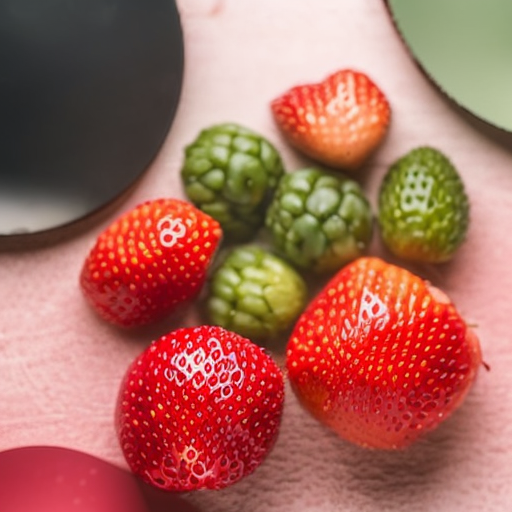}
\end{minipage}\hfill
\begin{minipage}{0.139\textwidth}
\vspace{-3mm}
\includegraphics[width=\linewidth,height=\linewidth]{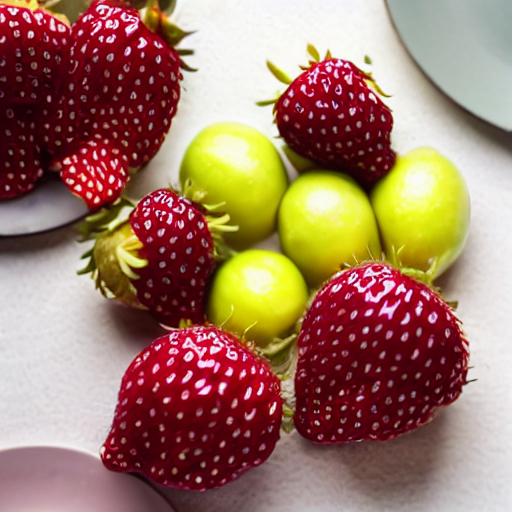}   
\end{minipage}\hfill
\begin{minipage}{0.139\textwidth}
\vspace{-3mm}
\includegraphics[width=\linewidth,height=\linewidth]{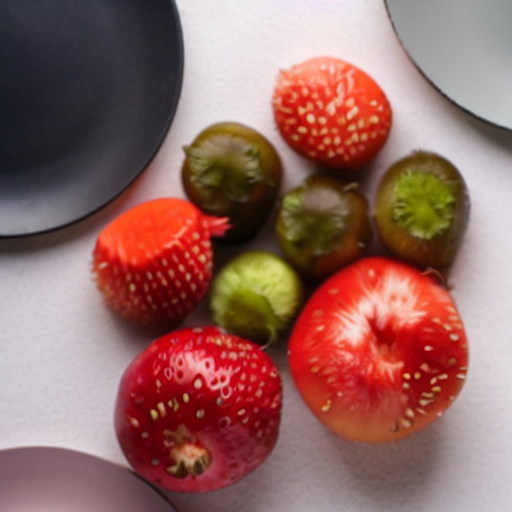} 
\end{minipage}\hfill
\begin{minipage}{0.139\textwidth}
\vspace{-3mm}
\includegraphics[width=\linewidth,height=\linewidth]{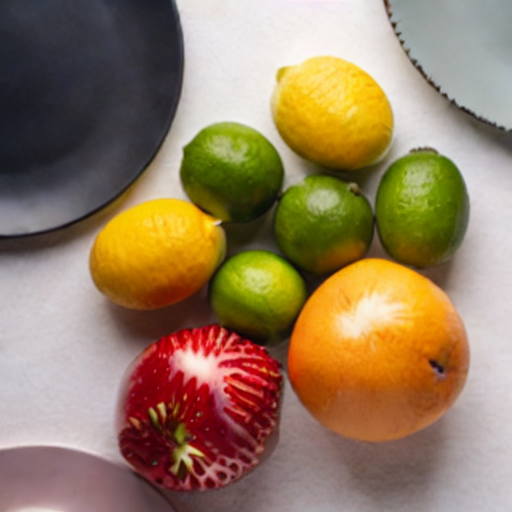} 
\end{minipage}\hfill
\begin{minipage}{0.139\textwidth}
\vspace{-3mm}
\includegraphics[width=\linewidth,height=\linewidth]{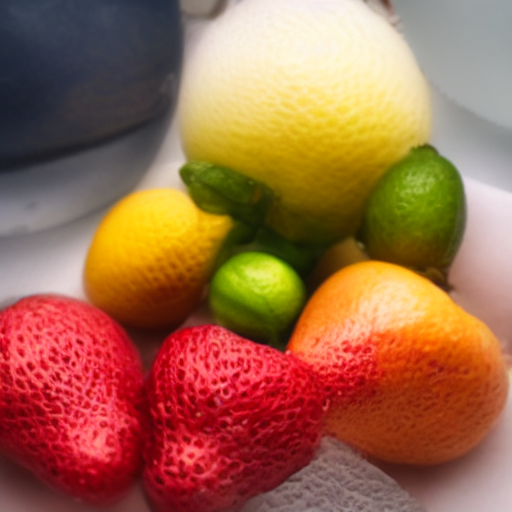} 
\end{minipage}\hfill
\begin{minipage}{0.139\textwidth}
\vspace{-3mm}
\includegraphics[width=\linewidth,height=\linewidth]{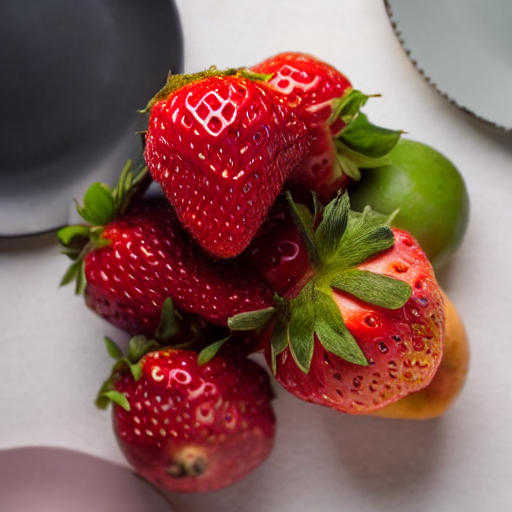} 
\end{minipage}
\vspace{1mm}
\\
\hspace{-2mm} \small{\textsl{Editing Text:}}   \hspace{46mm} \textcolor{black}{\textit{some strawberries and other fruit}}
\vspace{4mm}
\\
\begin{minipage}{0.139\linewidth}
\vspace{-3mm}
\includegraphics[width=\linewidth,height=\linewidth]{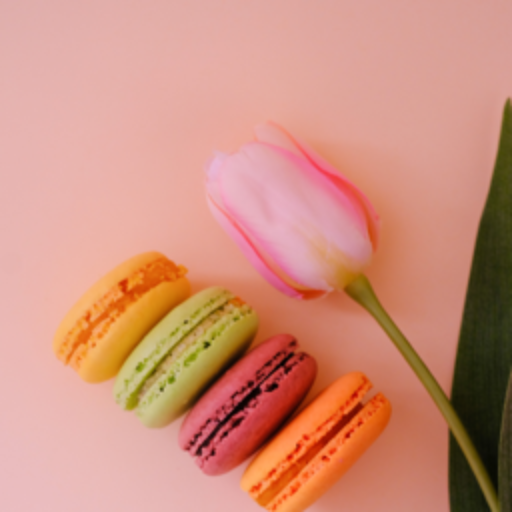}
\end{minipage}\hfill
\begin{minipage}{0.139\linewidth}
\vspace{-3mm}
\includegraphics[width=\linewidth,height=\linewidth]{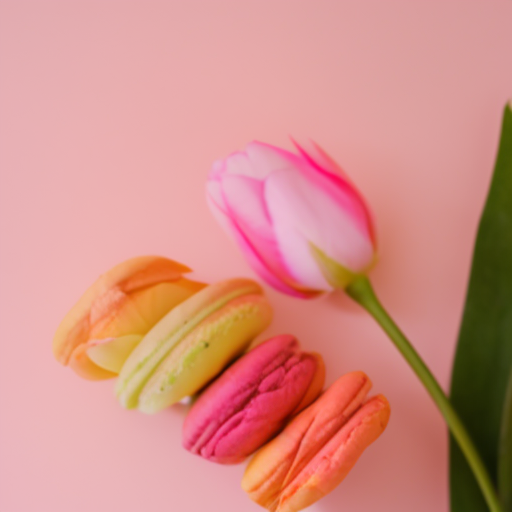}
\end{minipage}\hfill
\begin{minipage}{0.139\textwidth}
\vspace{-3mm}
\includegraphics[width=\linewidth,height=\linewidth]{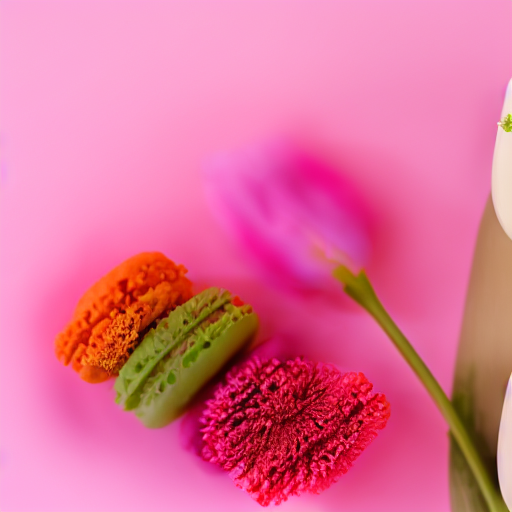}   
\end{minipage}\hfill
\begin{minipage}{0.139\textwidth}
\vspace{-3mm}
\includegraphics[width=\linewidth,height=\linewidth]{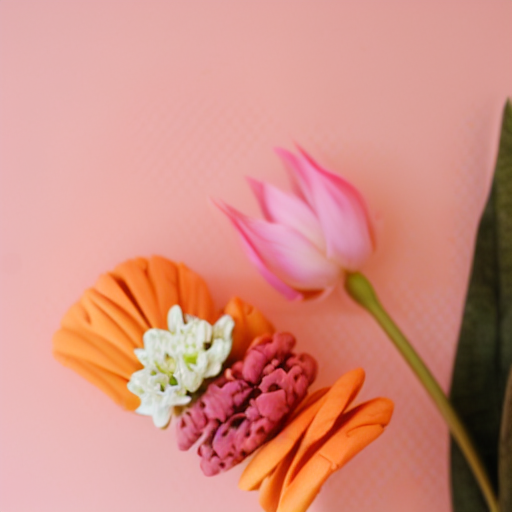} 
\end{minipage}\hfill
\begin{minipage}{0.139\textwidth}
\vspace{-3mm}
\includegraphics[width=\linewidth,height=\linewidth]{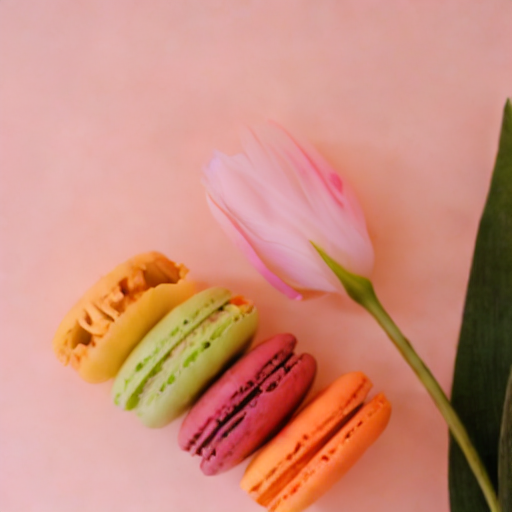} 
\end{minipage}\hfill
\begin{minipage}{0.139\textwidth}
\vspace{-3mm}
\includegraphics[width=\linewidth,height=\linewidth]{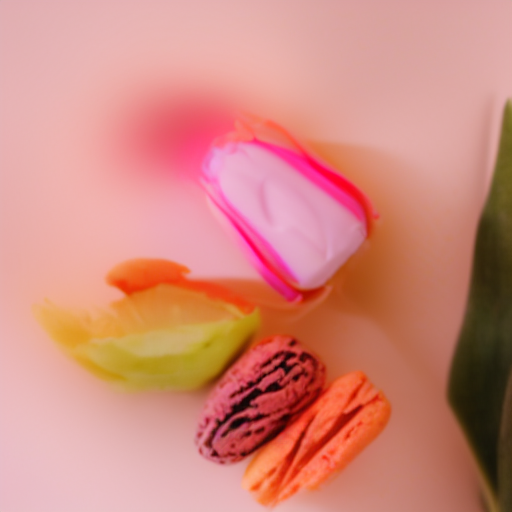} 
\end{minipage}\hfill
\begin{minipage}{0.139\textwidth}
\vspace{-3mm}
\includegraphics[width=\linewidth,height=\linewidth]{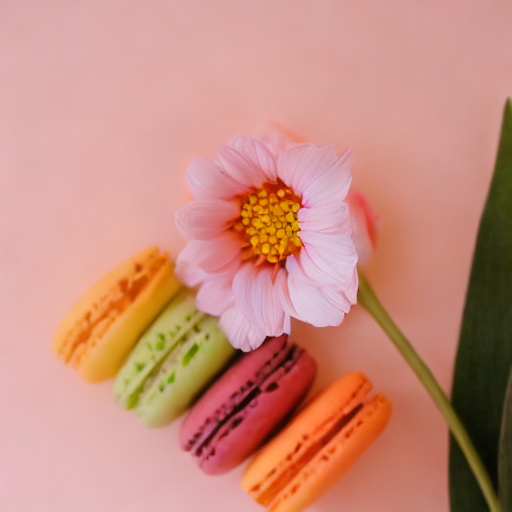} 
\end{minipage}
\vspace{1mm}
\\
\hspace{-2mm} \small{\textcolor{black}{\textsl{Editing Text:}}}   \hspace{48mm} \textcolor{black}{\textit{a blooming flower and dessert}}
\vspace{-1mm}
\captionof{figure}{\textbf{Comparison with existing methods.} We compare our method with existing text-driven image editing methods. From left to right: Input image, Plug-and-Play \cite{tumanyan2023plug}, InstructPix2Pix \cite{brooks2023instructpix2pix}, Null-text \cite{mokady2023null}, DiffEdit \cite{couairon2022diffedit}, MasaCtrl \cite{cao2023masactrl}, and ours.
\vspace{1.5mm}
}
\label{figure5}
\end{figure*}

\begin{figure*}[htp]
\begin{minipage}{0.16\linewidth}
\captionof*{figure}{\small{\textcolor{black}{Input Image}}}
\vspace{-3mm}
\includegraphics[width=\linewidth]{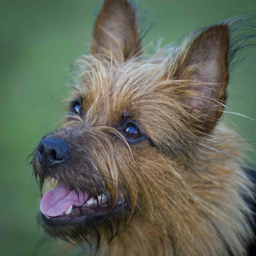}
\end{minipage}\hfill
\begin{minipage}{0.16\textwidth}
\captionof*{figure}{\small{\textcolor{black}{Edited Image}}}
\vspace{-3mm}
\includegraphics[width=\linewidth]{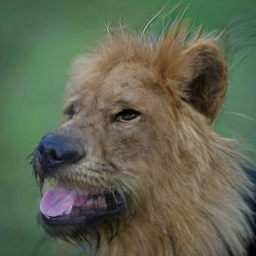}   
\end{minipage}\hfill
\begin{minipage}{0.16\textwidth}
\captionof*{figure}{\small{\textcolor{black}{Input Image}}}
\vspace{-3mm}
\includegraphics[width=\linewidth]{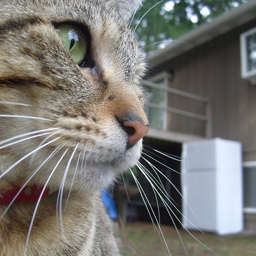} 
\end{minipage}\hfill
\begin{minipage}{0.16\textwidth}
\captionof*{figure}{\small{\textcolor{black}{Edited Image}}}
\vspace{-3mm}
\includegraphics[width=\linewidth]{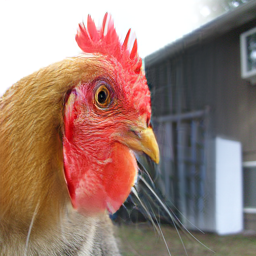} 
\end{minipage}\hfill
\begin{minipage}{0.16\textwidth}
\captionof*{figure}{\small{\textcolor{black}{Input Image}}}
\vspace{-3mm}
\includegraphics[width=\linewidth]{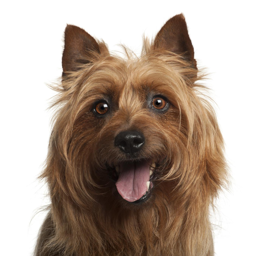} 
\end{minipage}\hfill
\begin{minipage}{0.16\textwidth}
\captionof*{figure}{\small{\textcolor{black}{Edited Image}}}
\vspace{-3mm}
\includegraphics[width=\linewidth]{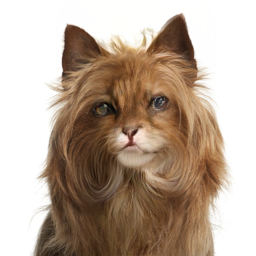} 
\end{minipage}
\vspace{1mm}
\\
\small{\textcolor{black}{\textsl{Editing Text:}}} \hspace{5mm} \textcolor{black}{\textit{a lion}   \hspace{48mm}   \textit{a rooster} \hspace{47mm}  \textit{cat face}}
\vspace{-2mm}
\captionof{figure}{\textbf{Generated results using MaskGIT \cite{chang2022maskgit} as the image synthesis model.} Aside from using the Stable Diffusion, our method can also generate reasonable editing results with the non-autoregressive transformer-based MaskGIT.}
\label{fig:maskgit}
\end{figure*}

\section{Experimental Results}
\paragraph{Implementation Details.}
\label{implementation}

In our evaluation, we collect high-resolution and free-to-use images covering a variety of objects from Unsplash ({\small \url{https://unsplash.com/}}).
For edit-region generation, the total number of bounding box proposals (\ie, $M$) is 7 and the CLIP guidance model is initialized with ViT-B/16 weights. We do not use super-resolution models to enhance the quality of the resultant images. By default, we adopt the pre-trained Stable Diffusion-v-1-2 as our editing model.
Our main experiments are conducted using two A5000 GPUs, where we train the model for 5 epochs using Adam optimizer \cite{kingma2014adam} with an initial learning rate of 0.003. 

\subsection{Qualitative Evaluation}
We assess the performance of our proposed method on a diverse set of high-quality images featuring various objects. Figure~\ref{fig:paragraph} shows that our approach takes an image and a language description to perform mask-free edits. We display complex text prompts that feature one category of object (the \nth{1} row), multiple objects (the \nth{2} row), geometric relations (the \nth{3} row), and long paragraphs (the \nth{4} row).

\subsection{Comparisons with Prior Work}
We compare our method with five state-of-the-art text-driven image editing approaches:
\textbf{Plug-and-Play} \cite{tumanyan2023plug} preserves the semantic layout of the source image by injecting features from the source image into the generation process of the target image.
\textbf{InstructPix2Pix} \cite{brooks2023instructpix2pix} first utilizes GPT-3 and Stable Diffusion to produce paired training data for image editing. It then trains a diffusion model with classifier-free guidance under conditions.
\textbf{Null-text Inversion} \cite{mokady2023null} enables text-based image editing with Stable Diffusion, using an initial DDIM inversion \cite{song2020denoising, dhariwal2021diffusion} as a pivot for optimization, tuning only the null-text embedding in classifier-free guidance.
\textbf{DiffEdit} \cite{couairon2022diffedit} automatically generates masks for the regions that require editing by contrasting predictions of the diffusion model conditioned on different text prompts.
\textbf{MasaCtrl} \cite{cao2023masactrl} performs text-based non-rigid image editing by converting self-attention in diffusion models into mutual self-attention, and it extracts the masks from the cross-attention maps as the editing regions.

In all experiments, we report the results of the compared methods using their official code, except for DiffEdit\footnote{As there's no official code of DiffEdit, we use the code \url{https://github.com/Xiang-cd/DiffEdit-stable-diffusion}}. Figure~\ref{figure5} displays the editing results from existing methods.
We have the following observations, InstructPix2Pix inevitably leads to undesired changes in the global appearance (e.g., background). DiffEdit and MasaCtrl will yield unsatisfactory results when using the more complex prompts containing multiple objects. Other methods generate less realistic results (\eg, coffee in the \nth{4} row) or results that do not correspond with the text prompt (\eg, only one apple and pear in the \nth{3} row).

\begin{table}[!t]
\centering
\begin{tabular}{lp{4.2cm}<{\centering}}
\toprule
Compared Methods & Preference for Ours \\ \midrule
vs. Plug-and-Play~\cite{tumanyan2023plug}            & 80.5\% $\pm$\mytiny{1.9\%}  \\
vs. InstructPix2Pix~\cite{brooks2023instructpix2pix} & 73.2\% $\pm$\mytiny{2.2\%}   \\
vs. Null-text~\cite{mokady2023null}                  & 88.2\% $\pm$\mytiny{1.6\%}   \\
vs. DiffEdit~\cite{couairon2022diffedit}             & 91.9\% $\pm$\mytiny{1.3\%}   \\
vs. MasaCtrl~\cite{cao2023masactrl}                  & 90.8\% $\pm$\mytiny{1.4\%}   \\ \midrule
Average                                              & \textbf{84.9\%}   \\ \bottomrule
\end{tabular}
\vspace{-2mm}
\caption{\textbf{User studies.} We show the percentage (mean, std) of user preference for our approach over compared methods.}
\label{tab:user_study}
\end{table}

\begin{figure}[t]
\begin{minipage}{0.24\linewidth}
\captionof*{figure}{\small{\textcolor{black}{Input Image}}}
\vspace{-3mm}
\includegraphics[width=\linewidth,height=\linewidth]{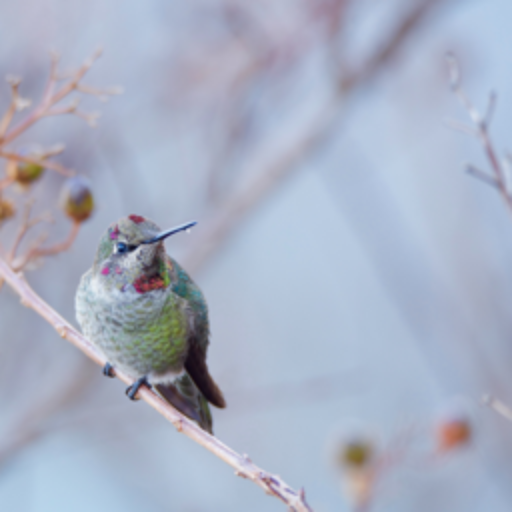}
\end{minipage}\hfill
\begin{minipage}{0.24\linewidth}
\captionof*{figure}{\small{\textcolor{black}{Result}}}
\vspace{-3mm}
\includegraphics[width=\linewidth,height=\linewidth]{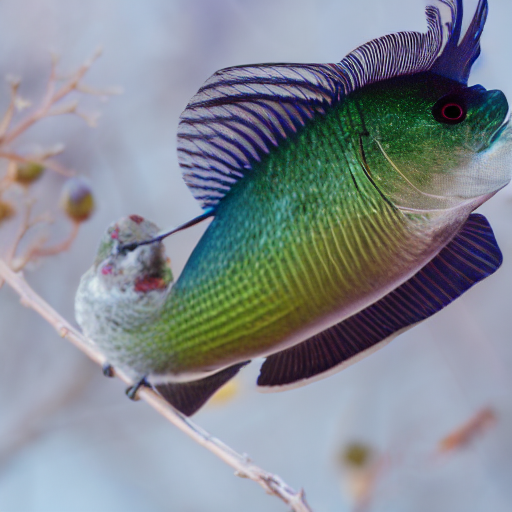}
\end{minipage}\hfill
\begin{minipage}{0.24\linewidth}
\captionof*{figure}{\small{\textcolor{black}{Input Image}}}
\vspace{-3mm}
\includegraphics[width=\linewidth,height=\linewidth]{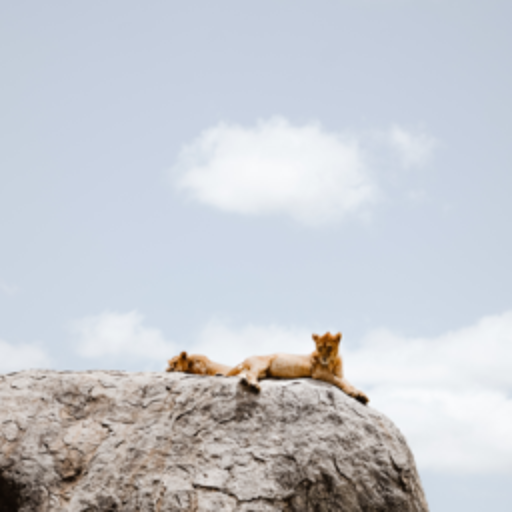}
\end{minipage}\hfill
\begin{minipage}{0.24\linewidth}
\captionof*{figure}{\small{\textcolor{black}{Result}}}
\vspace{-3mm}
\includegraphics[width=\linewidth,height=\linewidth]{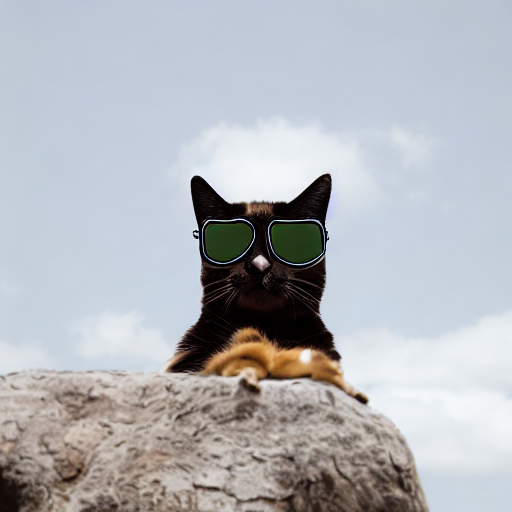}
\end{minipage}
\vspace{2mm}
\\
\hspace{-2mm} \small{\textcolor{black}{\textsl{Editing Text:}}}  \textcolor{black}{\textit{a large flying fish} \hspace{5mm} \textit{a cat wearing sunglasses}}
\vspace{1mm}
\vspace{-6mm}
\\
\captionof{figure}{\textbf{Failure cases.} We show two failure cases generated by our method.}
\label{figure:failure}
\end{figure}

\subsection{User Study}

To evaluate the quality of the edited images, we conduct a user study using 60 input images and text prompts. We employ paired comparisons to measure user preference.
In each test, we show an input image, a text prompt, and two edited images generated by our method and one of the compared approaches. We ask the subject to choose the one that performs better in coherence with the text prompt while maintaining fidelity to the input image.

There are 203 participants in this study, where each participant evaluates 40 pairs of images. The image set and compared method for each image are randomly selected for each user. The order in each comparison pair is shuffled when presenting to each user. All the methods are compared for the same number of times.

Table~\ref{tab:user_study} shows the user study results. The proposed method performs favorably against all five compared approaches. On average, our method is preferred in \textbf{84.9\%} of all the comparisons, which demonstrates the effectiveness of the proposed method.

\vspace{-2mm}
\paragraph{Failure cases.} We present some failure cases of our approach to analyze the reasons. As shown in Figure~\ref{figure:failure}, the failure results can be caused by improper anchor point initialization, especially when the anchor points fall into the background area.

\subsection{Ablation Study}
\paragraph{Compatibility with image synthesis models.}

To demonstrate the generalizability of the proposed method, we conduct experiments using MaskGIT \cite{chang2022maskgit}, a distinct image generative transformer. As shown in Figure~\ref{fig:maskgit}, we can generate results that adhere to the text prompt while preserving the background content. Note that the latent spaces within MaskGIT exclusively accommodate box-like masks, lacking the requisite precision for manipulating pixel-level masks in the context of image editing.

\begin{figure}[!t]
\begin{minipage}{0.24\linewidth}
\captionof*{figure}{\small{\textcolor{black}{Input Image}}}
\vspace{-3mm}
\includegraphics[width=\linewidth,height=\linewidth]{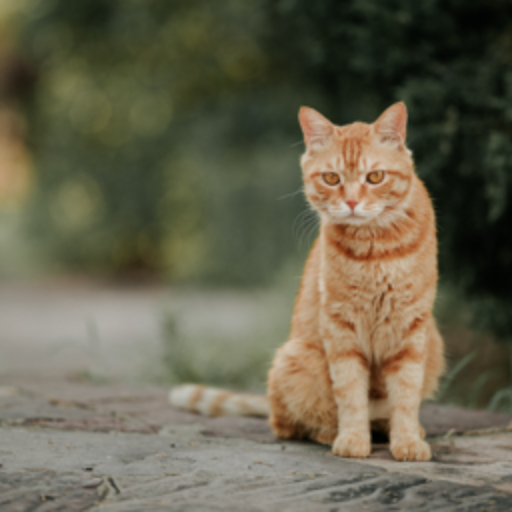}
\end{minipage}\hfill
\begin{minipage}{0.24\linewidth}
\captionof*{figure}{\small{\textcolor{black}{w/o $\mathcal{L}_{Dir}$}}}
\vspace{-3mm}
\includegraphics[width=\linewidth,height=\linewidth]{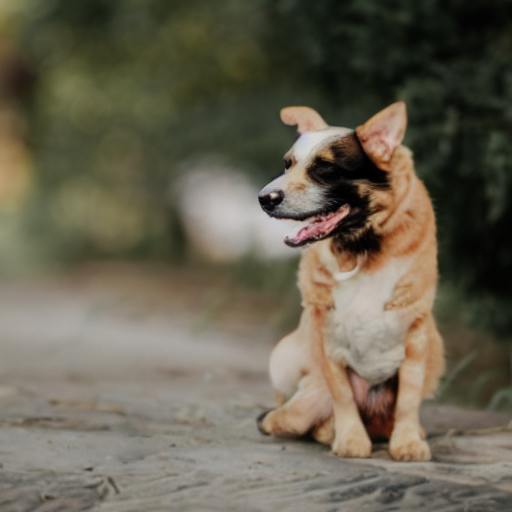}
\end{minipage}\hfill
\begin{minipage}{0.24\linewidth}
\captionof*{figure}{\small{\textcolor{black}{w/o $\mathcal{L}_{Str}$}}}
\vspace{-3mm}
\includegraphics[width=\linewidth,height=\linewidth]{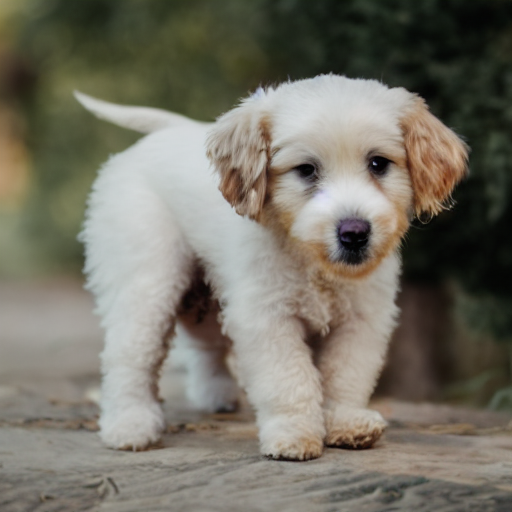}
\end{minipage}\hfill
\begin{minipage}{0.24\linewidth}
\captionof*{figure}{\small{\textcolor{black}{Ours}}}
\vspace{-3mm}
\includegraphics[width=\linewidth,height=\linewidth]{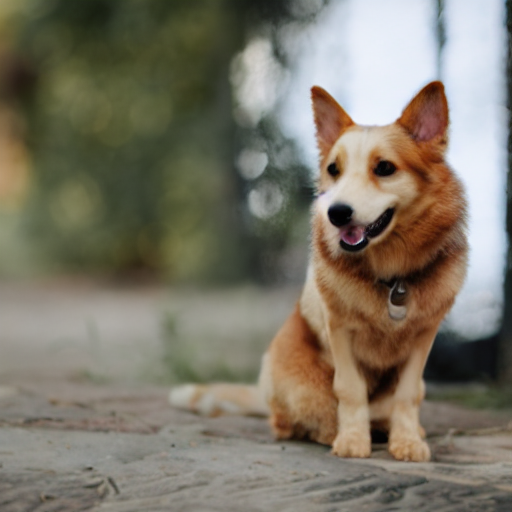}
\end{minipage}
\vspace{2mm}
\\
\hspace{-2mm} \small{\textcolor{black}{\textsl{Editing Text:}}}   \hspace{7mm} \textcolor{black}{\textit{a high quality photo of a lovely dog}}
\vspace{-5mm}
\\
\captionof{figure}{\textbf{Effect of different loss components}. The \nth{2} and \nth{3} columns present results without $\mathcal{L}_{Dir}$ and $\mathcal{L}_{Str}$ respectively. The last column is generated by the model using all loss components.}
\label{figure:loss}
\vspace{-1mm}
\end{figure}

\paragraph{Effect of different loss components.} To evaluate the influences of different loss components in our training loss, in Figure~\ref{figure:loss}, we show the results generated without the directional loss $\mathcal{L}_{Dir}$ that controls the directional edit, or without the structural loss $\mathcal{L}_{Str}$ that focuses on preserving the appearance of the source image. 
We observe that the result without $\mathcal{L}_{Dir}$ does not fully match the context of the text prompt, and the result without $\mathcal{L}_{Str}$ fails to preserve the posture and shape of the object from the source image.
In contrast, our method using all loss components can generate results that adhere to the text prompt while preserving the concept in the source image.

\paragraph{Effect of region generation methods.}
In Table~\ref{tab:user_study2}, we present the user study results by comparing our method with two other baselines for bounding box generation. (1) \textbf{Random-anchor-random-size}: The editing regions are bounding boxes centered at anchor points uniformly sampled from the whole image, with height and width uniformly sampled from $[0, H]$ and $[0, W]$, where $H$ and $W$ are the height and width of the image. We clamp the regions exceeding the image boundary.
 (2) \textbf{DINO-anchor-random-size}: The editing regions are bounding boxes centered at anchor points selected from the DINO self-attention map, which are identical to those generated by our method, but with height and width uniformly sampled from $[0, H]$ and $[0, W]$. For both baselines, we use the same number of anchor points as our method, and select the image with the highest quality score $S$ to present to the user.

The results show that our method is preferred in \textbf{83.9\%} compared with Random-anchor baseline. Even when compared to the competitive baseline where the anchor point is selected from the DINO self-attention map with the randomly chosen bounding box size, the proposed method is still preferred in \textbf{71.0\%} of all comparisons. These results validate the effectiveness of our model in generating meaningful editing regions.

\begin{table}[!t]
\centering
\begin{tabular}{lc}
\toprule
Compared Methods & Preference for Ours \\ \midrule
vs. Random-anchor-random-size & 83.9\% $\pm$\mytiny{2.6\%}  \\
vs. DINO-anchor-random-size   & 71.0\% $\pm$\mytiny{3.2\%}  \\
\bottomrule

\end{tabular}
\vspace{-2mm}
\caption{\textbf{Ablation study of region generation methods.} We show the percentage (mean, std) of user preference for our approach over two compared baselines.}
\label{tab:user_study2}
\end{table}

\subsection{Limitation}
We observe two limitations of our method. First, the performance is affected by the choice of the self-supervised model, particularly regarding anchor initialization. Second, since no user-specified region guidance is provided, the predicted region may include background areas, resulting in unintentional modifications in certain image contents. To address this, we plan to model the mask using more fine-grained representations (\eg, patches).

\section{Conclusion}
In this paper, we propose a method for editing given images based on freely provided language descriptions, including paragraphs, without the need for user-specified edit regions. We introduce a region generation network and incorporate text-driven editing training losses to generate high-quality and realistic images. The proposed method seamlessly integrates with various image synthesis models. Experiments including user studies are conducted, demonstrating the competitive performance of our proposed method.

{
    \small
    \bibliographystyle{ieeenat_fullname}
    \bibliography{main}

\begin{thebibliography}{62}
\providecommand{\natexlab}[1]{#1}
\providecommand{\url}[1]{\texttt{#1}}
\expandafter\ifx\csname urlstyle\endcsname\relax
  \providecommand{\doi}[1]{doi: #1}\else
  \providecommand{\doi}{doi: \begingroup \urlstyle{rm}\Url}\fi

\bibitem[Avrahami et~al.(2022)Avrahami, Lischinski, and Fried]{avrahami2022blended}
Omri Avrahami, Dani Lischinski, and Ohad Fried.
\newblock Blended diffusion for text-driven editing of natural images.
\newblock In \emph{CVPR}, 2022.

\bibitem[Bar-Tal et~al.(2022)Bar-Tal, Ofri-Amar, Fridman, Kasten, and Dekel]{bar2022text2live}
Omer Bar-Tal, Dolev Ofri-Amar, Rafail Fridman, Yoni Kasten, and Tali Dekel.
\newblock {Text2LIVE}: Text-driven layered image and video editing.
\newblock In \emph{ECCV}, 2022.

\bibitem[Bau et~al.(2021)Bau, Andonian, Cui, Park, Jahanian, Oliva, and Torralba]{bau2021paint}
David Bau, Alex Andonian, Audrey Cui, YeonHwan Park, Ali Jahanian, Aude Oliva, and Antonio Torralba.
\newblock Paint by word.
\newblock \emph{arXiv:2103.10951}, 2021.

\bibitem[Brooks et~al.(2023)Brooks, Holynski, and Efros]{brooks2023instructpix2pix}
Tim Brooks, Aleksander Holynski, and Alexei~A Efros.
\newblock Instructpix2pix: Learning to follow image editing instructions.
\newblock In \emph{CVPR}, 2023.

\bibitem[Brown et~al.(2020)Brown, Mann, Ryder, Subbiah, Kaplan, Dhariwal, Neelakantan, Shyam, Sastry, Askell, et~al.]{brown2020language}
Tom Brown, Benjamin Mann, Nick Ryder, Melanie Subbiah, Jared~D Kaplan, Prafulla Dhariwal, Arvind Neelakantan, Pranav Shyam, Girish Sastry, Amanda Askell, et~al.
\newblock Language models are few-shot learners.
\newblock In \emph{NeurIPS}, 2020.

\bibitem[Cao et~al.(2023)Cao, Wang, Qi, Shan, Qie, and Zheng]{cao2023masactrl}
Mingdeng Cao, Xintao Wang, Zhongang Qi, Ying Shan, Xiaohu Qie, and Yinqiang Zheng.
\newblock Masactrl: Tuning-free mutual self-attention control for consistent image synthesis and editing.
\newblock In \emph{ICCV}, 2023.

\bibitem[Caron et~al.(2021)Caron, Touvron, Misra, J{\'e}gou, Mairal, Bojanowski, and Joulin]{caron2021emerging}
Mathilde Caron, Hugo Touvron, Ishan Misra, Herv{\'e} J{\'e}gou, Julien Mairal, Piotr Bojanowski, and Armand Joulin.
\newblock Emerging properties in self-supervised vision transformers.
\newblock In \emph{ICCV}, 2021.

\bibitem[Chang et~al.(2022)Chang, Zhang, Jiang, Liu, and Freeman]{chang2022maskgit}
Huiwen Chang, Han Zhang, Lu Jiang, Ce Liu, and William~T Freeman.
\newblock {MaskGIT}: Masked generative image transformer.
\newblock In \emph{CVPR}, 2022.

\bibitem[Chang et~al.(2023)Chang, Zhang, Barber, Maschinot, Lezama, Jiang, Yang, Murphy, Freeman, Rubinstein, et~al.]{chang2023muse}
Huiwen Chang, Han Zhang, Jarred Barber, AJ Maschinot, Jose Lezama, Lu Jiang, Ming-Hsuan Yang, Kevin Murphy, William~T Freeman, Michael Rubinstein, et~al.
\newblock Muse: Text-to-image generation via masked generative transformers.
\newblock \emph{arXiv:2301.00704}, 2023.

\bibitem[Couairon et~al.(2023)Couairon, Verbeek, Schwenk, and Cord]{couairon2022diffedit}
Guillaume Couairon, Jakob Verbeek, Holger Schwenk, and Matthieu Cord.
\newblock Diffedit: Diffusion-based semantic image editing with mask guidance.
\newblock In \emph{ICLR}, 2023.

\bibitem[Crowson()]{katherine1}
Katherine Crowson.
\newblock {CLIP guided diffusion HQ 256x256}.
\newblock \url{https://colab.research.google.com/drive/12a_Wrfi2_gwwAuN3VvMTwVMz9TfqctNj}.

\bibitem[Crowson et~al.(2022)Crowson, Biderman, Kornis, Stander, Hallahan, Castricato, and Raff]{crowson2022vqgan}
Katherine Crowson, Stella Biderman, Daniel Kornis, Dashiell Stander, Eric Hallahan, Louis Castricato, and Edward Raff.
\newblock {VQGAN-CLIP}: Open domain image generation and editing with natural language guidance.
\newblock In \emph{ECCV}, 2022.

\bibitem[Dhariwal and Nichol(2021)]{dhariwal2021diffusion}
Prafulla Dhariwal and Alexander Nichol.
\newblock Diffusion models beat {GANs} on image synthesis.
\newblock In \emph{NeurIPS}, 2021.

\bibitem[Ding et~al.(2022)Ding, Zheng, Hong, and Tang]{ding2022cogview2}
Ming Ding, Wendi Zheng, Wenyi Hong, and Jie Tang.
\newblock {CogView2}: Faster and better text-to-image generation via hierarchical transformers.
\newblock In \emph{NeurIPS}, 2022.

\bibitem[Dosovitskiy et~al.(2020)Dosovitskiy, Beyer, Kolesnikov, Weissenborn, Zhai, Unterthiner, Dehghani, Minderer, Heigold, Gelly, et~al.]{dosovitskiy2020image}
Alexey Dosovitskiy, Lucas Beyer, Alexander Kolesnikov, Dirk Weissenborn, Xiaohua Zhai, Thomas Unterthiner, Mostafa Dehghani, Matthias Minderer, Georg Heigold, Sylvain Gelly, et~al.
\newblock An image is worth 16x16 words: Transformers for image recognition at scale.
\newblock \emph{arXiv:2010.11929}, 2020.

\bibitem[Esser et~al.(2021)Esser, Rombach, and Ommer]{Esser21vqgan}
Patrick Esser, Robin Rombach, and Bj{\"{o}}rn Ommer.
\newblock Taming transformers for high-resolution image synthesis.
\newblock In \emph{CVPR}, 2021.

\bibitem[Gal et~al.(2022)Gal, Patashnik, Maron, Bermano, Chechik, and Cohen-Or]{gal2022stylegan}
Rinon Gal, Or Patashnik, Haggai Maron, Amit~H Bermano, Gal Chechik, and Daniel Cohen-Or.
\newblock {StyleGAN-NADA: CLIP-guided domain adaptation of image generators}.
\newblock \emph{ACM Transactions on Graphics (TOG)}, 41\penalty0 (4):\penalty0 1--13, 2022.

\bibitem[Gal et~al.(2023)Gal, Alaluf, Atzmon, Patashnik, Bermano, Chechik, and Cohen-Or]{gal2023textual}
Rinon Gal, Yuval Alaluf, Yuval Atzmon, Or Patashnik, Amit~H. Bermano, Gal Chechik, and Daniel Cohen-Or.
\newblock An image is worth one word: Personalizing text-to-image generation using textual inversion.
\newblock In \emph{ICLR}, 2023.

\bibitem[Girshick(2015)]{girshick2015fast}
Ross Girshick.
\newblock Fast {R-CNN}.
\newblock In \emph{ICCV}, 2015.

\bibitem[Gu et~al.(2022)Gu, Chen, Bao, Wen, Zhang, Chen, Yuan, and Guo]{gu2022vector}
Shuyang Gu, Dong Chen, Jianmin Bao, Fang Wen, Bo Zhang, Dongdong Chen, Lu Yuan, and Baining Guo.
\newblock Vector quantized diffusion model for text-to-image synthesis.
\newblock In \emph{CVPR}, 2022.

\bibitem[Hertz et~al.(2023)Hertz, Mokady, Tenenbaum, Aberman, Pritch, and Cohen-Or]{hertz2023prompt}
Amir Hertz, Ron Mokady, Jay Tenenbaum, Kfir Aberman, Yael Pritch, and Daniel Cohen-Or.
\newblock Prompt-to-prompt image editing with cross-attention control.
\newblock In \emph{ICLR}, 2023.

\bibitem[Ho et~al.(2020)Ho, Jain, and Abbeel]{ho2020denoising}
Jonathan Ho, Ajay Jain, and Pieter Abbeel.
\newblock Denoising diffusion probabilistic models.
\newblock In \emph{NeurIPS}, 2020.

\bibitem[Ho et~al.(2022)Ho, Chan, Saharia, Whang, Gao, Gritsenko, Kingma, Poole, Norouzi, Fleet, et~al.]{ho2022imagen}
Jonathan Ho, William Chan, Chitwan Saharia, Jay Whang, Ruiqi Gao, Alexey Gritsenko, Diederik~P Kingma, Ben Poole, Mohammad Norouzi, David~J Fleet, et~al.
\newblock Imagen video: High definition video generation with diffusion models.
\newblock \emph{arXiv:2210.02303}, 2022.

\bibitem[Jang et~al.(2016)Jang, Gu, and Poole]{jang2016categorical}
Eric Jang, Shixiang Gu, and Ben Poole.
\newblock Categorical reparameterization with gumbel-softmax.
\newblock \emph{arXiv:1611.01144}, 2016.

\bibitem[Karras et~al.(2020)Karras, Laine, Aittala, Hellsten, Lehtinen, and Aila]{karras2020analyzing}
Tero Karras, Samuli Laine, Miika Aittala, Janne Hellsten, Jaakko Lehtinen, and Timo Aila.
\newblock Analyzing and improving the image quality of {StyleGAN}.
\newblock In \emph{CVPR}, 2020.

\bibitem[Kawar et~al.(2022)Kawar, Zada, Lang, Tov, Chang, Dekel, Mosseri, and Irani]{kawar2022imagic}
Bahjat Kawar, Shiran Zada, Oran Lang, Omer Tov, Huiwen Chang, Tali Dekel, Inbar Mosseri, and Michal Irani.
\newblock Imagic: Text-based real image editing with diffusion models.
\newblock \emph{arXiv:2210.09276}, 2022.

\bibitem[Kim et~al.(2022)Kim, Kwon, and Ye]{kim2021diffusionclip}
Gwanghyun Kim, Taesung Kwon, and Jong~Chul Ye.
\newblock {DiffusionCLIP}: Text-guided diffusion models for robust image manipulation.
\newblock In \emph{CVPR}, 2022.

\bibitem[Kingma and Ba(2014)]{kingma2014adam}
Diederik~P Kingma and Jimmy Ba.
\newblock Adam: A method for stochastic optimization.
\newblock \emph{arXiv:1412.6980}, 2014.

\bibitem[Kolkin et~al.(2019)Kolkin, Salavon, and Shakhnarovich]{kolkin2019style}
Nicholas Kolkin, Jason Salavon, and Gregory Shakhnarovich.
\newblock Style transfer by relaxed optimal transport and self-similarity.
\newblock In \emph{CVPR}, 2019.

\bibitem[Kwon and Ye(2022)]{kwon2022clipstyler}
Gihyun Kwon and Jong~Chul Ye.
\newblock {CLIPstyler}: Image style transfer with a single text condition.
\newblock In \emph{CVPR}, 2022.

\bibitem[Li et~al.(2020)Li, Qi, Lukasiewicz, and Torr]{li2020manigan}
Bowen Li, Xiaojuan Qi, Thomas Lukasiewicz, and Philip~HS Torr.
\newblock {ManiGAN}: Text-guided image manipulation.
\newblock In \emph{CVPR}, 2020.

\bibitem[Lin et~al.(2021)Lin, Guo, and Lu]{lin2021self}
Yuanze Lin, Xun Guo, and Yan Lu.
\newblock Self-supervised video representation learning with meta-contrastive network.
\newblock In \emph{Proceedings of the IEEE/CVF international conference on computer vision}, pages 8239--8249, 2021.

\bibitem[Lin et~al.(2022)Lin, Xie, Chen, Xu, Zhu, and Yuan]{lin2022revive}
Yuanze Lin, Yujia Xie, Dongdong Chen, Yichong Xu, Chenguang Zhu, and Lu Yuan.
\newblock Revive: Regional visual representation matters in knowledge-based visual question answering.
\newblock \emph{Advances in Neural Information Processing Systems}, 35:\penalty0 10560--10571, 2022.

\bibitem[Lin et~al.(2023)Lin, Wei, Wang, Yuille, and Xie]{lin2023smaug}
Yuanze Lin, Chen Wei, Huiyu Wang, Alan Yuille, and Cihang Xie.
\newblock Smaug: Sparse masked autoencoder for efficient video-language pre-training.
\newblock In \emph{Proceedings of the IEEE/CVF International Conference on Computer Vision}, pages 2459--2469, 2023.

\bibitem[Lin et~al.(2024)Lin, Clark, and Torr]{lin2024dreampolisher}
Yuanze Lin, Ronald Clark, and Philip Torr.
\newblock Dreampolisher: Towards high-quality text-to-3d generation via geometric diffusion.
\newblock \emph{arXiv preprint arXiv:2403.17237}, 2024.

\bibitem[Liu et~al.(2021)Liu, Gong, Wu, Zhang, Su, and Liu]{liu2021fusedream}
Xingchao Liu, Chengyue Gong, Lemeng Wu, Shujian Zhang, Hao Su, and Qiang Liu.
\newblock {FuseDream}: Training-free text-to-image generation with improved {CLIP+GAN} space optimization.
\newblock \emph{arXiv:2112.01573}, 2021.

\bibitem[Meng et~al.(2021)Meng, He, Song, Song, Wu, Zhu, and Ermon]{meng2021sdedit}
Chenlin Meng, Yutong He, Yang Song, Jiaming Song, Jiajun Wu, Jun-Yan Zhu, and Stefano Ermon.
\newblock {SDEdit}: Guided image synthesis and editing with stochastic differential equations.
\newblock In \emph{ICLR}, 2021.

\bibitem[Mokady et~al.(2023)Mokady, Hertz, Aberman, Pritch, and Cohen-Or]{mokady2023null}
Ron Mokady, Amir Hertz, Kfir Aberman, Yael Pritch, and Daniel Cohen-Or.
\newblock Null-text inversion for editing real images using guided diffusion models.
\newblock In \emph{CVPR}, 2023.

\bibitem[Nichol et~al.(2021)Nichol, Dhariwal, Ramesh, Shyam, Mishkin, McGrew, Sutskever, and Chen]{nichol2021glide}
Alex Nichol, Prafulla Dhariwal, Aditya Ramesh, Pranav Shyam, Pamela Mishkin, Bob McGrew, Ilya Sutskever, and Mark Chen.
\newblock {GLIDE}: Towards photorealistic image generation and editing with text-guided diffusion models.
\newblock \emph{arXiv:2112.10741}, 2021.

\bibitem[Nichol and Dhariwal(2021)]{nichol2021improved}
Alexander~Quinn Nichol and Prafulla Dhariwal.
\newblock Improved denoising diffusion probabilistic models.
\newblock In \emph{ICML}, 2021.

\bibitem[Patashnik et~al.(2021)Patashnik, Wu, Shechtman, Cohen-Or, and Lischinski]{patashnik2021styleclip}
Or Patashnik, Zongze Wu, Eli Shechtman, Daniel Cohen-Or, and Dani Lischinski.
\newblock {StyleCLIP}: Text-driven manipulation of {StyleGAN} imagery.
\newblock In \emph{ICCV}, 2021.

\bibitem[Radford et~al.(2021)Radford, Kim, Hallacy, Ramesh, Goh, Agarwal, Sastry, Askell, Mishkin, Clark, et~al.]{radford2021learning}
Alec Radford, Jong~Wook Kim, Chris Hallacy, Aditya Ramesh, Gabriel Goh, Sandhini Agarwal, Girish Sastry, Amanda Askell, Pamela Mishkin, Jack Clark, et~al.
\newblock Learning transferable visual models from natural language supervision.
\newblock In \emph{ICML}, 2021.

\bibitem[Ramesh et~al.(2021)Ramesh, Pavlov, Goh, Gray, Voss, Radford, Chen, and Sutskever]{ramesh2021zero}
Aditya Ramesh, Mikhail Pavlov, Gabriel Goh, Scott Gray, Chelsea Voss, Alec Radford, Mark Chen, and Ilya Sutskever.
\newblock Zero-shot text-to-image generation.
\newblock In \emph{ICML}, 2021.

\bibitem[Ramesh et~al.(2022)Ramesh, Dhariwal, Nichol, Chu, and Chen]{Ramesh2022dalle2}
Aditya Ramesh, Prafulla Dhariwal, Alex Nichol, Casey Chu, and Mark Chen.
\newblock Hierarchical text-conditional image generation with {CLIP} latents.
\newblock \emph{arXiv:2204.06125}, 2022.

\bibitem[Rombach et~al.(2022)Rombach, Blattmann, Lorenz, Esser, and Ommer]{rombach2022high}
Robin Rombach, Andreas Blattmann, Dominik Lorenz, Patrick Esser, and Bj{\"o}rn Ommer.
\newblock High-resolution image synthesis with latent diffusion models.
\newblock In \emph{CVPR}, 2022.

\bibitem[Ronneberger et~al.(2015)Ronneberger, Fischer, and Brox]{ronneberger2015u}
Olaf Ronneberger, Philipp Fischer, and Thomas Brox.
\newblock U-net: Convolutional networks for biomedical image segmentation.
\newblock In \emph{MICCAI}, 2015.

\bibitem[Ruiz et~al.(2022)Ruiz, Li, Jampani, Pritch, Rubinstein, and Aberman]{ruiz2022dreambooth}
Nataniel Ruiz, Yuanzhen Li, Varun Jampani, Yael Pritch, Michael Rubinstein, and Kfir Aberman.
\newblock {DreamBooth}: Fine tuning text-to-image diffusion models for subject-driven generation.
\newblock \emph{arXiv:2208.12242}, 2022.

\bibitem[Saharia et~al.(2022)Saharia, Chan, Saxena, Li, Whang, Denton, Ghasemipour, Ayan, Mahdavi, Lopes, Salimans, Ho, Fleet, and Norouzi]{saharia2022imagen}
Chitwan Saharia, William Chan, Saurabh Saxena, Lala Li, Jay Whang, Emily~L. Denton, Seyed Kamyar~Seyed Ghasemipour, Burcu~Karagol Ayan, Seyedeh~Sara Mahdavi, Raphael~Gontijo Lopes, Tim Salimans, Jonathan Ho, David~J. Fleet, and Mohammad Norouzi.
\newblock Photorealistic text-to-image diffusion models with deep language understanding.
\newblock In \emph{NeurIPS}, 2022.

\bibitem[Schuhmann et~al.(2021)Schuhmann, Vencu, Beaumont, Kaczmarczyk, Mullis, Katta, Coombes, Jitsev, and Komatsuzaki]{schuhmann2021laion}
Christoph Schuhmann, Richard Vencu, Romain Beaumont, Robert Kaczmarczyk, Clayton Mullis, Aarush Katta, Theo Coombes, Jenia Jitsev, and Aran Komatsuzaki.
\newblock {LAION-400M}: Open dataset of {CLIP}-filtered 400 million image-text pairs.
\newblock \emph{arXiv:2111.02114}, 2021.

\bibitem[Shechtman and Irani(2007)]{shechtman2007matching}
Eli Shechtman and Michal Irani.
\newblock Matching local self-similarities across images and videos.
\newblock In \emph{CVPR}, 2007.

\bibitem[Sim{\'e}oni et~al.(2021)Sim{\'e}oni, Puy, Vo, Roburin, Gidaris, Bursuc, P{\'e}rez, Marlet, and Ponce]{simeoni2021localizing}
Oriane Sim{\'e}oni, Gilles Puy, Huy~V Vo, Simon Roburin, Spyros Gidaris, Andrei Bursuc, Patrick P{\'e}rez, Renaud Marlet, and Jean Ponce.
\newblock Localizing objects with self-supervised transformers and no labels.
\newblock \emph{arXiv:2109.14279}, 2021.

\bibitem[Song et~al.(2021)Song, Meng, and Ermon]{song2020denoising}
Jiaming Song, Chenlin Meng, and Stefano Ermon.
\newblock Denoising diffusion implicit models.
\newblock In \emph{ICLR}, 2021.

\bibitem[Su et~al.(2022)Su, Song, Meng, and Ermon]{su2022dual}
Xuan Su, Jiaming Song, Chenlin Meng, and Stefano Ermon.
\newblock Dual diffusion implicit bridges for image-to-image translation.
\newblock In \emph{ICLR}, 2022.

\bibitem[Tumanyan et~al.(2023)Tumanyan, Geyer, Bagon, and Dekel]{tumanyan2023plug}
Narek Tumanyan, Michal Geyer, Shai Bagon, and Tali Dekel.
\newblock Plug-and-play diffusion features for text-driven image-to-image translation.
\newblock In \emph{CVPR}, 2023.

\bibitem[Van Den~Oord et~al.(2017)Van Den~Oord, Vinyals, et~al.]{van2017neural}
Aaron Van Den~Oord, Oriol Vinyals, et~al.
\newblock Neural discrete representation learning.
\newblock In \emph{NeurIPS}, 2017.

\bibitem[Xu et~al.(2018)Xu, Zhang, Huang, Zhang, Gan, Huang, and He]{xu2018attngan}
Tao Xu, Pengchuan Zhang, Qiuyuan Huang, Han Zhang, Zhe Gan, Xiaolei Huang, and Xiaodong He.
\newblock {AttnGAN}: Fine-grained text to image generation with attentional generative adversarial networks.
\newblock In \emph{CVPR}, 2018.

\bibitem[Ye et~al.(2022)Ye, Xie, Chen, Xu, Yuan, Zhu, and Liao]{ye2022improving}
Shuquan Ye, Yujia Xie, Dongdong Chen, Yichong Xu, Lu Yuan, Chenguang Zhu, and Jing Liao.
\newblock Improving commonsense in vision-language models via knowledge graph riddles.
\newblock \emph{arXiv:2211.16504}, 2022.

\bibitem[Yu et~al.(2022)Yu, Xu, Koh, Luong, Baid, Wang, Vasudevan, Ku, Yang, Ayan, Hutchinson, Han, Parekh, Li, Zhang, Baldridge, and Wu]{Yu2022parti}
Jiahui Yu, Yuanzhong Xu, Jing~Yu Koh, Thang Luong, Gunjan Baid, Zirui Wang, Vijay Vasudevan, Alexander Ku, Yinfei Yang, Burcu~Karagol Ayan, Ben Hutchinson, Wei Han, Zarana Parekh, Xin Li, Han Zhang, Jason Baldridge, and Yonghui Wu.
\newblock Scaling autoregressive models for content-rich text-to-image generation.
\newblock \emph{arXiv:2206.10789}, 2022.

\bibitem[Zhang et~al.(2017)Zhang, Xu, Li, Zhang, Wang, Huang, and Metaxas]{zhang2017stackgan}
Han Zhang, Tao Xu, Hongsheng Li, Shaoting Zhang, Xiaogang Wang, Xiaolei Huang, and Dimitris~N Metaxas.
\newblock {StackGAN}: Text to photo-realistic image synthesis with stacked generative adversarial networks.
\newblock In \emph{ICCV}, 2017.

\bibitem[Zhang et~al.(2018)Zhang, Xu, Li, Zhang, Wang, Huang, and Metaxas]{zhang2018stackgan++}
Han Zhang, Tao Xu, Hongsheng Li, Shaoting Zhang, Xiaogang Wang, Xiaolei Huang, and Dimitris~N Metaxas.
\newblock {StackGAN++}: Realistic image synthesis with stacked generative adversarial networks.
\newblock \emph{IEEE TPAMI}, 41\penalty0 (8):\penalty0 1947--1962, 2018.

\bibitem[Zhang et~al.(2023)Zhang, Rao, and Agrawala]{zhang2023adding}
Lvmin Zhang, Anyi Rao, and Maneesh Agrawala.
\newblock Adding conditional control to text-to-image diffusion models.
\newblock In \emph{ICCV}, 2023.

\bibitem[Zhang et~al.(2021)Zhang, Tseng, Jiang, Yang, Lee, and Essa]{zhang2021text}
Tianhao Zhang, Hung-Yu Tseng, Lu Jiang, Weilong Yang, Honglak Lee, and Irfan Essa.
\newblock Text as neural operator: Image manipulation by text instruction.
\newblock In \emph{ACM MM}, pages 1893--1902, 2021.

\end{thebibliography}
}

\newpage
\appendix 

\section{Overview}
This supplementary material includes the following additional content:
\begin{compactitem}
\item Further implementation details in Section \ref{sup_implementation}.
\item Additional experimental results in Section \ref{sup_experiment}.
\item Additional visualization results in Section \ref{vis}, as shown in Figure \ref{figure3}, \ref{figure4} and \ref{figure5_2}.
\end{compactitem}

\section{Implementation Details}
\label{sup_implementation}
In our experiments in the main paper, we chose 8 anchor points whose attention values are the top-8 largest. The training time for each sample is approximately 4 minutes, with a batch size set to 1 using two A5000 GPUs. For the diffusion model, we set the guidance scale and strength as 7.5 and 0.75, respectively.

\section{Additional Experimental Results}
\label{sup_experiment}

\subsection{Comparison with existing methods}
In Figure \ref{figure_comp1} and \ref{figure_comp2} of this Appendix, we present additional results comparing our proposed method with state-of-the-art text-driven image editing baseline approaches, including Plug-and-Play \cite{tumanyan2023plug}, InstructPix2Pix \cite{brooks2023instructpix2pix}, Null-text inversion \cite{mokady2023null}, DiffEdit \cite{couairon2022diffedit} and MasaCtrl \cite{cao2023masactrl}. 

We observe that the comparison results are consistent with the user studies reported in the main paper: our method generates images of higher quality that adhere to the text prompts while preserving the background content. The baseline methods appear to fall short in terms of generation quality and controllability by the prompt. For example, Plug-and-Play and InstructPix2Pix tend to modify the style of the entire image. The Null-text inversion method appears to have difficulty generating objects that align with the text prompts. Meanwhile, techniques like DiffEdit and MasaCtrl yield results that are of lower quality or poorer alignment with the text prompt.

\subsection{Additional Ablation Study}
In this section, we present three additional ablation studies employing qualitative metrics to complement the study based on user ratings reported in the main paper. Two metric scores are used: (1) the CLIP~~\cite{radford2021learning} text-to-image similarity score $S_{t2i}$, which evaluates the cosine similarity between the given prompt and the edited image; (2) the CLIP image-to-image similarity score $S_{i2i}$, which represents the cosine similarity between the source image and the edited image. We adopt the CLIP model initialized from ViT-B/16 weights to calculate the similarity scores. Results highlighted in \colorbox{blue_new!20}{blue} denote the default settings used in our main experiments. 

\begin{table}[!t]
\centering
\begin{tabular}{cp{2cm}<{\centering}p{2cm}<{\centering}}
\toprule
Loss Component & S$_{t2i}$ $\uparrow$ & S$_{i2i}$ $\uparrow$ \\ \midrule
$\mathcal{L}_{Clip}$ &  0.301 & 0.801 \\
$\mathcal{L}_{Clip}+L_{Str}$  & 0.294 &     0.806   \\ 
\rowcolor{blue_new!15} $\mathcal{L}_{Clip}+L_{Str}+L_{Dir}$ & 0.301 & 0.805   \\ \bottomrule
\end{tabular}
\vspace{-2mm}
\caption{Ablation study on adopting different loss components.}
\label{tab:1}
\end{table}

\begin{table}[!t]
\centering
\begin{tabular}{cp{2cm}<{\centering}p{2cm}<{\centering}}
\toprule
\# of region proposals & S$_{t2i}$ $\uparrow$ & S$_{i2i}$ $\uparrow$ \\ \midrule
1  & 0.231 & 0.915  \\
3  & 0.273 &  0.837 \\
5  & 0.295 & 0.809 \\
\rowcolor{blue_new!15} 7  & 0.300 &   0.805     \\ 
9 & 0.301 & 0.802     \\ \bottomrule
\end{tabular}
\vspace{-2mm}
\caption{Ablation study on the number of region proposals.}
\label{tab:2}
\end{table}

\begin{table}[!t]
\centering
\begin{tabular}{p{3cm}<{\centering}p{2cm}<{\centering}p{2cm}<{\centering}p{2cm}<{\centering}}
\toprule
\# of anchor points & S$_{t2i}$ $\uparrow$ & S$_{i2i}$ $\uparrow$ \\ \midrule
1  & 0.275 & 0.824   \\
4  & 0.296 & 0.805  \\
6  &0.301 &  0.802 \\
\rowcolor{blue_new!15} 8  & 0.300 &   0.805    \\ 
10 & 0.298 &  0.803    \\ \bottomrule
\end{tabular}
\vspace{-2mm}
\caption{Ablation study on the number of anchor points.}
\label{tab:3}
\end{table}

\paragraph{Effect of different loss components.} To evaluate the influences of the introduced loss components (\ie, $\mathcal{L}_{Clip}$, $\mathcal{L}_{Str}$ and $\mathcal{L}_{Dir}$) in our training loss, we report the metric scores of adopting different loss components in Table \ref{tab:1}. 
The results show that using all losses yields the best overall score (\ie, average of $S_{t2i}$ and $S_{i2i}$), thereby verifying their contributions.

\paragraph{Effect of the number of region proposals.} In Table \ref{tab:2}, we evaluate the performance of using different numbers of region proposals. 
The results indicate that it's likely to achieve the performance bottleneck with a larger region proposal (\eg, 9 region proposals), and our method achieves reasonable quality (\eg, a superb balance between $S_{t2i}$ and $S_{i2i}$) even using 7 region proposals.

\paragraph{Effect of the number of anchor points.} To analyze the influence of the number of anchor points, we display the ablation study in Table \ref{tab:3}. It can be observed that when the number of anchor points increases from 1 to 8, the text-to-image similarity score $S_{t2i}$ consistently increases. However, when the number is too large (\eg, 10), the $S_{t2i}$ decreases, it may be caused by the larger possibility of choosing noisy anchor points, which may be located in the background area, hurting the performance of editing.

\begin{figure}[t]
\begin{minipage}{0.24\linewidth}
\captionof*{figure}{\small{\textcolor{black}{Input Image}}}
\vspace{-3mm}
\includegraphics[width=\linewidth,height=\linewidth]{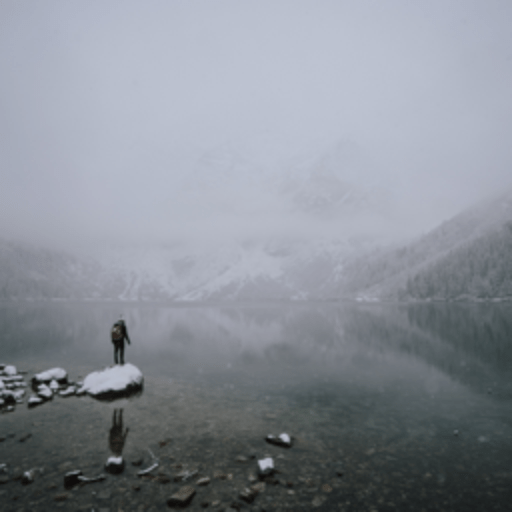}
\end{minipage}\hfill
\begin{minipage}{0.24\linewidth}
\captionof*{figure}{\small{\textcolor{black}{Result}}}
\vspace{-3mm}
\includegraphics[width=\linewidth,height=\linewidth]{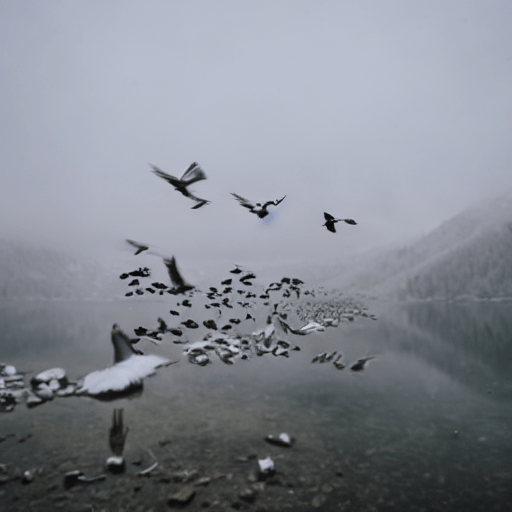}
\end{minipage}\hfill
\begin{minipage}{0.24\linewidth}
\captionof*{figure}{\small{\textcolor{black}{Input Image}}}
\vspace{-3mm}
\includegraphics[width=\linewidth,height=\linewidth]{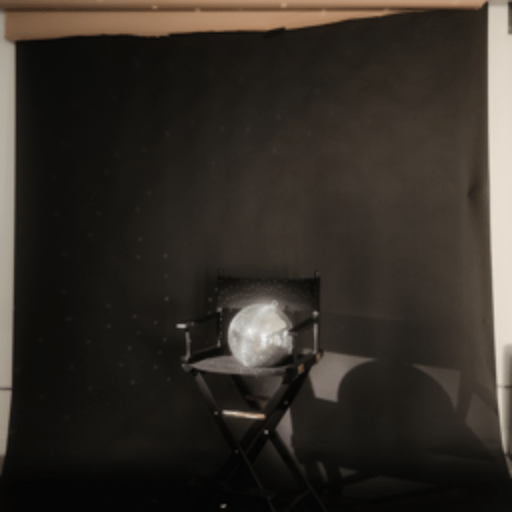}
\end{minipage}\hfill
\begin{minipage}{0.24\linewidth}
\captionof*{figure}{\small{\textcolor{black}{Result}}}
\vspace{-3mm}
\includegraphics[width=\linewidth,height=\linewidth]{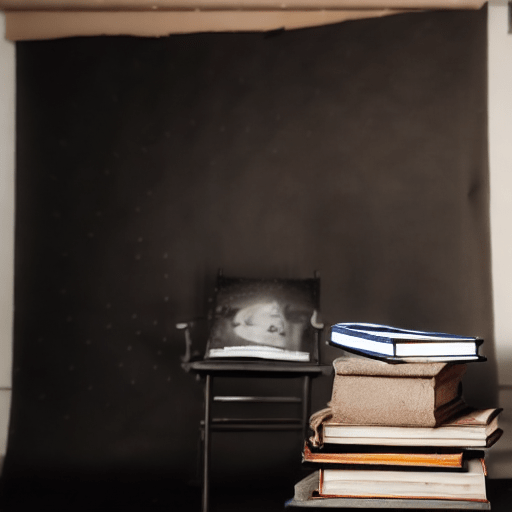}
\end{minipage}
\vspace{2mm}
\\
\vspace{1mm}
\hspace{1mm} \small{\textcolor{black}{\textit{a flock of pigeons takes flight} \hspace{1mm} \textit{several books rest on the chair}}}
\vspace{1mm}
\\
\begin{minipage}{0.24\linewidth}
\includegraphics[width=\linewidth,height=\linewidth]{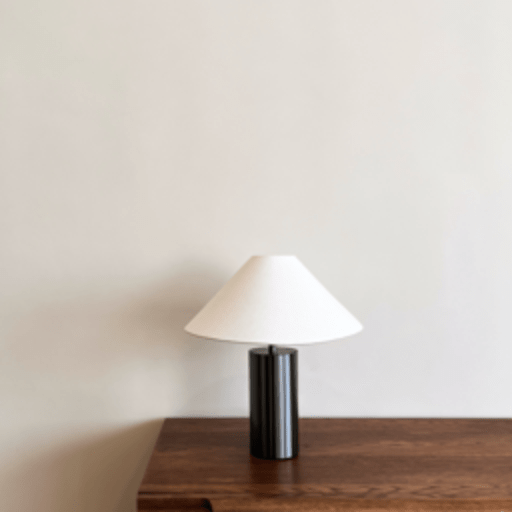}
\end{minipage}\hfill
\begin{minipage}{0.24\linewidth}
\includegraphics[width=\linewidth,height=\linewidth]{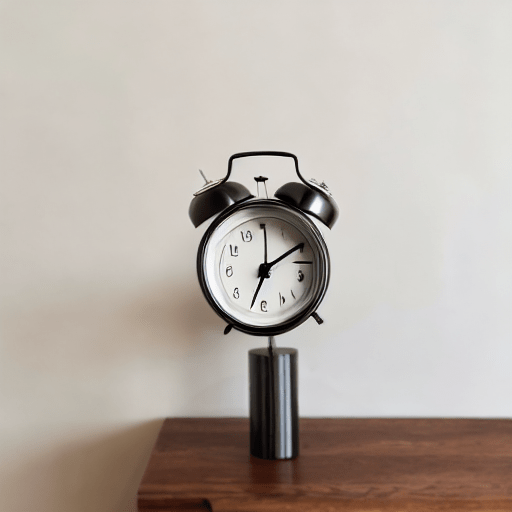}
\end{minipage}\hfill
\begin{minipage}{0.24\linewidth}
\includegraphics[width=\linewidth,height=\linewidth]{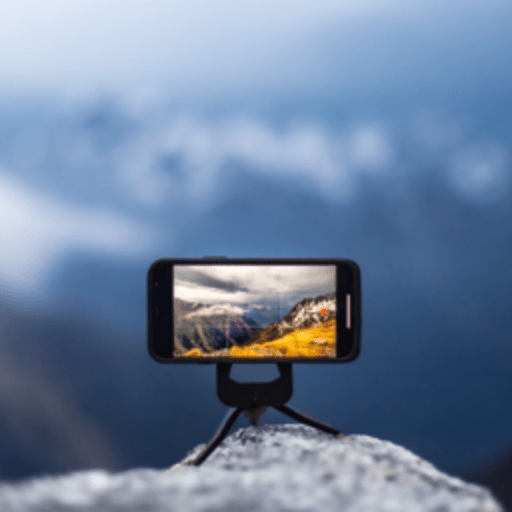}
\end{minipage}\hfill
\begin{minipage}{0.24\linewidth}
\includegraphics[width=\linewidth,height=\linewidth]{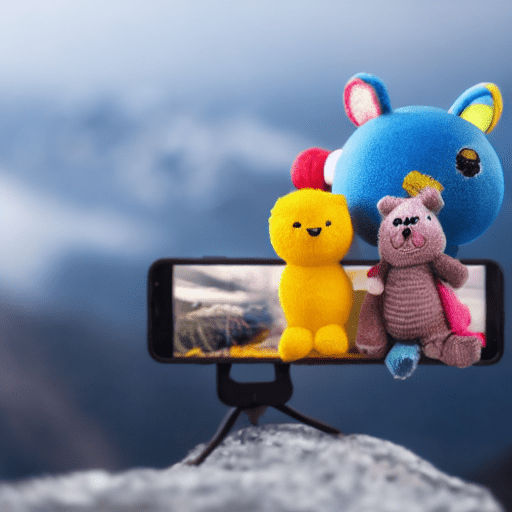}
\end{minipage}
\vspace{2mm}
\\
\hspace{-2mm} \small{\textcolor{black}{\textit{an alarm clock with functional simplicity} \hspace{1mm} \textit{an assortment of toys}}}
\vspace{-3mm}
\\
\captionof{figure}{\textbf{Failure cases.} We show more failure cases generated by our method.}
\label{figure:failure2}
\end{figure}

\subsection{Analysis}
\paragraph{More failure cases.} 
In Figure~\ref{figure:failure2}, we display more failure examples generated by our method. We can see that the proposed method may generate unsatisfactory results when the anchor points are sampled from the background regions. 

\section{Visualizations}
\label{vis}
We provide more visualization results of text-driven image editing generated by our method in Figure \ref{figure3}, \ref{figure4} and \ref{figure5_2}. Thanks to the flexibility of the learned bounding box guidance, our method is capable of handling a wide range of prompts, it can generate satisfactory results while preserving the original appearance of the background.

\begin{figure*}[htp]
\begin{minipage}{0.139\linewidth}
\captionof*{figure}{\small{\textcolor{black}{Input Image}}}
\vspace{-3mm}
\includegraphics[width=\linewidth,height=\linewidth]{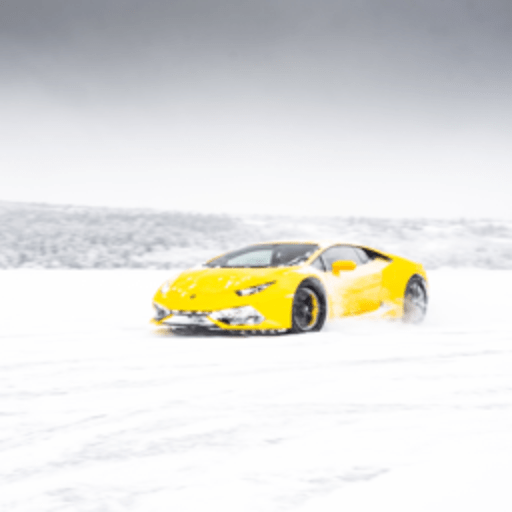}
\end{minipage}\hfill
\begin{minipage}{0.139\textwidth}
\captionof*{figure}{\small{\textcolor{black}{Plug-and-Play}}}
\vspace{-3mm}
\includegraphics[width=\linewidth,height=\linewidth]{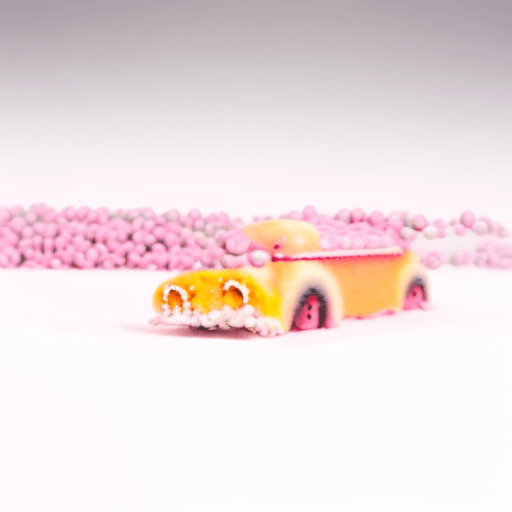}   
\end{minipage}\hfill
\begin{minipage}{0.139\textwidth}
\captionof*{figure}{\small{\textcolor{black}{InstructPix2Pix}}}
\vspace{-3mm}
\includegraphics[width=\linewidth,height=\linewidth]{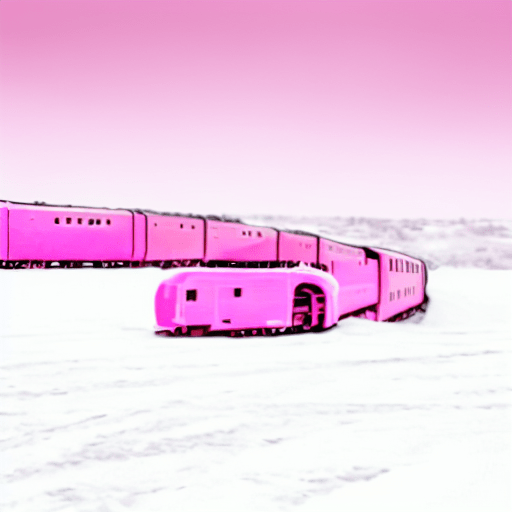}   
\end{minipage}\hfill
\begin{minipage}{0.139\textwidth}
\captionof*{figure}{\small{\textcolor{black}{Null-text}}}
\vspace{-3mm}
\includegraphics[width=\linewidth,height=\linewidth]{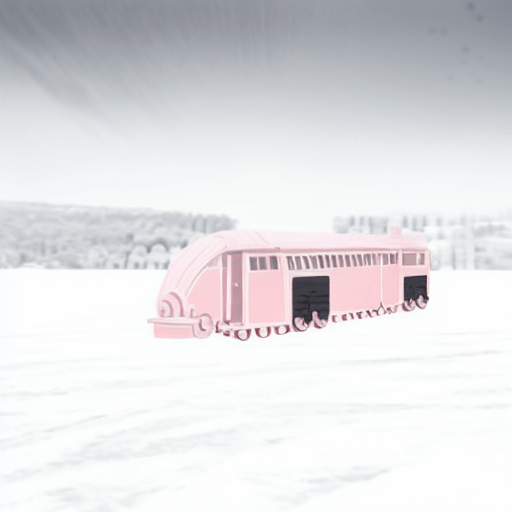} 
\end{minipage}\hfill
\begin{minipage}{0.139\textwidth}
\captionof*{figure}{\small{\textcolor{black}{DiffEdit}}}
\vspace{-3mm}
\includegraphics[width=\linewidth,height=\linewidth]{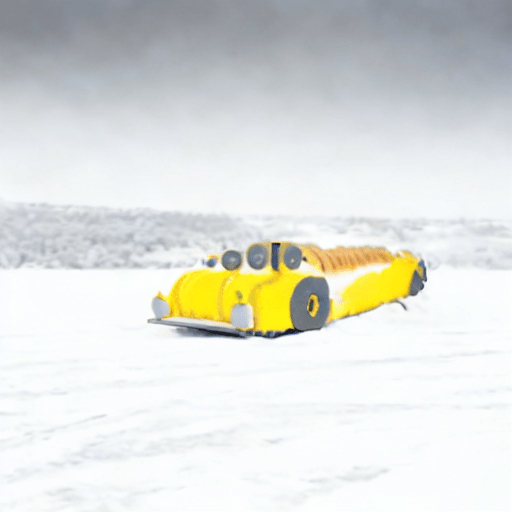} 
\end{minipage}\hfill
\begin{minipage}{0.139\textwidth}
\captionof*{figure}{\small{\textcolor{black}{MasaCtrl}}}
\vspace{-3mm}
\includegraphics[width=\linewidth,height=\linewidth]{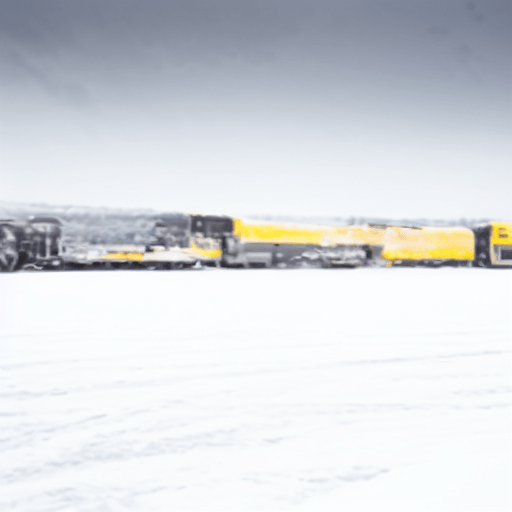} 
\end{minipage}\hfill
\begin{minipage}{0.139\textwidth}
\captionof*{figure}{\small{\textcolor{black}{Ours}}}
\vspace{-3mm}
\includegraphics[width=\linewidth,height=\linewidth]{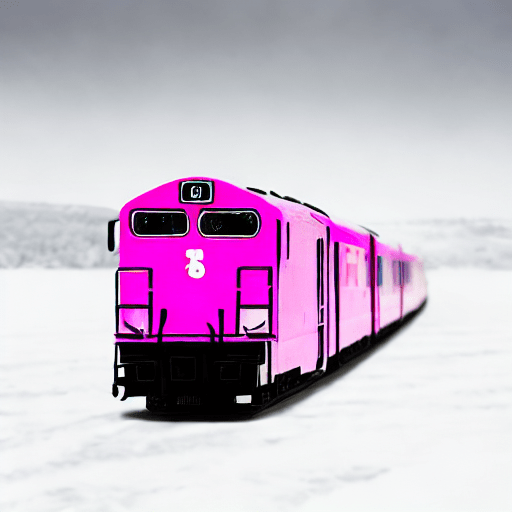} 
\end{minipage}
\vspace{1mm}
\\
\hspace{-2mm} \small{\textcolor{black}{\textsl{Editing Text:}}}   \hspace{55mm} \textcolor{black}{\textit{a high quality photo of a pink train}}
\vspace{4mm}
\\
\begin{minipage}{0.139\linewidth}
\vspace{-3mm}
\includegraphics[width=\linewidth,height=\linewidth]{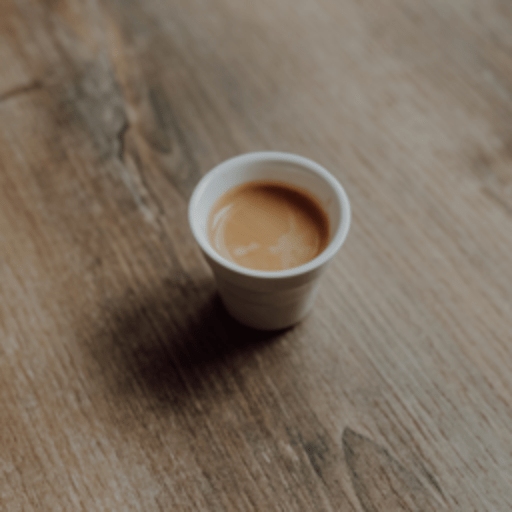}
\end{minipage}\hfill
\begin{minipage}{0.139\linewidth}
\vspace{-3mm}
\includegraphics[width=\linewidth,height=\linewidth]{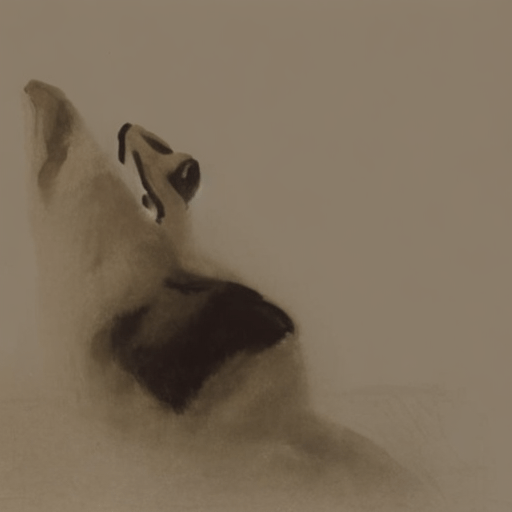}
\end{minipage}\hfill
\begin{minipage}{0.139\textwidth}
\vspace{-3mm}
\includegraphics[width=\linewidth,height=\linewidth]{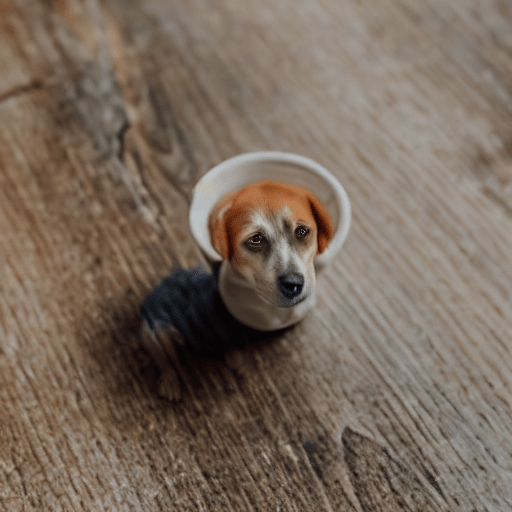}   
\end{minipage}\hfill
\begin{minipage}{0.139\textwidth}
\vspace{-3mm}
\includegraphics[width=\linewidth,height=\linewidth]{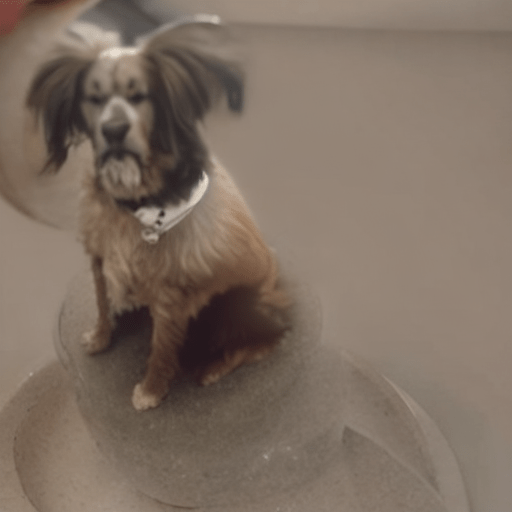} 
\end{minipage}\hfill
\begin{minipage}{0.139\textwidth}
\vspace{-3mm}
\includegraphics[width=\linewidth,height=\linewidth]{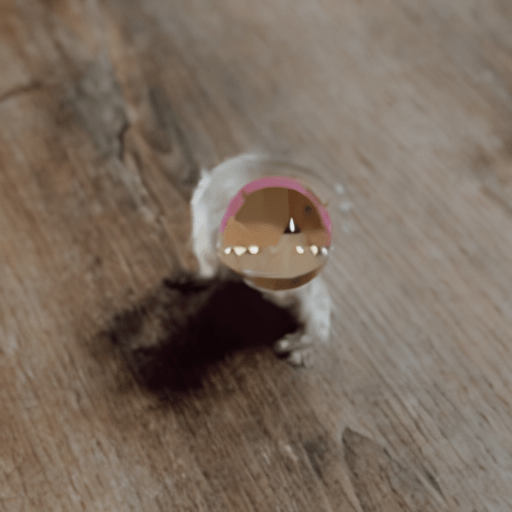} 
\end{minipage}\hfill
\begin{minipage}{0.139\textwidth}
\vspace{-3mm}
\includegraphics[width=\linewidth,height=\linewidth]{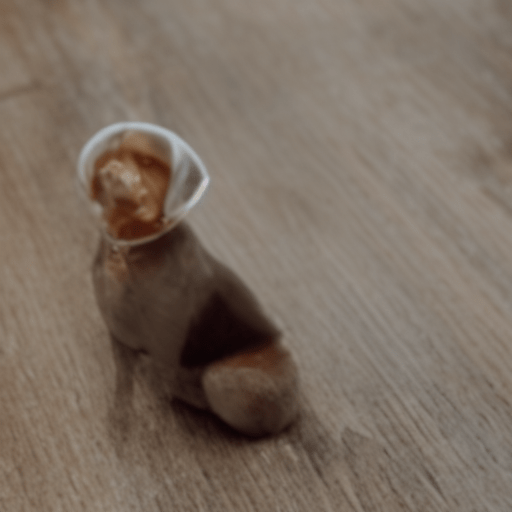} 
\end{minipage}\hfill
\begin{minipage}{0.139\textwidth}
\vspace{-3mm}
\includegraphics[width=\linewidth,height=\linewidth]{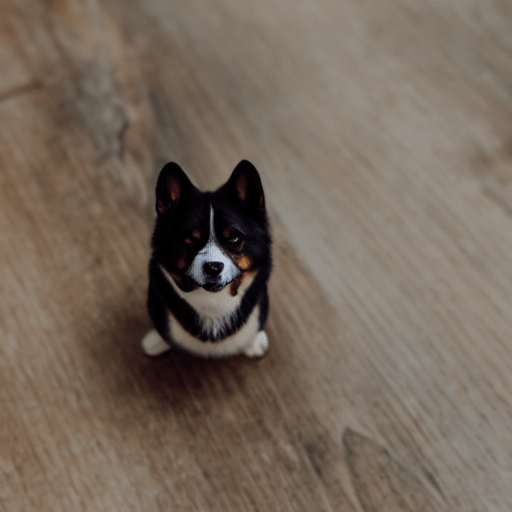} 
\end{minipage}
\vspace{1mm}
\\
\hspace{-2mm} \small{\textcolor{black}{\textsl{Editing Text:}}}   \hspace{61mm} \textcolor{black}{\textit{a dog in a seated position}}
\vspace{4mm}
\\
\begin{minipage}{0.139\linewidth}
\vspace{-3mm}
\includegraphics[width=\linewidth,height=\linewidth]{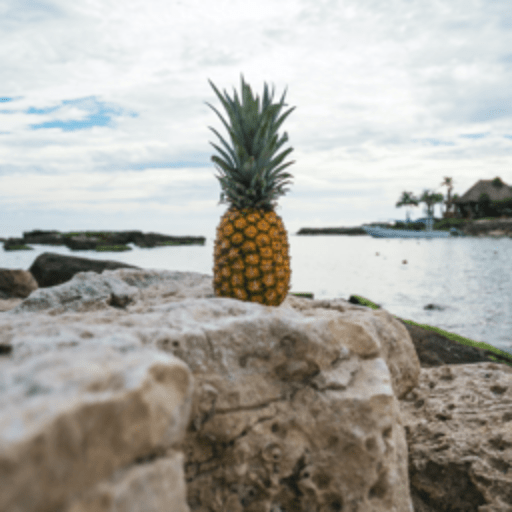}
\end{minipage}\hfill
\begin{minipage}{0.139\linewidth}
\vspace{-3mm}
\includegraphics[width=\linewidth,height=\linewidth]{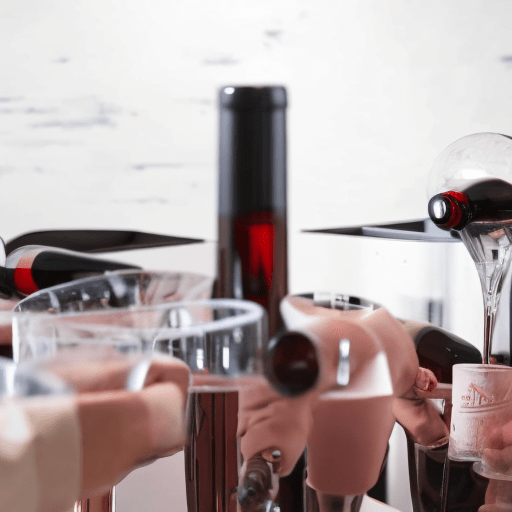}
\end{minipage}\hfill
\begin{minipage}{0.139\textwidth}
\vspace{-3mm}
\includegraphics[width=\linewidth,height=\linewidth]{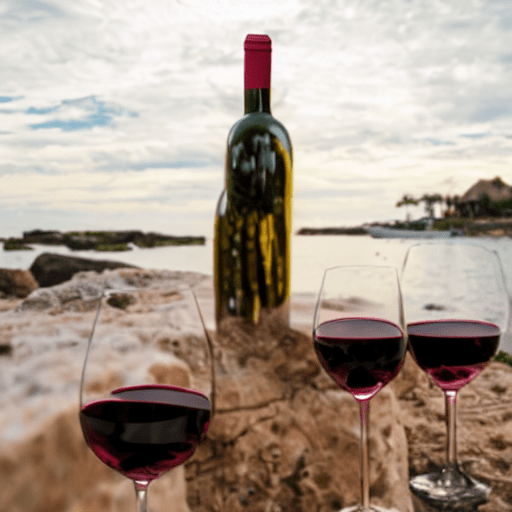}   
\end{minipage}\hfill
\begin{minipage}{0.139\textwidth}
\vspace{-3mm}
\includegraphics[width=\linewidth,height=\linewidth]{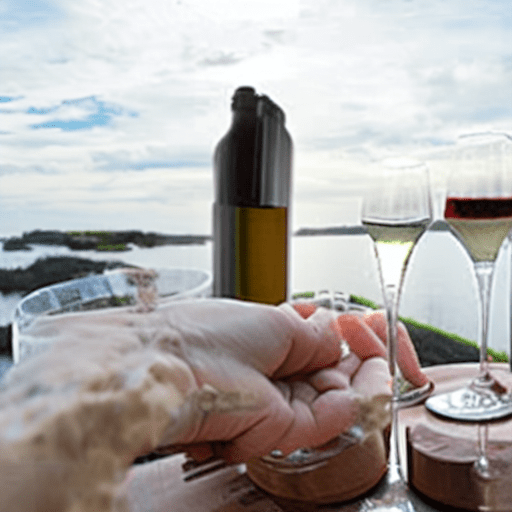} 
\end{minipage}\hfill
\begin{minipage}{0.139\textwidth}
\vspace{-3mm}
\includegraphics[width=\linewidth,height=\linewidth]{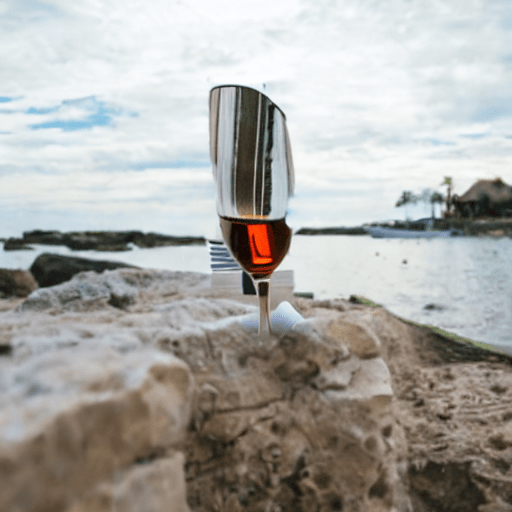} 
\end{minipage}\hfill
\begin{minipage}{0.139\textwidth}
\vspace{-3mm}
\includegraphics[width=\linewidth,height=\linewidth]{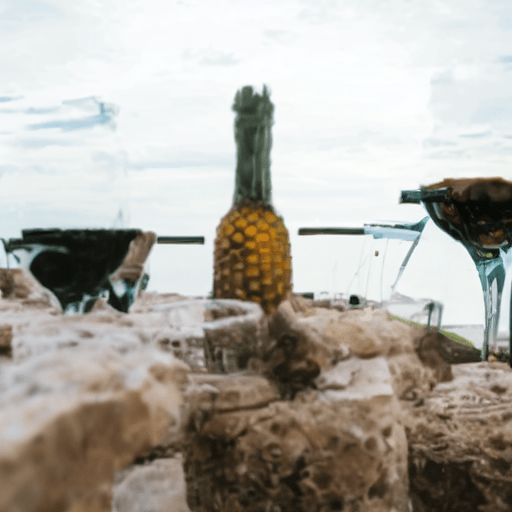} 
\end{minipage}\hfill
\begin{minipage}{0.139\textwidth}
\vspace{-3mm}
\includegraphics[width=\linewidth,height=\linewidth]{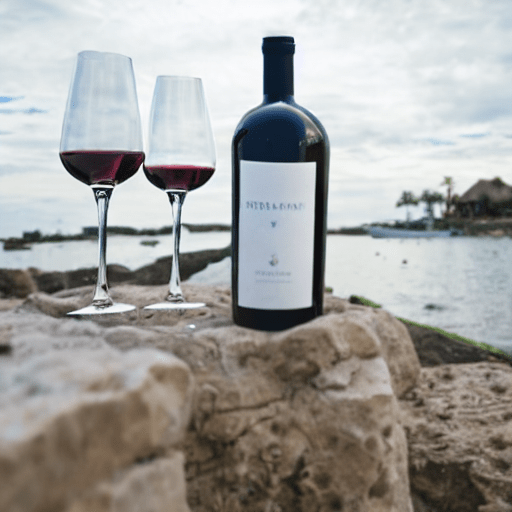} 
\end{minipage}
\vspace{1mm}

\hspace{-1mm} \small{\textcolor{black}{\textsl{Editing Text:}}}   \hspace{53mm} \textcolor{black}{\textit{a bottle of wine and several wine cups}}
\vspace{4mm}
\\
\begin{minipage}{0.139\linewidth}
\vspace{-3mm}
\includegraphics[width=\linewidth,height=\linewidth]{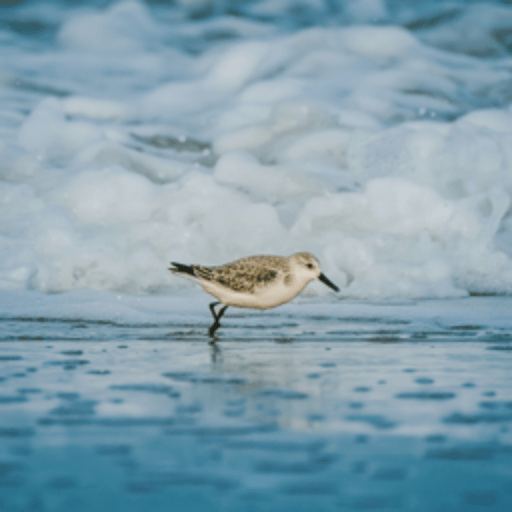}
\end{minipage}\hfill
\begin{minipage}{0.139\linewidth}
\vspace{-3mm}
\includegraphics[width=\linewidth,height=\linewidth]{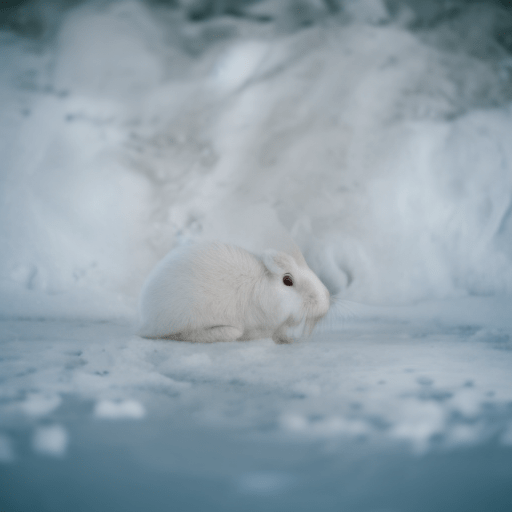}
\end{minipage}\hfill
\begin{minipage}{0.139\textwidth}
\vspace{-3mm}
\includegraphics[width=\linewidth,height=\linewidth]{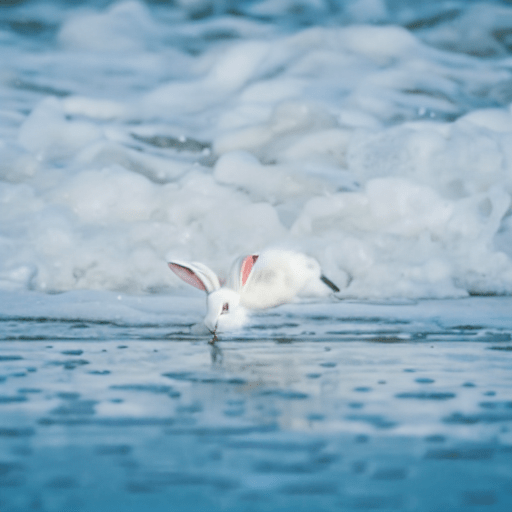}   
\end{minipage}\hfill
\begin{minipage}{0.139\textwidth}
\vspace{-3mm}
\includegraphics[width=\linewidth,height=\linewidth]{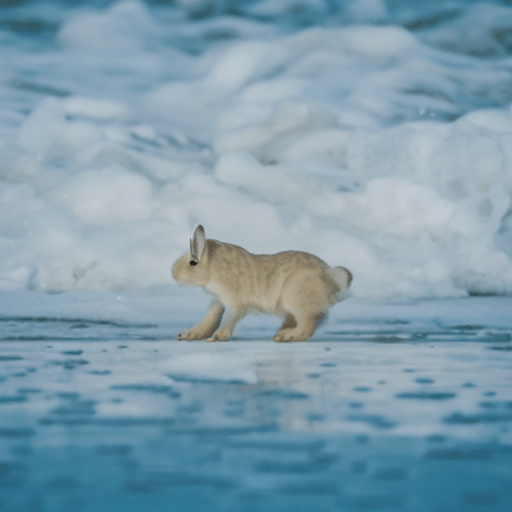} 
\end{minipage}\hfill
\begin{minipage}{0.139\textwidth}
\vspace{-3mm}
\includegraphics[width=\linewidth,height=\linewidth]{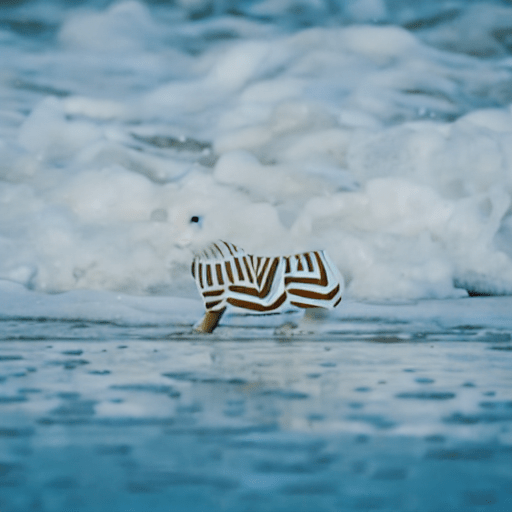} 
\end{minipage}\hfill
\begin{minipage}{0.139\textwidth}
\vspace{-3mm}
\includegraphics[width=\linewidth,height=\linewidth]{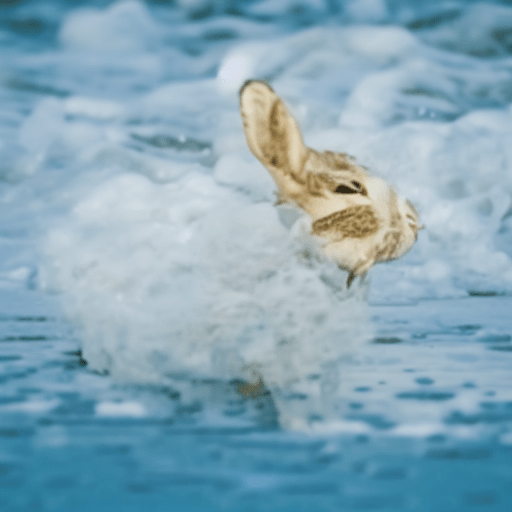} 
\end{minipage}\hfill
\begin{minipage}{0.139\textwidth}
\vspace{-3mm}
\includegraphics[width=\linewidth,height=\linewidth]{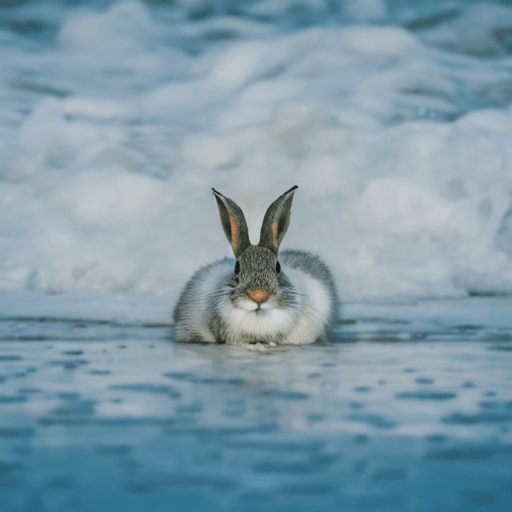} 
\end{minipage}
\vspace{1mm}
\\
\hspace{-2mm} \small{\textcolor{black}{\textsl{Editing Text:}}}   \hspace{62mm} \textcolor{black}{\textit{a rabbit with white fur}}
\vspace{4mm}
\\
\begin{minipage}{0.139\linewidth}
\vspace{-3mm}
\includegraphics[width=\linewidth,height=\linewidth]{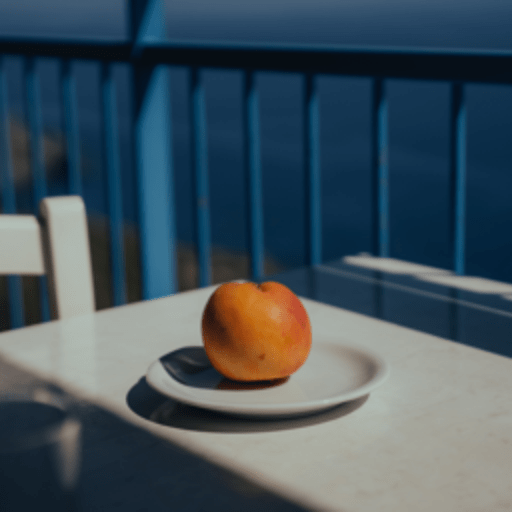}
\end{minipage}\hfill
\begin{minipage}{0.139\linewidth}
\vspace{-3mm}
\includegraphics[width=\linewidth,height=\linewidth]{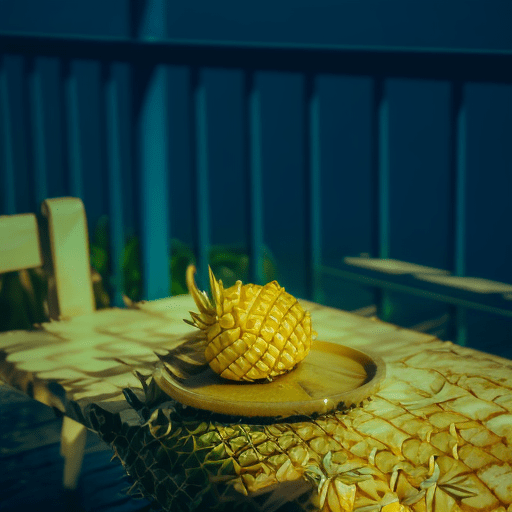}
\end{minipage}\hfill
\begin{minipage}{0.139\textwidth}
\vspace{-3mm}
\includegraphics[width=\linewidth,height=\linewidth]{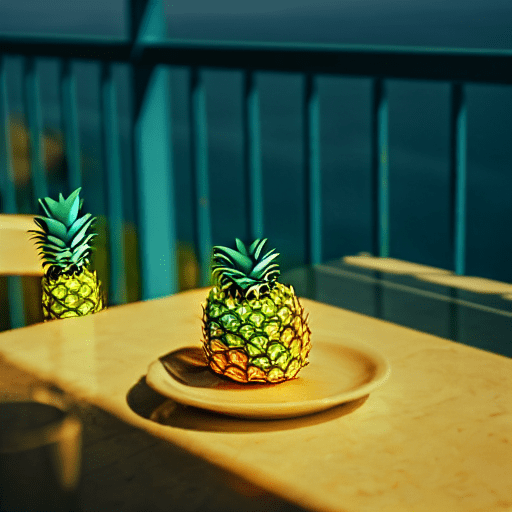}   
\end{minipage}\hfill
\begin{minipage}{0.139\textwidth}
\vspace{-3mm}
\includegraphics[width=\linewidth,height=\linewidth]{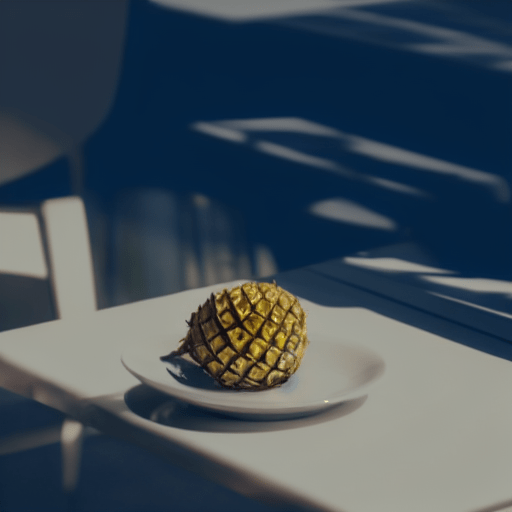} 
\end{minipage}\hfill
\begin{minipage}{0.139\textwidth}
\vspace{-3mm}
\includegraphics[width=\linewidth,height=\linewidth]{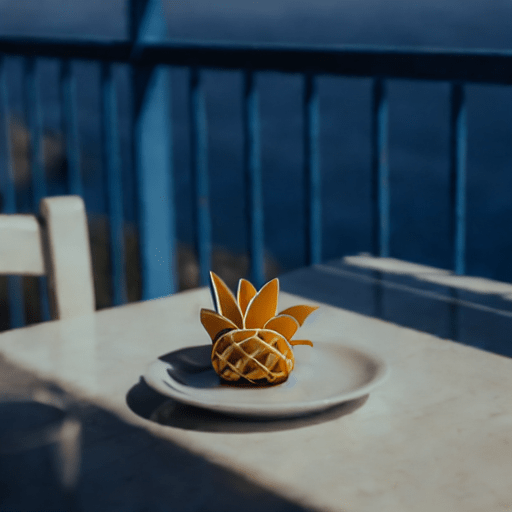} 
\end{minipage}\hfill
\begin{minipage}{0.139\textwidth}
\vspace{-3mm}
\includegraphics[width=\linewidth,height=\linewidth]{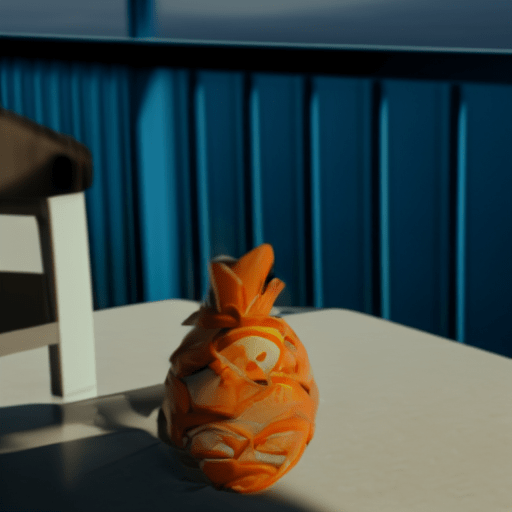} 
\end{minipage}\hfill
\begin{minipage}{0.139\textwidth}
\vspace{-3mm}
\includegraphics[width=\linewidth,height=\linewidth]{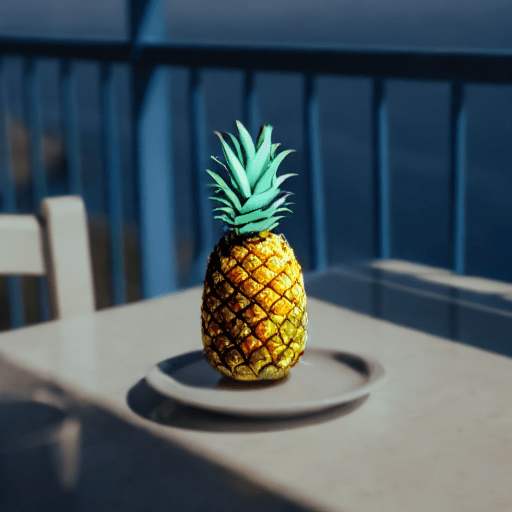} 
\end{minipage}
\vspace{1mm}
\\
\hspace{-2mm} \small{\textsl{Editing Text:}}   \hspace{61mm} \textcolor{black}{\textit{a pineapple on the plate}}
\vspace{4mm}
\\
\begin{minipage}{0.139\linewidth}
\vspace{-3mm}
\includegraphics[width=\linewidth,height=\linewidth]{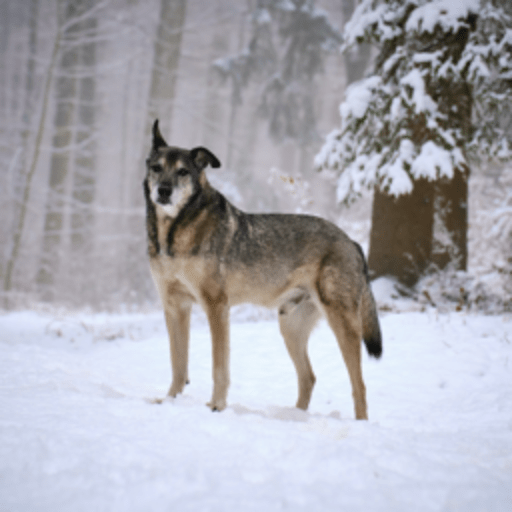}
\end{minipage}\hfill
\begin{minipage}{0.139\linewidth}
\vspace{-3mm}
\includegraphics[width=\linewidth,height=\linewidth]{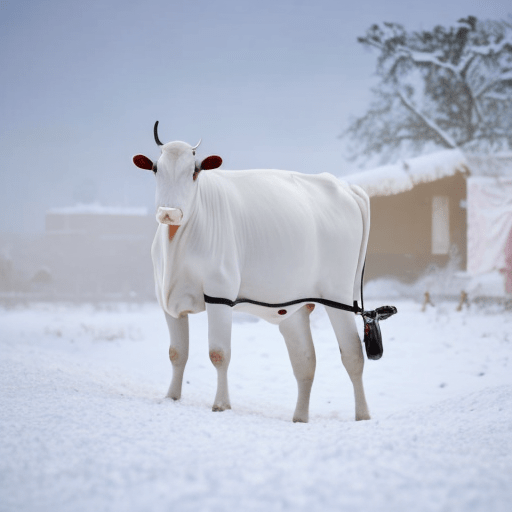}
\end{minipage}\hfill
\begin{minipage}{0.139\textwidth}
\vspace{-3mm}
\includegraphics[width=\linewidth,height=\linewidth]{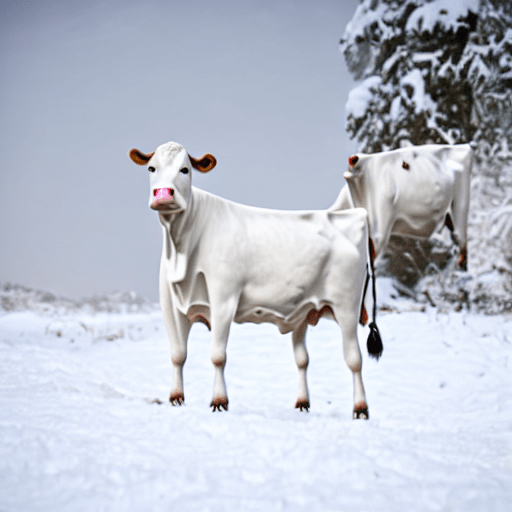}   
\end{minipage}\hfill
\begin{minipage}{0.139\textwidth}
\vspace{-3mm}
\includegraphics[width=\linewidth,height=\linewidth]{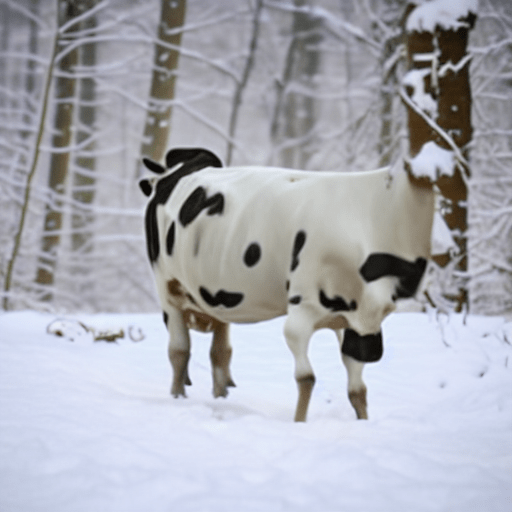} 
\end{minipage}\hfill
\begin{minipage}{0.139\textwidth}
\vspace{-3mm}
\includegraphics[width=\linewidth,height=\linewidth]{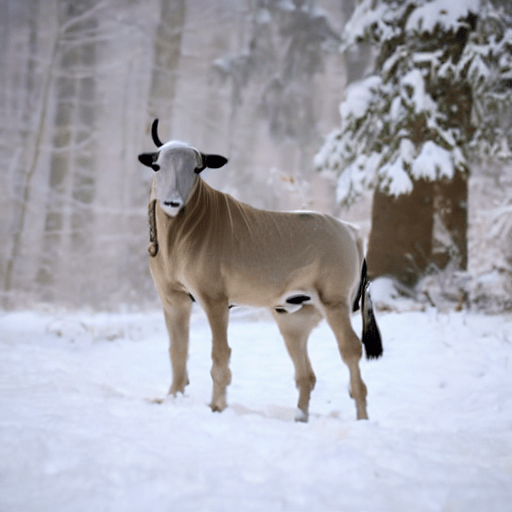} 
\end{minipage}\hfill
\begin{minipage}{0.139\textwidth}
\vspace{-3mm}
\includegraphics[width=\linewidth,height=\linewidth]{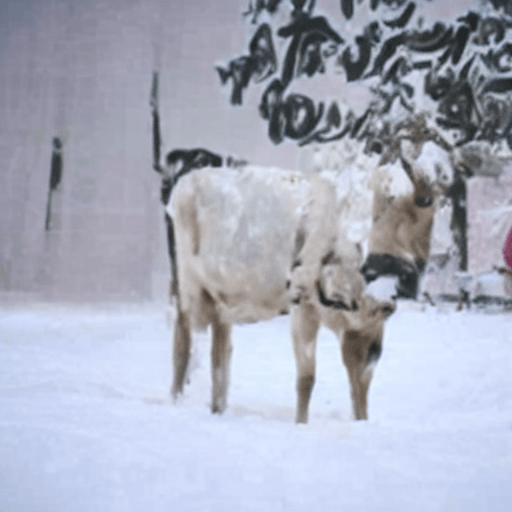} 
\end{minipage}\hfill
\begin{minipage}{0.139\textwidth}
\vspace{-3mm}
\includegraphics[width=\linewidth,height=\linewidth]{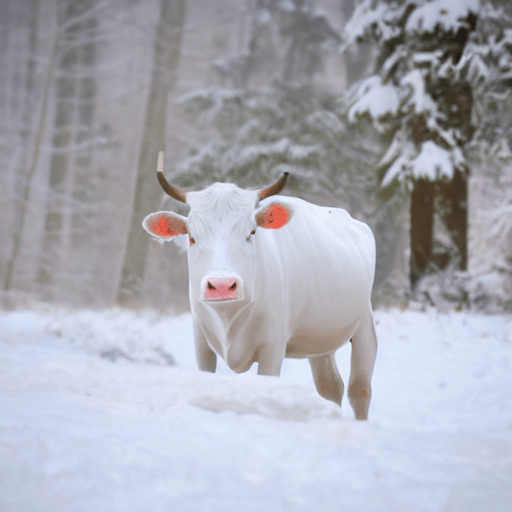} 
\end{minipage}
\vspace{1mm}
\\
\hspace{-2mm} \small{\textcolor{black}{\textsl{Editing Text:}}}   \hspace{55mm} \textcolor{black}{\textit{a white cow standing in the snow}}
\vspace{-1mm}
\captionof{figure}{\textbf{Comparison with existing methods.} We compare our method with existing text-driven image editing methods. From left to right: Input image, Plug-and-Play \cite{tumanyan2023plug}, InstructPix2Pix \cite{brooks2023instructpix2pix}, Null-text \cite{mokady2023null}, DiffEdit \cite{couairon2022diffedit}, MasaCtrl \cite{cao2023masactrl}, and ours.
\vspace{1.5mm}
}
\label{figure_comp1}
\end{figure*}

\clearpage

\begin{figure*}[htp]
\begin{minipage}{0.139\linewidth}
\captionof*{figure}{\small{\textcolor{black}{Input Image}}}
\vspace{-3mm}
\includegraphics[width=\linewidth,height=\linewidth]{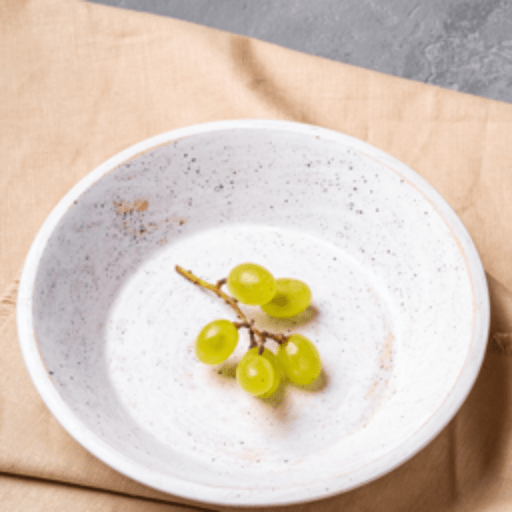}
\end{minipage}\hfill
\begin{minipage}{0.139\textwidth}
\captionof*{figure}{\small{\textcolor{black}{Plug-and-Play}}}
\vspace{-3mm}
\includegraphics[width=\linewidth,height=\linewidth]{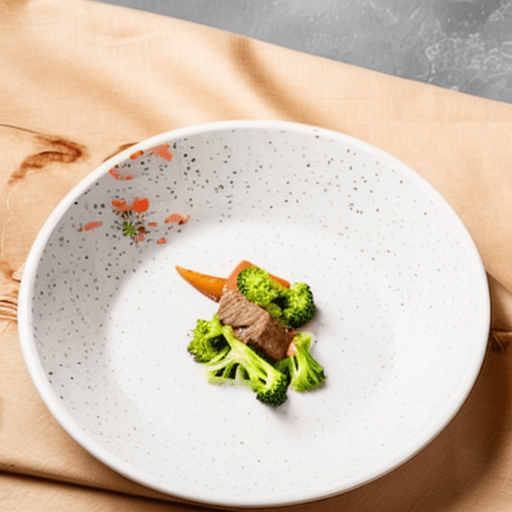}   
\end{minipage}\hfill
\begin{minipage}{0.139\textwidth}
\captionof*{figure}{\small{\textcolor{black}{InstructPix2Pix}}}
\vspace{-3mm}
\includegraphics[width=\linewidth,height=\linewidth]{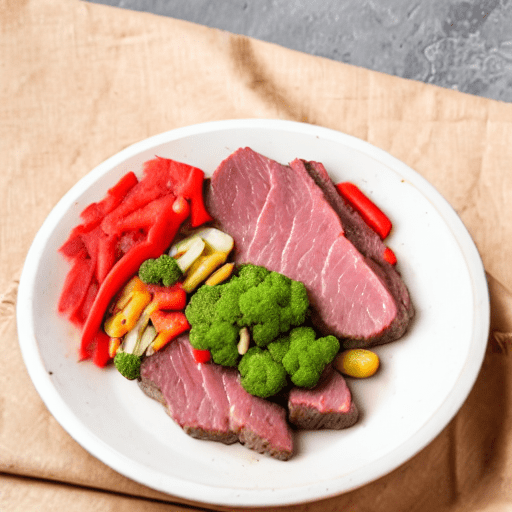}   
\end{minipage}\hfill
\begin{minipage}{0.139\textwidth}
\captionof*{figure}{\small{\textcolor{black}{Null-text}}}
\vspace{-3mm}
\includegraphics[width=\linewidth,height=\linewidth]{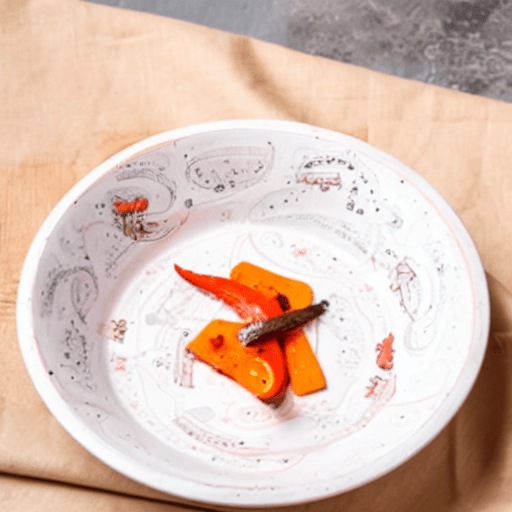} 
\end{minipage}\hfill
\begin{minipage}{0.139\textwidth}
\captionof*{figure}{\small{\textcolor{black}{DiffEdit}}}
\vspace{-3mm}
\includegraphics[width=\linewidth,height=\linewidth]{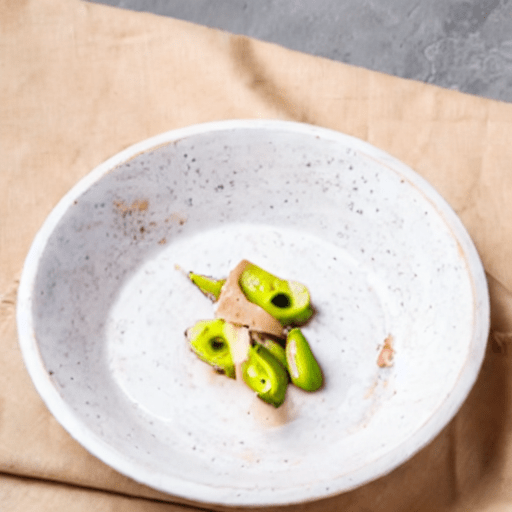} 
\end{minipage}\hfill
\begin{minipage}{0.139\textwidth}
\captionof*{figure}{\small{\textcolor{black}{MasaCtrl}}}
\vspace{-3mm}
\includegraphics[width=\linewidth,height=\linewidth]{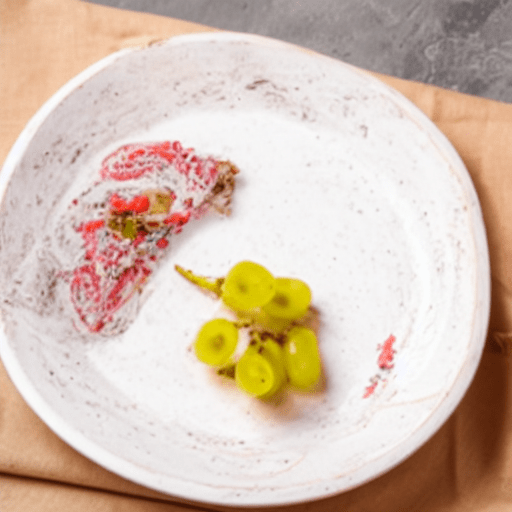} 
\end{minipage}\hfill
\begin{minipage}{0.139\textwidth}
\captionof*{figure}{\small{\textcolor{black}{Ours}}}
\vspace{-3mm}
\includegraphics[width=\linewidth,height=\linewidth]{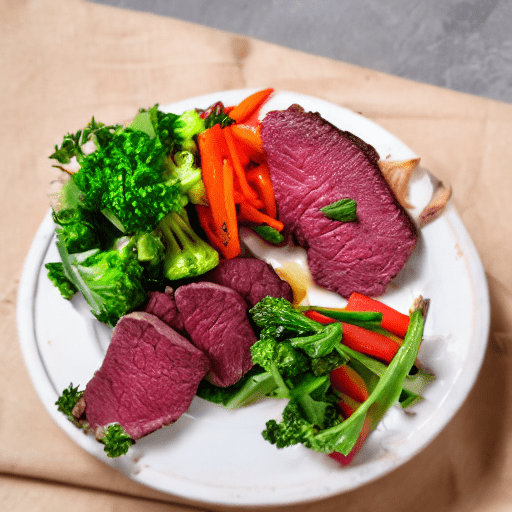} 
\end{minipage}
\vspace{1mm}
\\
\hspace{-2mm} \small{\textcolor{black}{\textsl{Editing Text:}}}   \hspace{56mm} \textcolor{black}{\textit{a plate of beefsteak and vegetable}}
\vspace{4mm}
\\
\begin{minipage}{0.139\linewidth}
\vspace{-3mm}
\includegraphics[width=\linewidth,height=\linewidth]{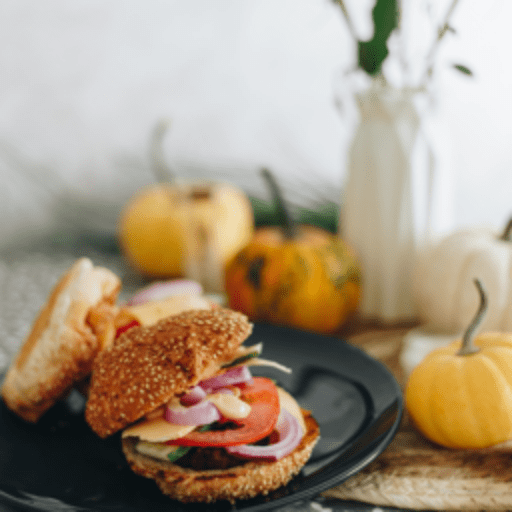}
\end{minipage}\hfill
\begin{minipage}{0.139\linewidth}
\vspace{-3mm}
\includegraphics[width=\linewidth,height=\linewidth]{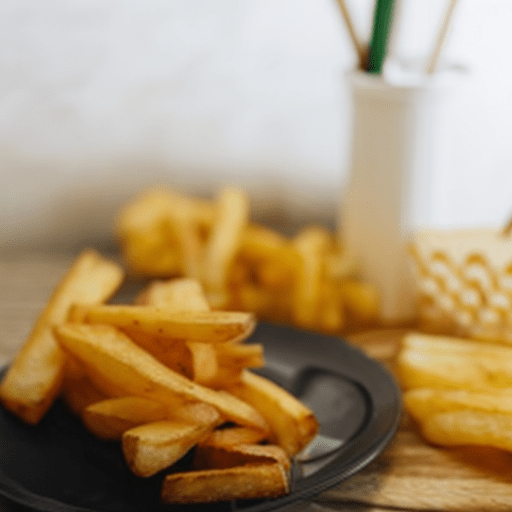}
\end{minipage}\hfill
\begin{minipage}{0.139\textwidth}
\vspace{-3mm}
\includegraphics[width=\linewidth,height=\linewidth]{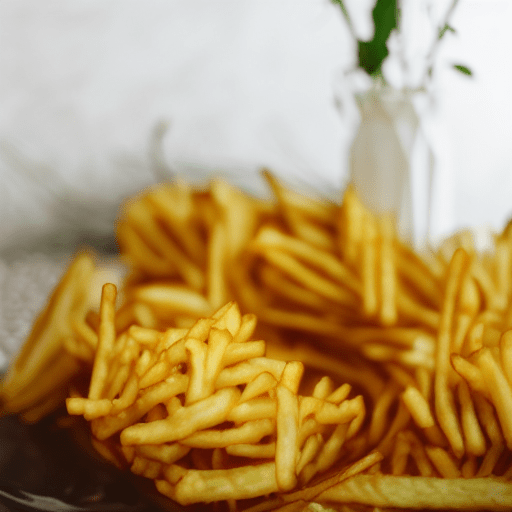}   
\end{minipage}\hfill
\begin{minipage}{0.139\textwidth}
\vspace{-3mm}
\includegraphics[width=\linewidth,height=\linewidth]{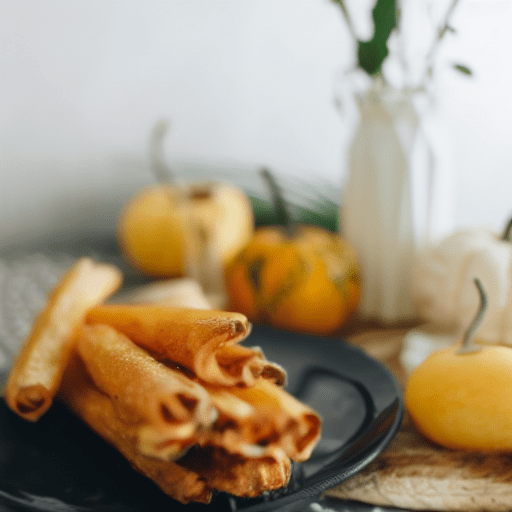} 
\end{minipage}\hfill
\begin{minipage}{0.139\textwidth}
\vspace{-3mm}
\includegraphics[width=\linewidth,height=\linewidth]{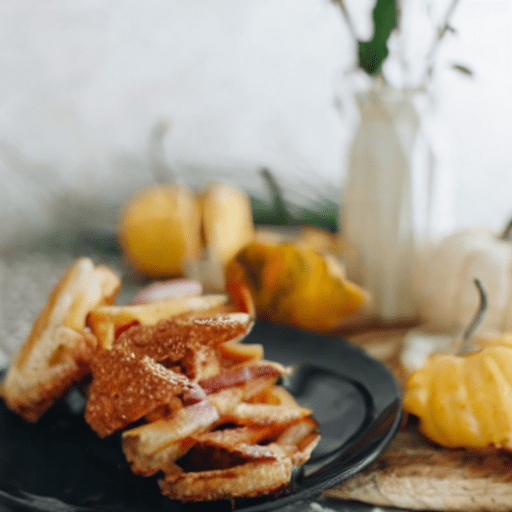} 
\end{minipage}\hfill
\begin{minipage}{0.139\textwidth}
\vspace{-3mm}
\includegraphics[width=\linewidth,height=\linewidth]{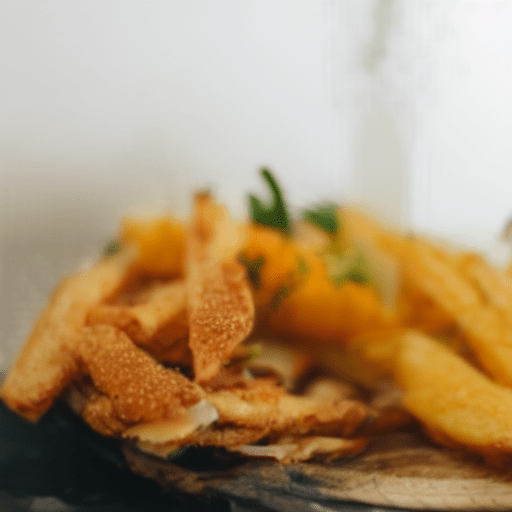} 
\end{minipage}\hfill
\begin{minipage}{0.139\textwidth}
\vspace{-3mm}
\includegraphics[width=\linewidth,height=\linewidth]{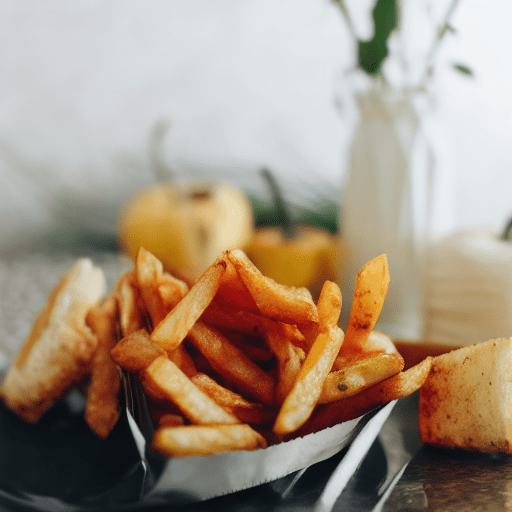} 
\end{minipage}
\vspace{1mm}
\\
\hspace{-2mm} \small{\textcolor{black}{\textsl{Editing Text:}}}   \hspace{50mm} \textcolor{black}{\textit{a generous portion of golden, crispy French fries}}
\vspace{4mm}
\\
\begin{minipage}{0.139\linewidth}
\vspace{-3mm}
\includegraphics[width=\linewidth,height=\linewidth]{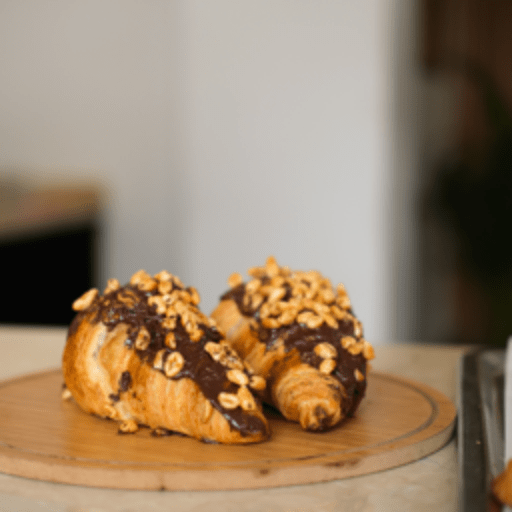}
\end{minipage}\hfill
\begin{minipage}{0.139\linewidth}
\vspace{-3mm}
\includegraphics[width=\linewidth,height=\linewidth]{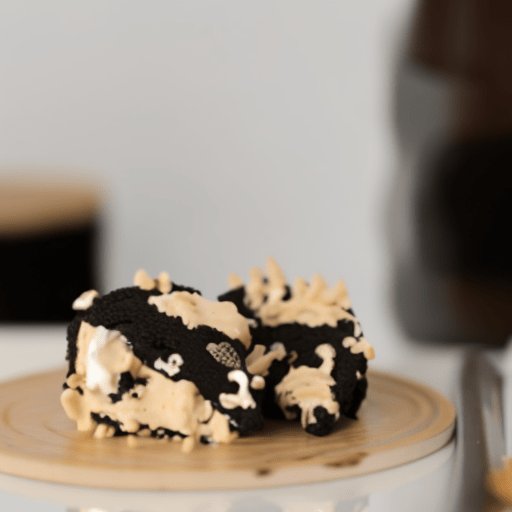}
\end{minipage}\hfill
\begin{minipage}{0.139\textwidth}
\vspace{-3mm}
\includegraphics[width=\linewidth,height=\linewidth]{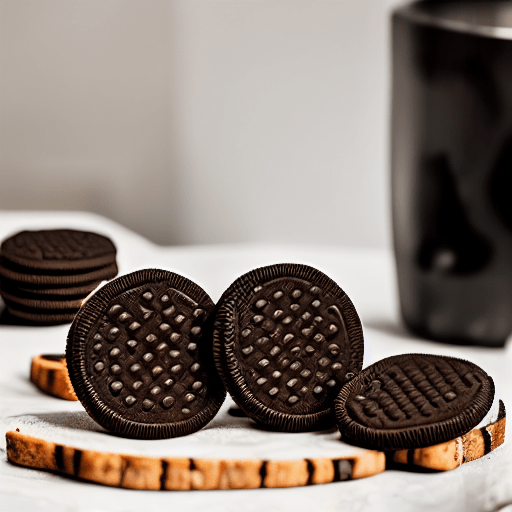}   
\end{minipage}\hfill
\begin{minipage}{0.139\textwidth}
\vspace{-3mm}
\includegraphics[width=\linewidth,height=\linewidth]{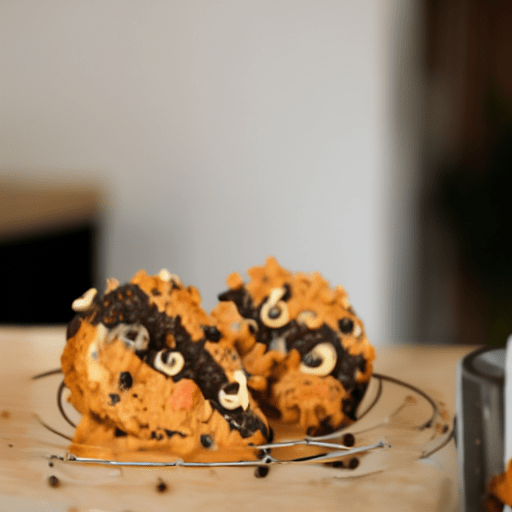} 
\end{minipage}\hfill
\begin{minipage}{0.139\textwidth}
\vspace{-3mm}
\includegraphics[width=\linewidth,height=\linewidth]{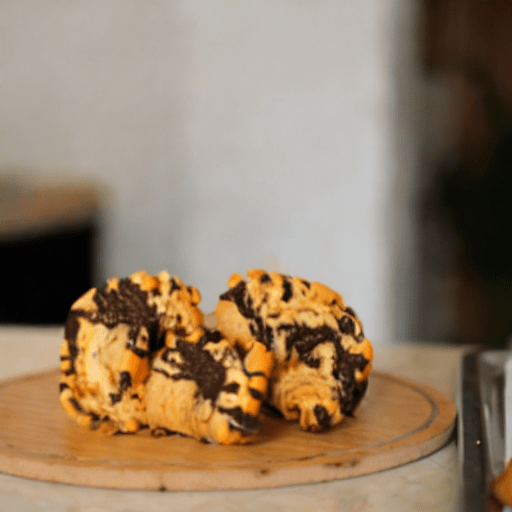} 
\end{minipage}\hfill
\begin{minipage}{0.139\textwidth}
\vspace{-3mm}
\includegraphics[width=\linewidth,height=\linewidth]{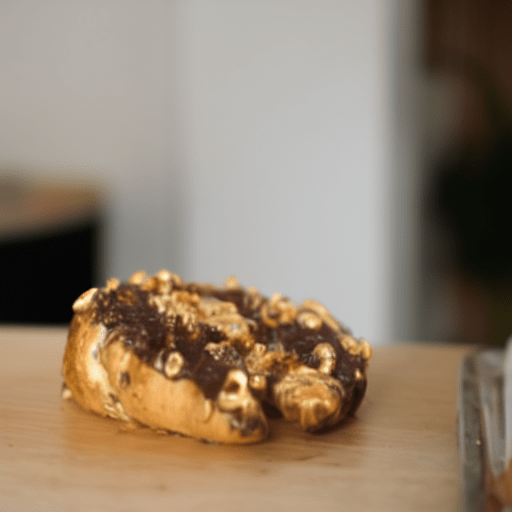} 
\end{minipage}\hfill
\begin{minipage}{0.139\textwidth}
\vspace{-3mm}
\includegraphics[width=\linewidth,height=\linewidth]{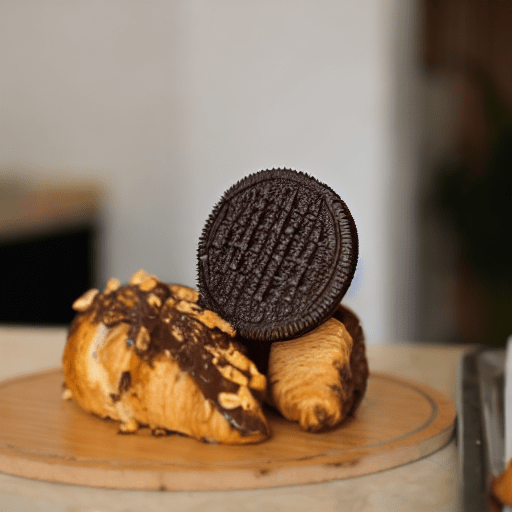} 
\end{minipage}
\vspace{1mm}

\hspace{-1mm} \small{\textcolor{black}{\textsl{Editing Text:}}}   \hspace{56mm} \textcolor{black}{\textit{a piece of Oreo cookie and bread}}
\vspace{4mm}
\\
\begin{minipage}{0.139\linewidth}
\vspace{-3mm}
\includegraphics[width=\linewidth,height=\linewidth]{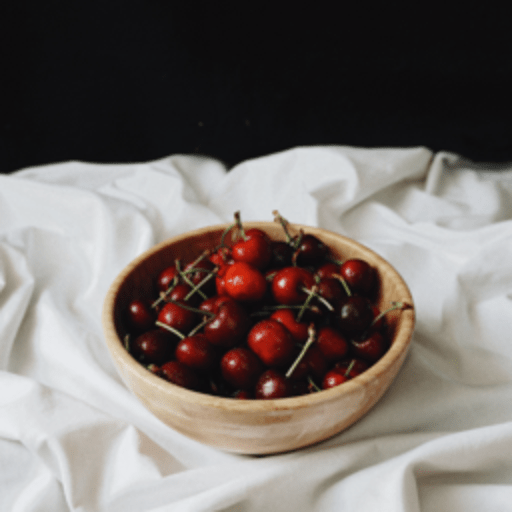}
\end{minipage}\hfill
\begin{minipage}{0.139\linewidth}
\vspace{-3mm}
\includegraphics[width=\linewidth,height=\linewidth]{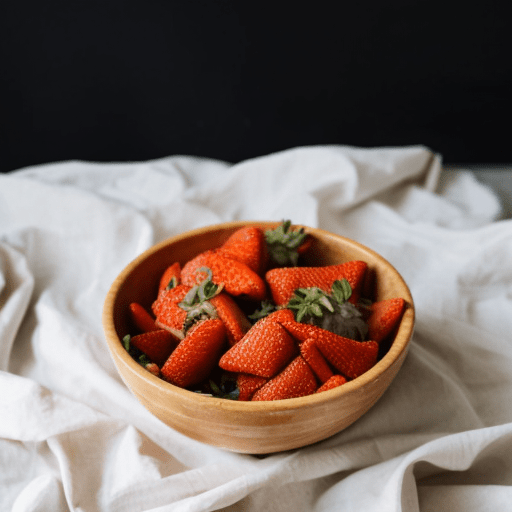}
\end{minipage}\hfill
\begin{minipage}{0.139\textwidth}
\vspace{-3mm}
\includegraphics[width=\linewidth,height=\linewidth]{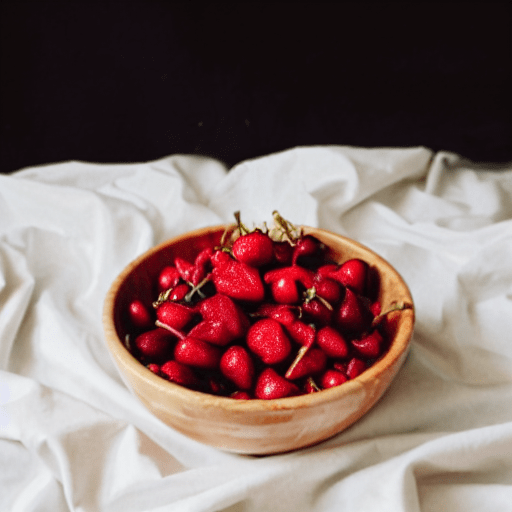}   
\end{minipage}\hfill
\begin{minipage}{0.139\textwidth}
\vspace{-3mm}
\includegraphics[width=\linewidth,height=\linewidth]{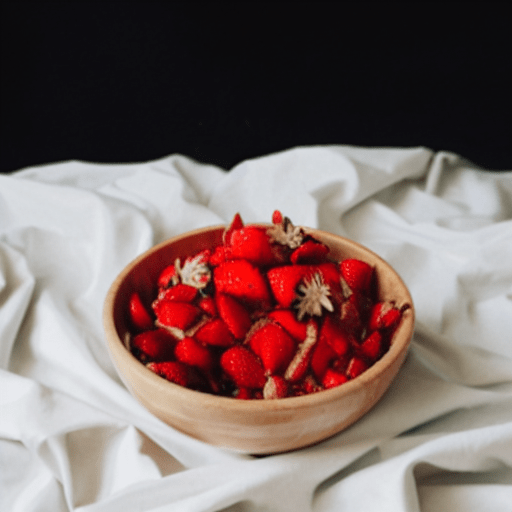} 
\end{minipage}\hfill
\begin{minipage}{0.139\textwidth}
\vspace{-3mm}
\includegraphics[width=\linewidth,height=\linewidth]{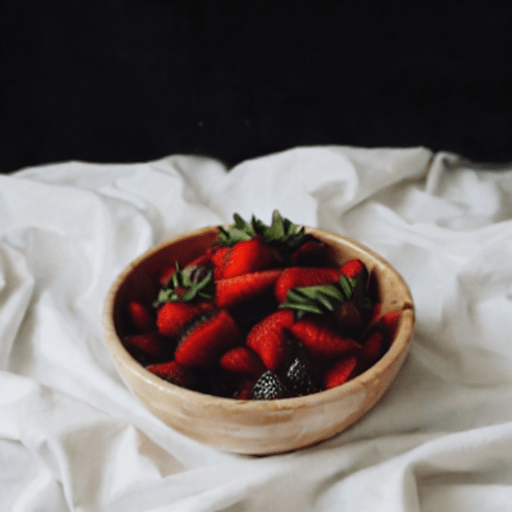} 
\end{minipage}\hfill
\begin{minipage}{0.139\textwidth}
\vspace{-3mm}
\includegraphics[width=\linewidth,height=\linewidth]{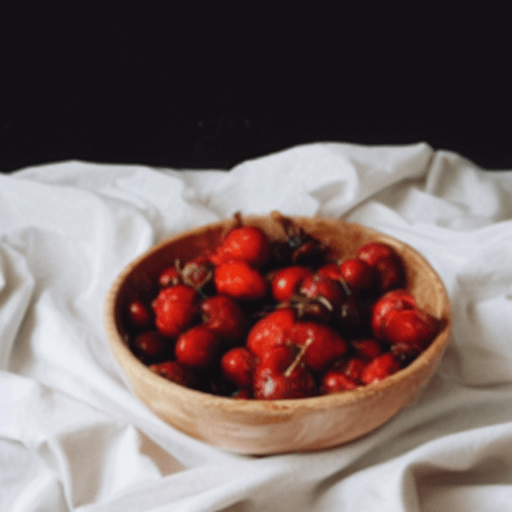} 
\end{minipage}\hfill
\begin{minipage}{0.139\textwidth}
\vspace{-3mm}
\includegraphics[width=\linewidth,height=\linewidth]{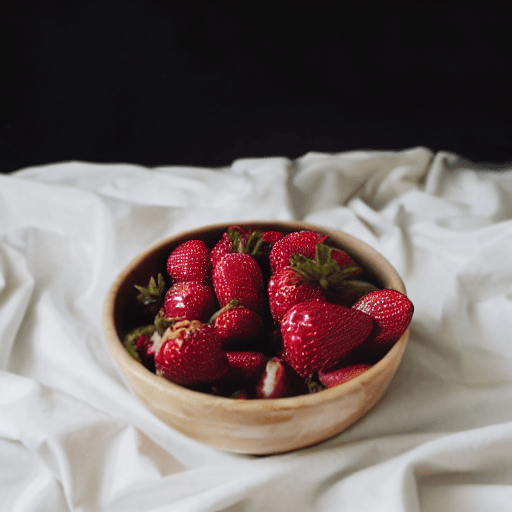} 
\end{minipage}
\vspace{1mm}
\\
\hspace{-2mm} \small{\textcolor{black}{\textsl{Editing Text:}}}   \hspace{65mm} \textcolor{black}{\textit{a bowl of strawberries}}
\vspace{4mm}
\\
\begin{minipage}{0.139\linewidth}
\vspace{-3mm}
\includegraphics[width=\linewidth,height=\linewidth]{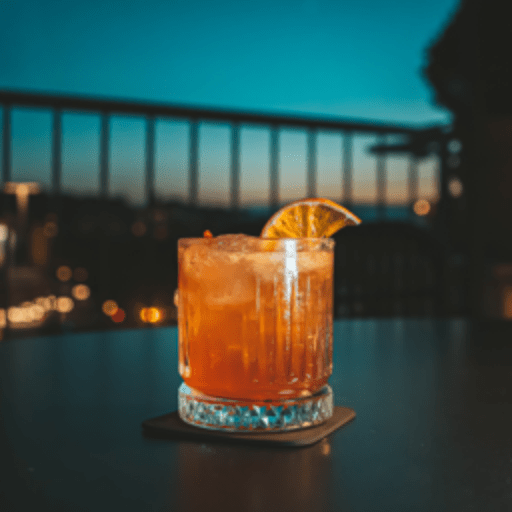}
\end{minipage}\hfill
\begin{minipage}{0.139\linewidth}
\vspace{-3mm}
\includegraphics[width=\linewidth,height=\linewidth]{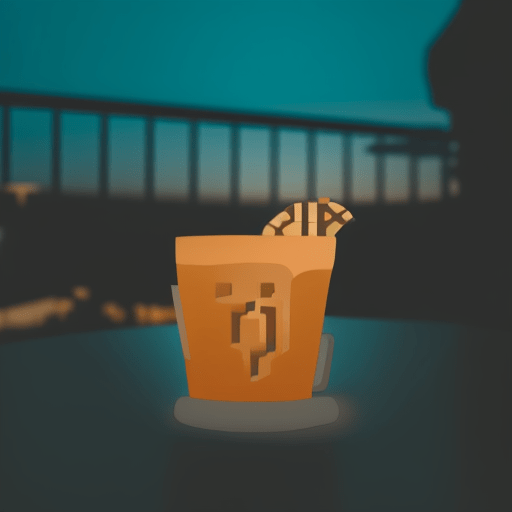}
\end{minipage}\hfill
\begin{minipage}{0.139\textwidth}
\vspace{-3mm}
\includegraphics[width=\linewidth,height=\linewidth]{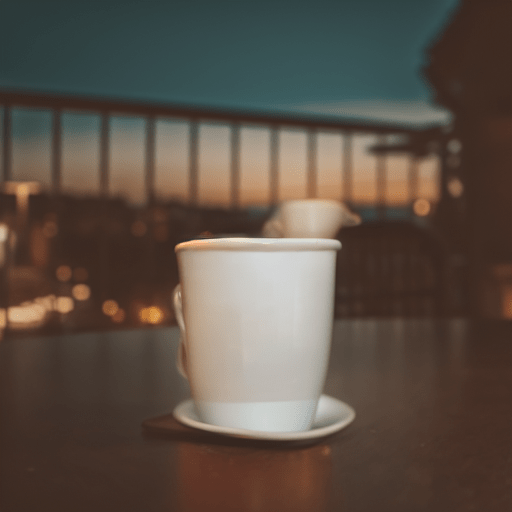}   
\end{minipage}\hfill
\begin{minipage}{0.139\textwidth}
\vspace{-3mm}
\includegraphics[width=\linewidth,height=\linewidth]{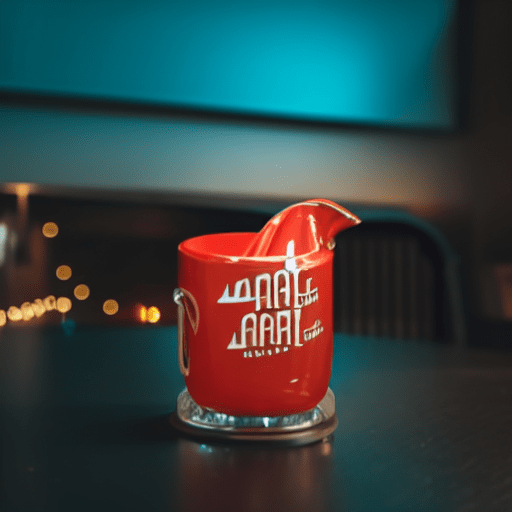} 
\end{minipage}\hfill
\begin{minipage}{0.139\textwidth}
\vspace{-3mm}
\includegraphics[width=\linewidth,height=\linewidth]{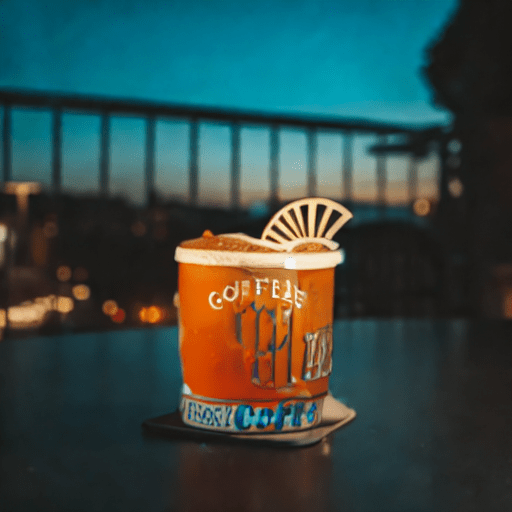} 
\end{minipage}\hfill
\begin{minipage}{0.139\textwidth}
\vspace{-3mm}
\includegraphics[width=\linewidth,height=\linewidth]{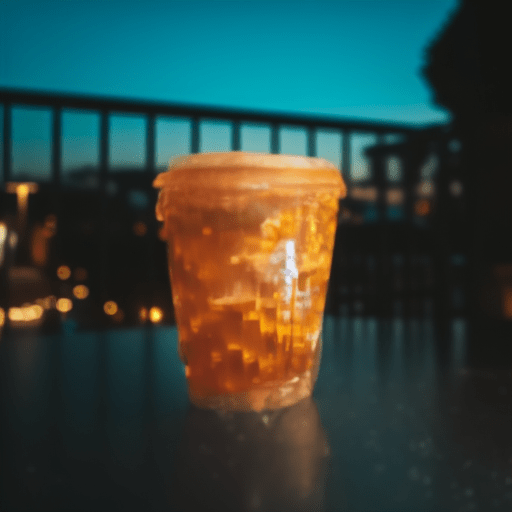} 
\end{minipage}\hfill
\begin{minipage}{0.139\textwidth}
\vspace{-3mm}
\includegraphics[width=\linewidth,height=\linewidth]{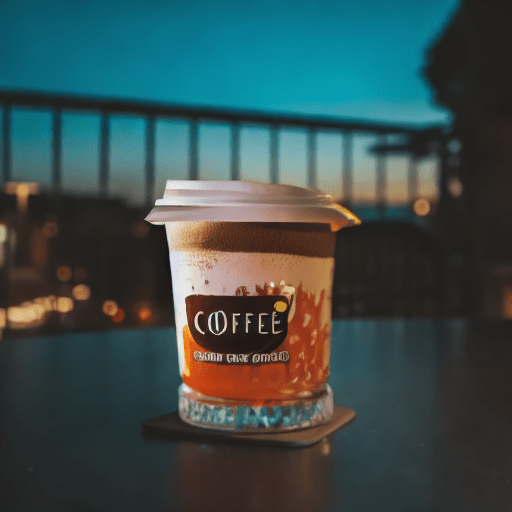} 
\end{minipage}
\vspace{1mm}
\\
\hspace{-2mm} \small{\textsl{Editing Text:}}   \hspace{55mm} \textcolor{black}{\textit{a cup whose logo is named as ``coffee"}}
\vspace{4mm}
\\
\begin{minipage}{0.139\linewidth}
\vspace{-3mm}
\includegraphics[width=\linewidth,height=\linewidth]{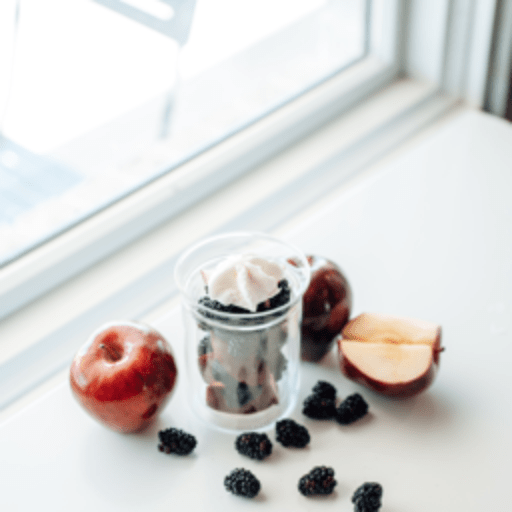}
\end{minipage}\hfill
\begin{minipage}{0.139\linewidth}
\vspace{-3mm}
\includegraphics[width=\linewidth,height=\linewidth]{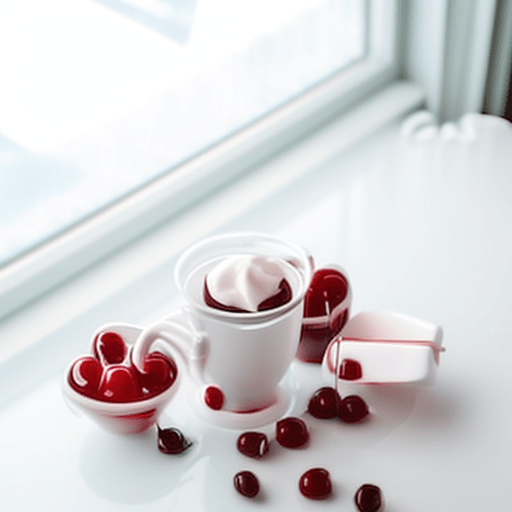}
\end{minipage}\hfill
\begin{minipage}{0.139\textwidth}
\vspace{-3mm}
\includegraphics[width=\linewidth,height=\linewidth]{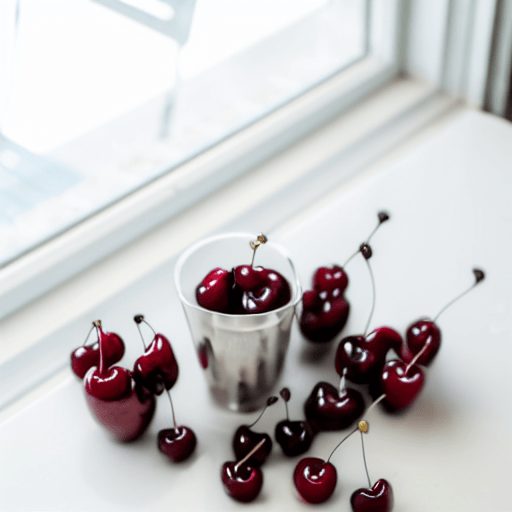}   
\end{minipage}\hfill
\begin{minipage}{0.139\textwidth}
\vspace{-3mm}
\includegraphics[width=\linewidth,height=\linewidth]{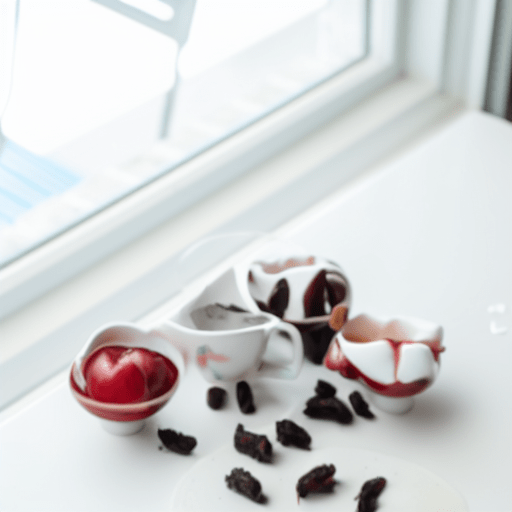} 
\end{minipage}\hfill
\begin{minipage}{0.139\textwidth}
\vspace{-3mm}
\includegraphics[width=\linewidth,height=\linewidth]{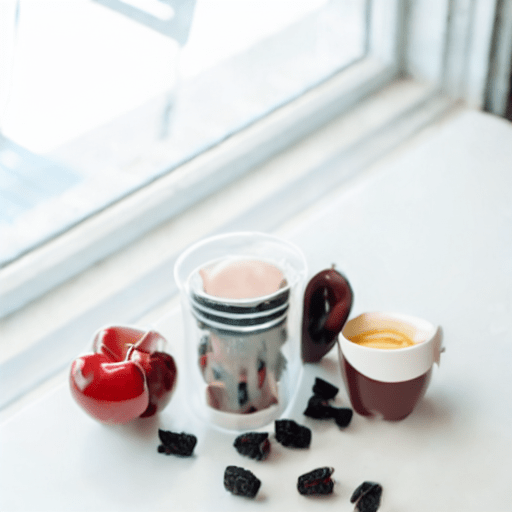} 
\end{minipage}\hfill
\begin{minipage}{0.139\textwidth}
\vspace{-3mm}
\includegraphics[width=\linewidth,height=\linewidth]{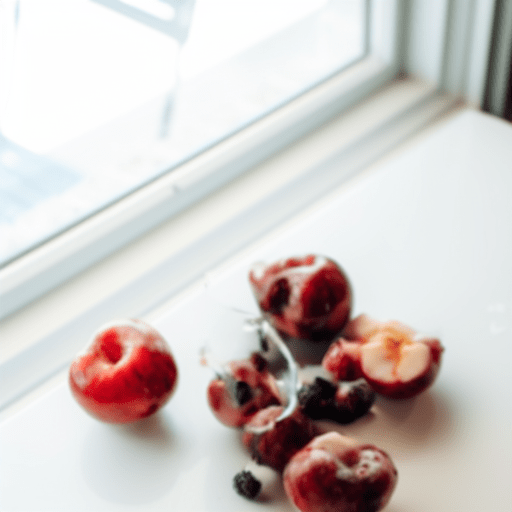} 
\end{minipage}\hfill
\begin{minipage}{0.139\textwidth}
\vspace{-3mm}
\includegraphics[width=\linewidth,height=\linewidth]{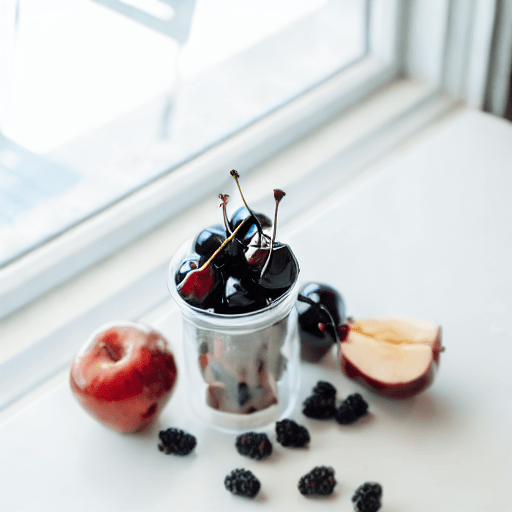} 
\end{minipage}
\vspace{1mm}
\\
\hspace{-2mm} \small{\textcolor{black}{\textsl{Editing Text:}}}   \hspace{55mm} \textcolor{black}{\textit{there are several cherries in the cup}}
\vspace{-1mm}
\captionof{figure}{\textbf{Comparison with existing methods.} We compare our method with existing text-driven image editing methods. From left to right: Input image, Plug-and-Play \cite{tumanyan2023plug}, InstructPix2Pix \cite{brooks2023instructpix2pix}, Null-text \cite{mokady2023null}, DiffEdit \cite{couairon2022diffedit}, MasaCtrl \cite{cao2023masactrl}, and ours.
\vspace{1.5mm}
}
\label{figure_comp2}
\end{figure*}

\clearpage

\begin{figure*}
\begin{minipage}{0.24\linewidth}
\captionof*{figure}{\small{\textcolor{black}{Input Image}}}
\vspace{-3mm}
\includegraphics[width=\linewidth]{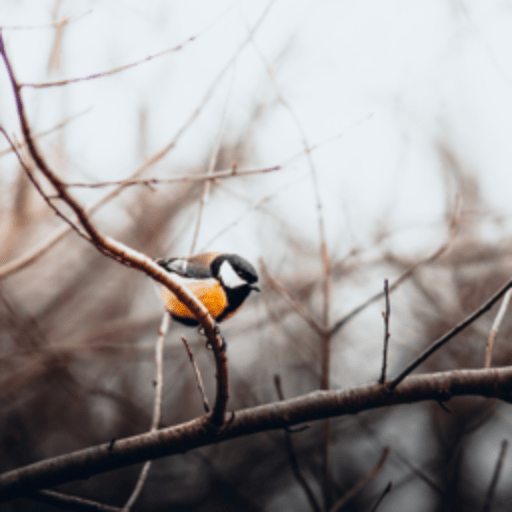}
\end{minipage}\hfill
\begin{minipage}{0.24\textwidth}
\captionof*{figure}{\small{\textcolor{black}{Edited Image}}} 
\vspace{-3mm}
\includegraphics[width=\linewidth]{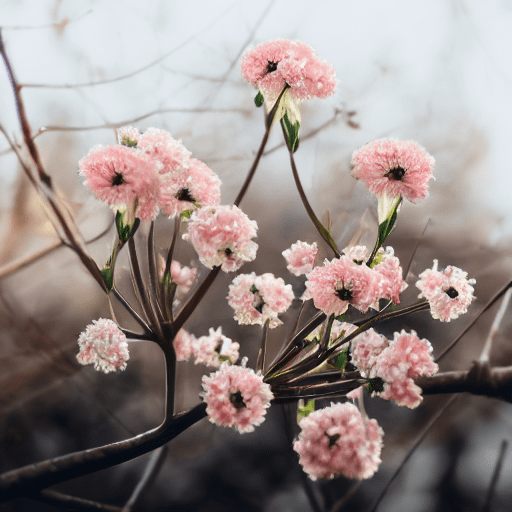}   
\end{minipage}\hfill
\begin{minipage}{0.24\textwidth}
\captionof*{figure}{\small{\textcolor{black}{Input Image}}}
\vspace{-3mm}
\includegraphics[width=\linewidth]{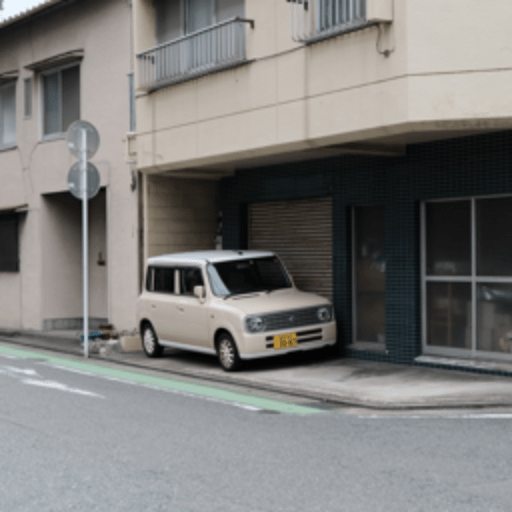} 
\end{minipage}\hfill
\begin{minipage}{0.24\textwidth}
\captionof*{figure}{\small{\textcolor{black}{Edited Image}}}
\vspace{-3mm}
\includegraphics[width=\linewidth]{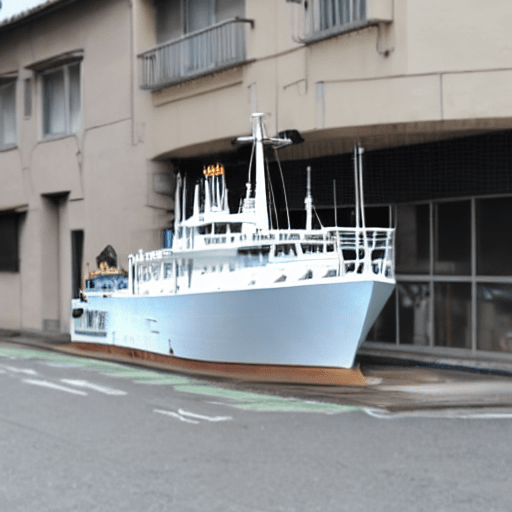} 
\end{minipage}
\vspace{2mm}
\\
\vspace{1mm}
\hspace{10mm} \textit{a multitude of pink flowers bursts into bloom}  \hspace{20mm}  \textit{a securely moored white ship on the solid ground}
\vspace{1mm}
\\
\begin{minipage}{0.24\textwidth}
\includegraphics[width=\linewidth]{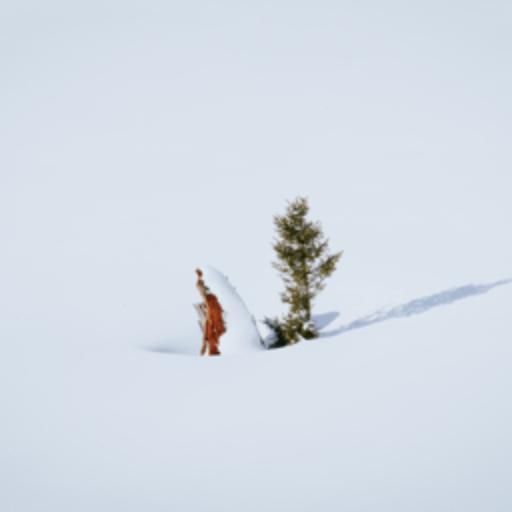}
\end{minipage}\hfill
\begin{minipage}{0.24\textwidth}
\includegraphics[width=\linewidth]{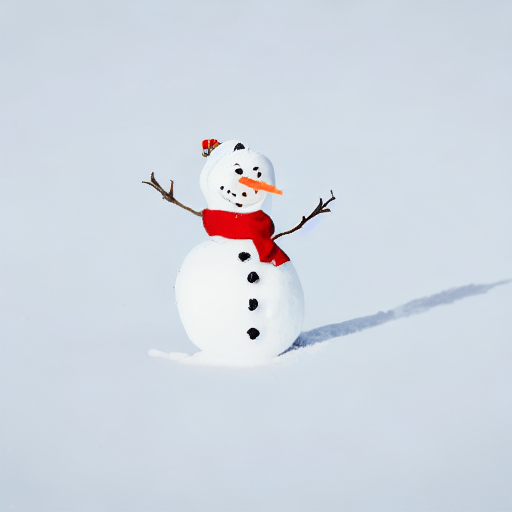}   
\end{minipage}\hfill
\begin{minipage}{0.24\textwidth}
\includegraphics[width=\linewidth]{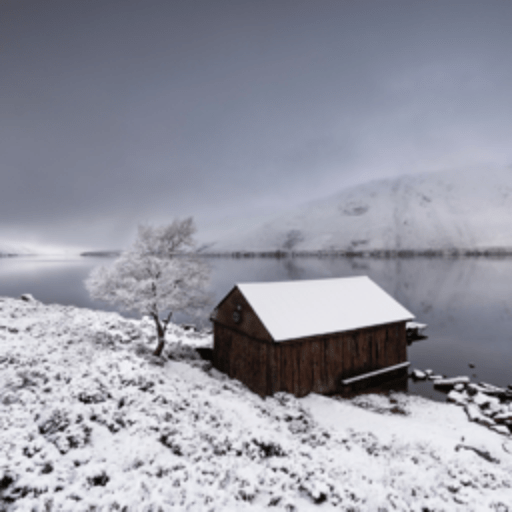} 
\end{minipage}\hfill
\begin{minipage}{0.24\textwidth}
\includegraphics[width=\linewidth]{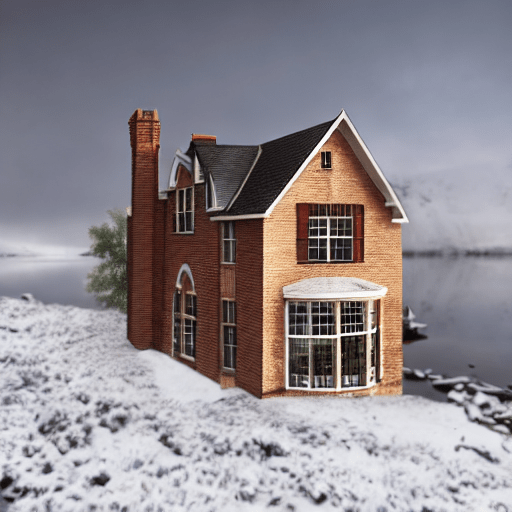} 
\end{minipage}
\vspace{2mm}
\\
\vspace{1mm}
\hspace{8mm} \textit{a cheerful snowman adorned with a carrot nose}  \hspace{16mm}  \textit{a classic brick dwelling with architectural solidity}
\vspace{1mm}
\\
\begin{minipage}{0.24\textwidth}
\includegraphics[width=\linewidth]{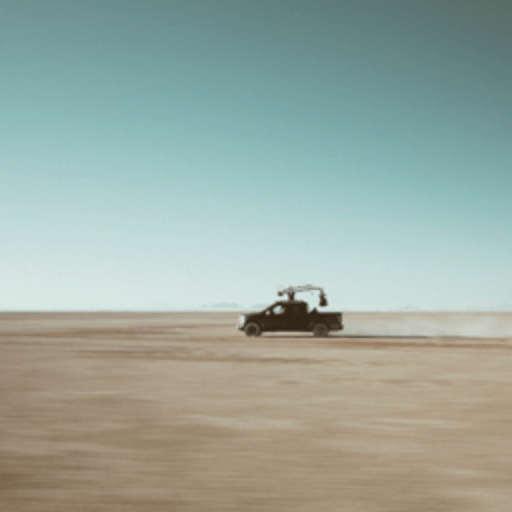}
\end{minipage}\hfill
\begin{minipage}{0.24\textwidth}
\includegraphics[width=\linewidth]{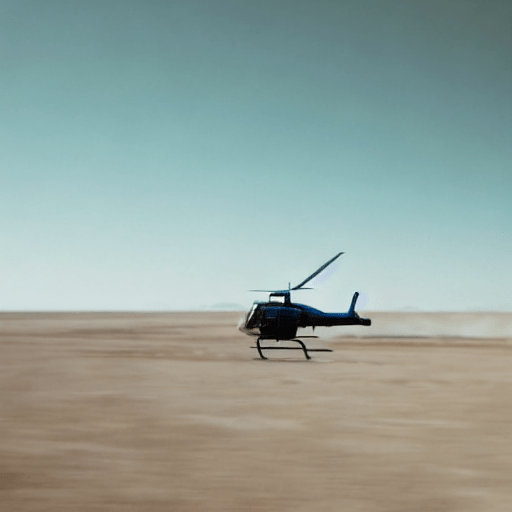}   
\end{minipage}\hfill
\begin{minipage}{0.24\textwidth}
\includegraphics[width=\linewidth]{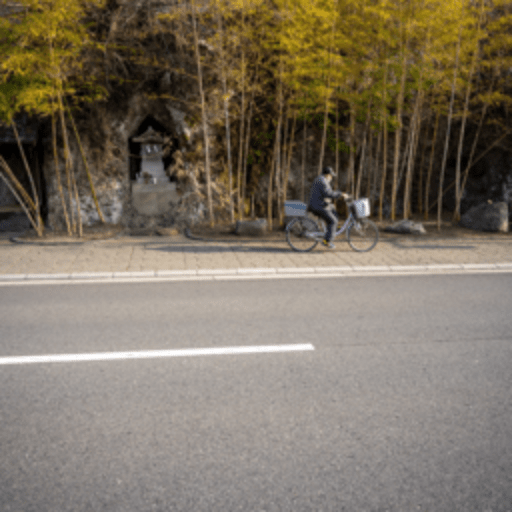} 
\end{minipage}\hfill
\begin{minipage}{0.24\textwidth}
\includegraphics[width=\linewidth]{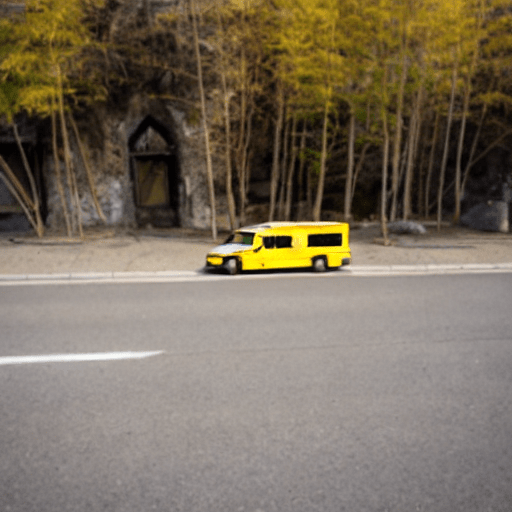} 
\end{minipage}
\vspace{2mm}
\\
\vspace{1mm}
\hspace{18mm} \textit{the helicopter readies for takeoff}  \hspace{32mm}  \textit{a yellow school bus toy rolls down the road}
\vspace{1mm}
\\
\begin{minipage}{0.24\textwidth}
\includegraphics[width=\linewidth]{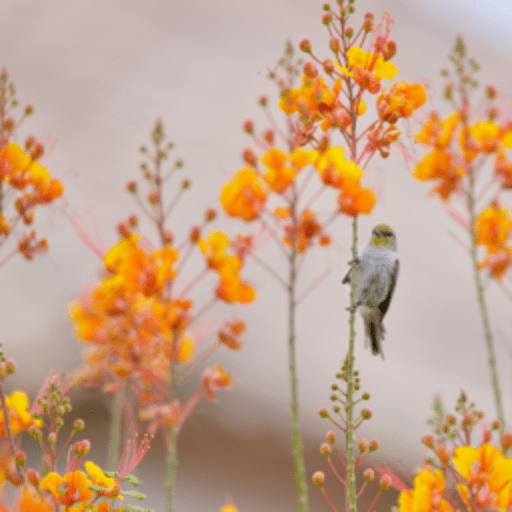}
\end{minipage}\hfill
\begin{minipage}{0.24\textwidth}
\includegraphics[width=\linewidth]{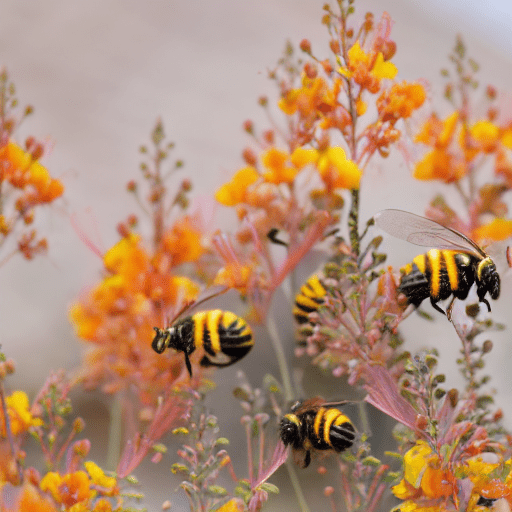}   
\end{minipage}\hfill
\begin{minipage}{0.24\textwidth}
\includegraphics[width=\linewidth]{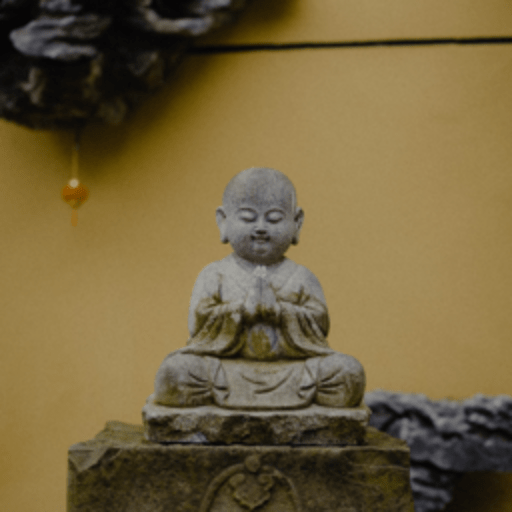} 
\end{minipage}\hfill
\begin{minipage}{0.24\textwidth}
\includegraphics[width=\linewidth]{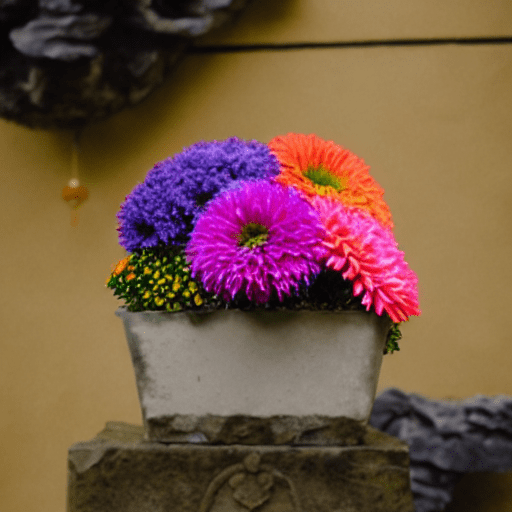} 
\end{minipage}
\vspace{2mm}
\\
\vspace{1mm}
\hspace{12mm} \textit{some small bees flit from flower to flower}  \hspace{17mm}  \textit{a pot of chrysanthemums bursts forth with vibrant colors}
\captionof{figure}{\textbf{Text-driven image editing results.} Given an input image and a language description, our method can generate realistic and relevant images without the need for user-specified regions for editing. It performs local image editing while preserving the image context.}
\label{figure3}
\end{figure*}

\begin{figure*}
\begin{minipage}{0.24\linewidth}
\captionof*{figure}{\small{\textcolor{black}{Input Image}}}
\vspace{-3mm}
\includegraphics[width=\linewidth]{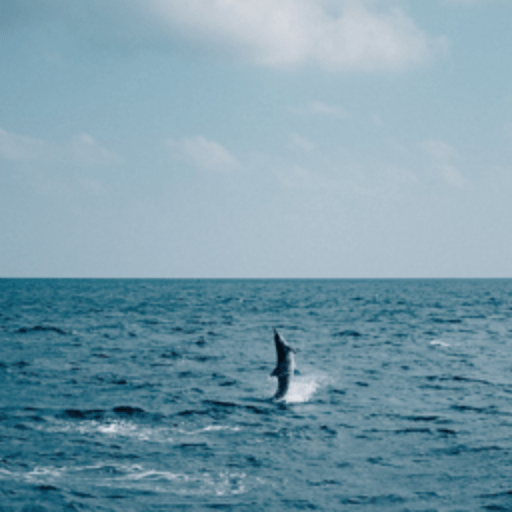}
\end{minipage}\hfill
\begin{minipage}{0.24\textwidth}
\captionof*{figure}{\small{\textcolor{black}{Edited Image}}}
\vspace{-3mm}
\includegraphics[width=\linewidth]{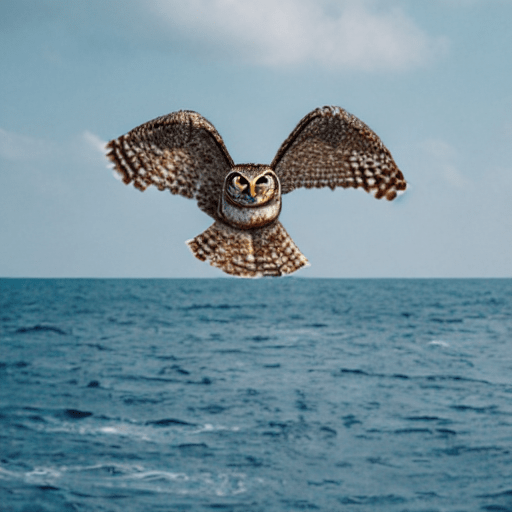}   
\end{minipage}\hfill
\begin{minipage}{0.24\textwidth}
\captionof*{figure}{\small{\textcolor{black}{Input Image}}}
\vspace{-3mm}
\includegraphics[width=\linewidth]{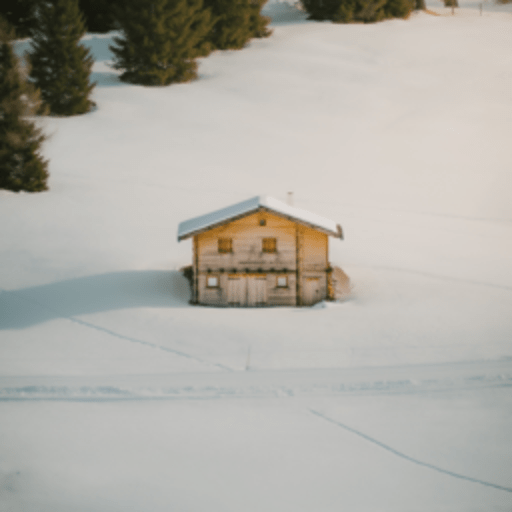} 
\end{minipage}\hfill
\begin{minipage}{0.24\textwidth}
\captionof*{figure}{\small{\textcolor{black}{Edited Image}}}
\vspace{-3mm}
\includegraphics[width=\linewidth]{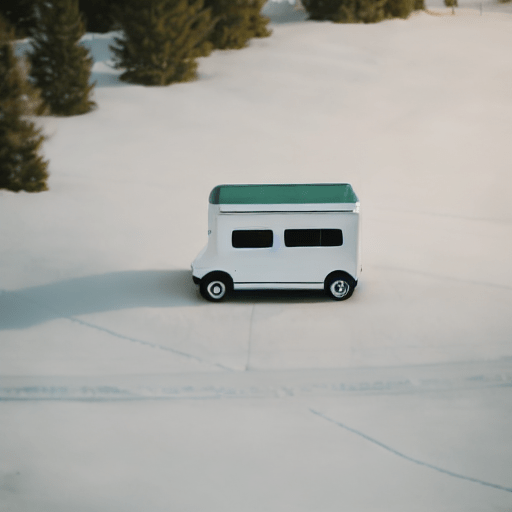} 
\end{minipage}
\vspace{2mm}
\\
\vspace{1mm}
\hspace{14mm} \textit{an owl gracefully hovers above the sea}  \hspace{30mm}  \textit{a parked caravan epitomizes travel tales}
\vspace{1mm}
\\
\begin{minipage}{0.24\textwidth}
\includegraphics[width=\linewidth]{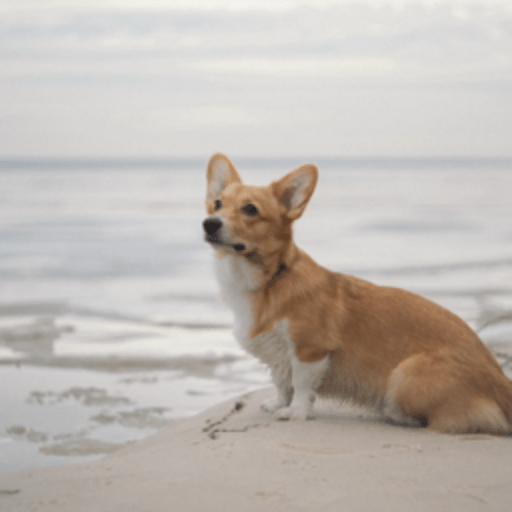}
\end{minipage}\hfill
\begin{minipage}{0.24\textwidth}
\includegraphics[width=\linewidth]{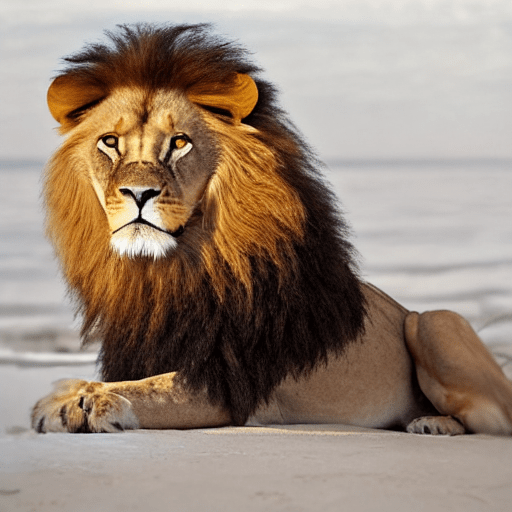}   
\end{minipage}\hfill
\begin{minipage}{0.24\textwidth}
\includegraphics[width=\linewidth]{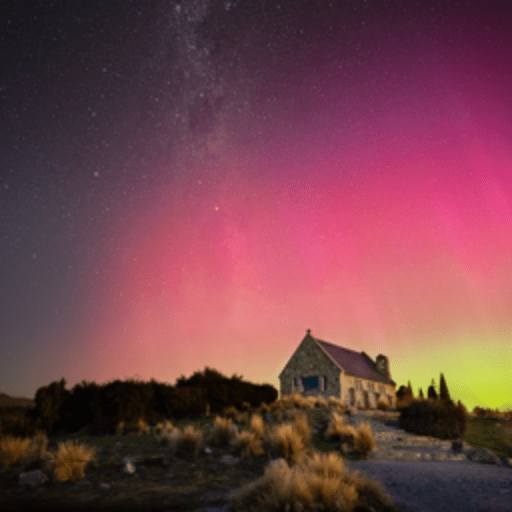} 
\end{minipage}\hfill
\begin{minipage}{0.24\textwidth}
\includegraphics[width=\linewidth]{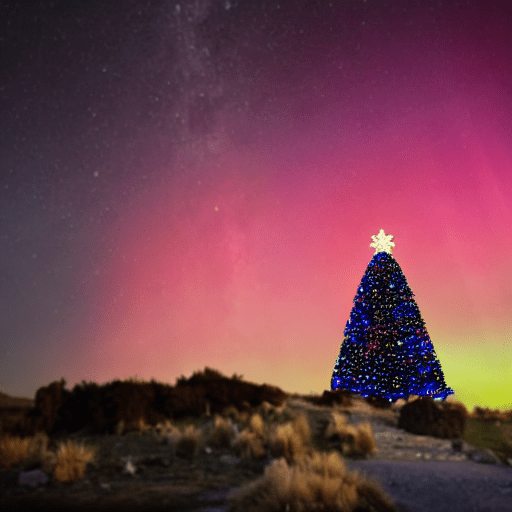} 
\end{minipage}
\vspace{2mm}
\\
\vspace{1mm}
\hspace{12mm} \textit{a powerful lion with a magnificent mane}  \hspace{26mm}  \textit{a Christmas tree adorned with twinkling lights}
\vspace{1mm}
\\
\begin{minipage}{0.24\textwidth}
\includegraphics[width=\linewidth]{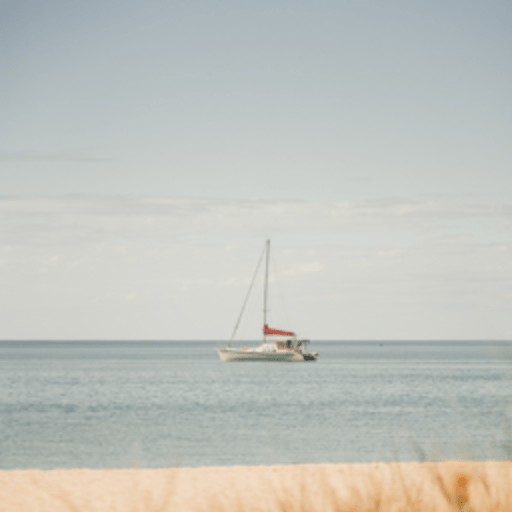}
\end{minipage}\hfill
\begin{minipage}{0.24\textwidth} 
\includegraphics[width=\linewidth]{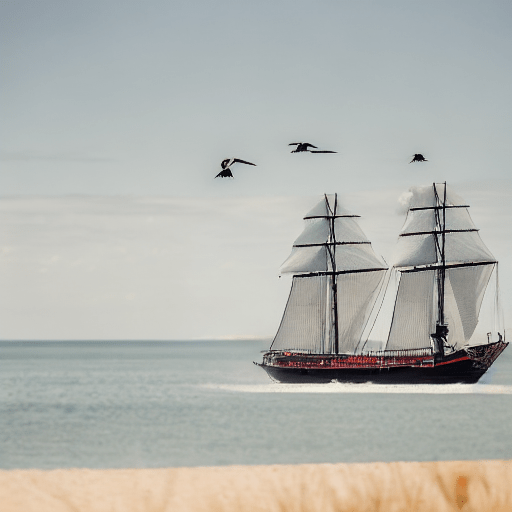}   
\end{minipage}\hfill
\begin{minipage}{0.24\textwidth}
\includegraphics[width=\linewidth]{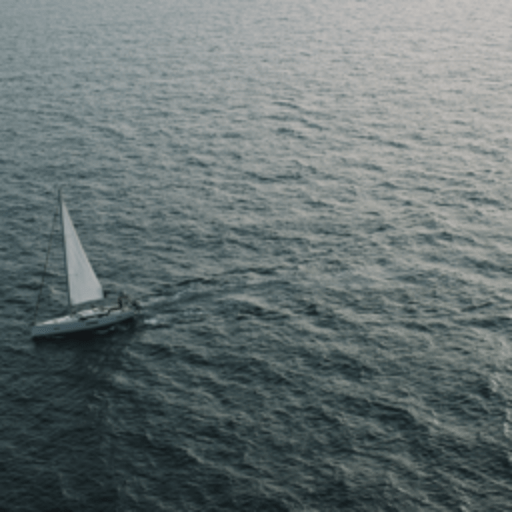} 
\end{minipage}\hfill
\begin{minipage}{0.24\textwidth}
\includegraphics[width=\linewidth]{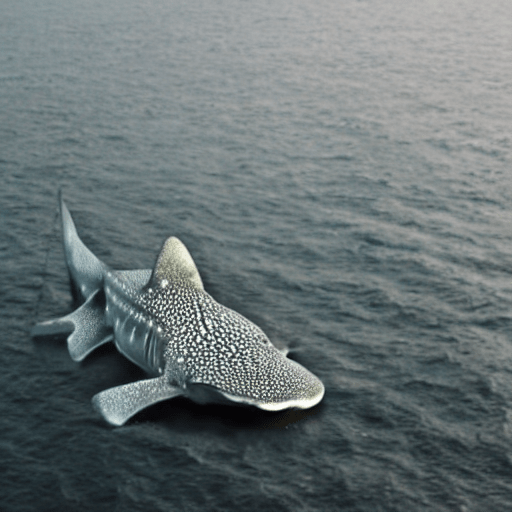} 
\end{minipage}
\vspace{2mm}
\\
\vspace{1mm}
\hspace{8mm} \textit{a large steamer sails accompanied by sea gulls}  \hspace{11mm}  \textit{a colossal whale shark floating on the top of the deep ocean}
\vspace{1mm}
\\
\begin{minipage}{0.24\textwidth}
\includegraphics[width=\linewidth]{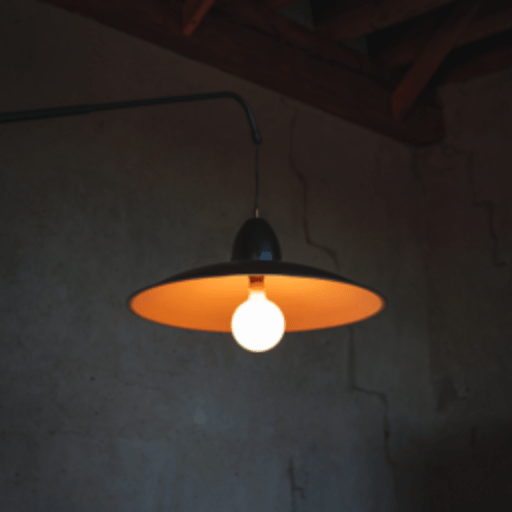}
\end{minipage}\hfill
\begin{minipage}{0.24\textwidth}
\includegraphics[width=\linewidth]{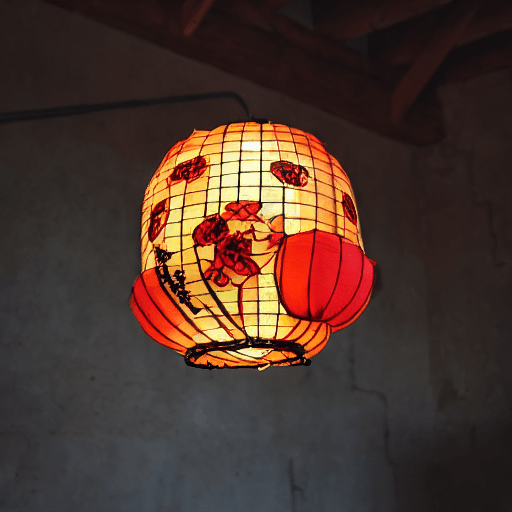}   
\end{minipage}\hfill
\begin{minipage}{0.24\textwidth}
\includegraphics[width=\linewidth]{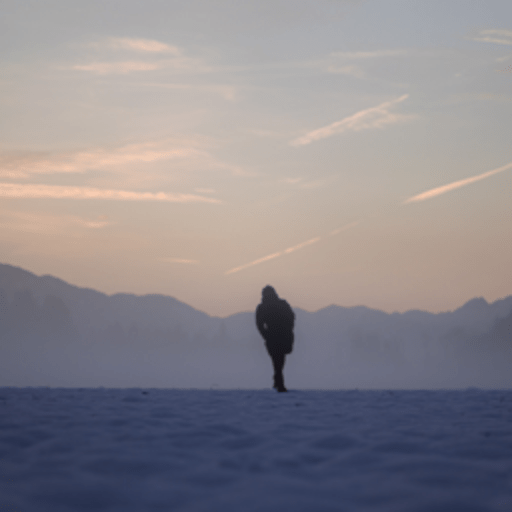} 
\end{minipage}\hfill
\begin{minipage}{0.24\textwidth}
\includegraphics[width=\linewidth]{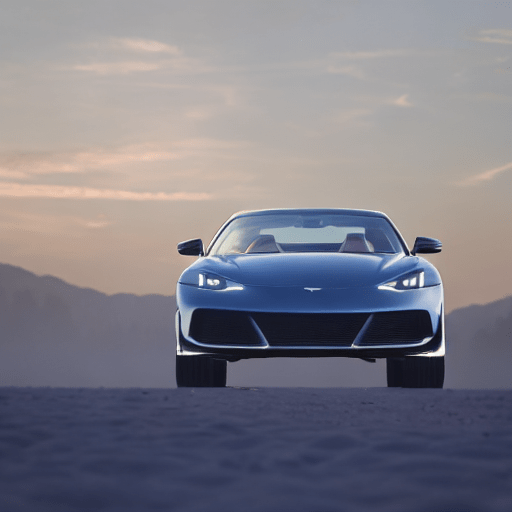} 
\end{minipage}
\vspace{2mm}
\\
\vspace{1mm}
 \textit{the Chinese lantern adds cultural elegance, radiating a soft glow}  \hspace{6mm}  \textit{a sleek sports car exudes an aura of speed}

\captionof{figure}{\textbf{Text-driven image editing results.} Given an input image and a language description, our method can generate realistic and relevant images without the need for user-specified regions for editing. It performs local image editing while preserving the image context.}
\label{figure4}
\end{figure*}

\begin{figure*}
\begin{minipage}{0.24\linewidth}
\captionof*{figure}{\small{\textcolor{black}{Input Image}}}
\vspace{-3mm}
\includegraphics[width=\linewidth]{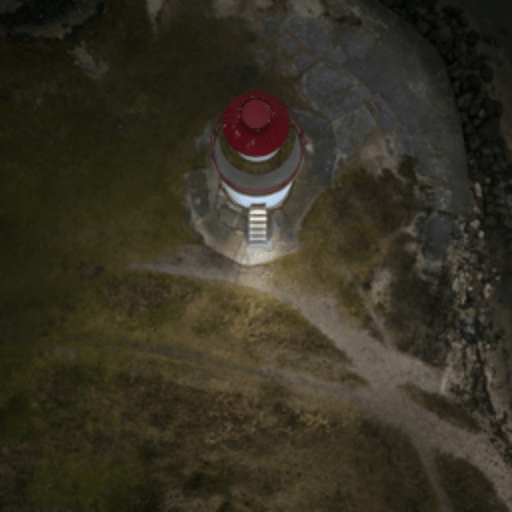}
\end{minipage}\hfill
\begin{minipage}{0.24\textwidth}
\captionof*{figure}{\small{\textcolor{black}{Edited Image}}}
\vspace{-3mm}
\includegraphics[width=\linewidth]{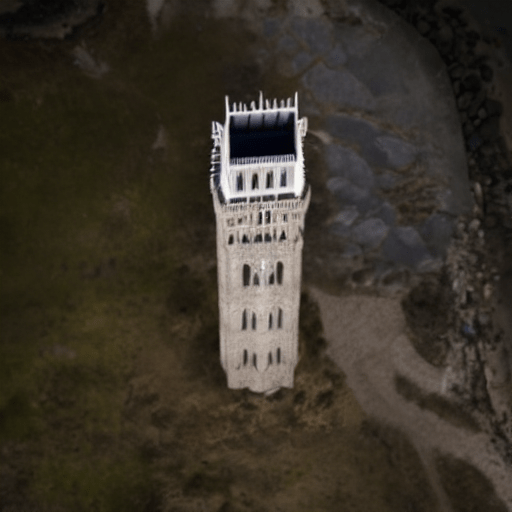}   
\end{minipage}\hfill
\begin{minipage}{0.24\textwidth}
\captionof*{figure}{\small{\textcolor{black}{Input Image}}}
\vspace{-3mm}
\includegraphics[width=\linewidth]{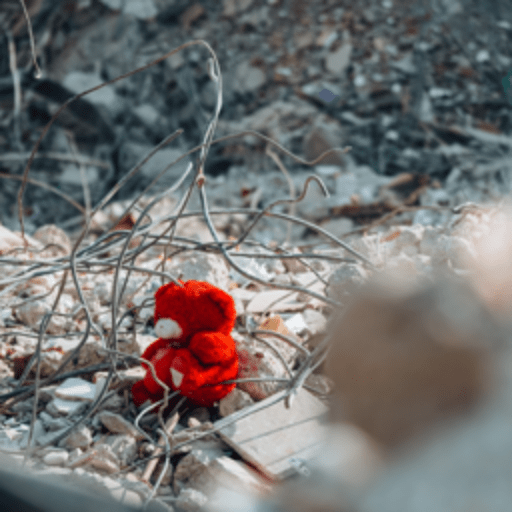} 
\end{minipage}\hfill
\begin{minipage}{0.24\textwidth}
\captionof*{figure}{\small{\textcolor{black}{Edited Image}}}
\vspace{-3mm}
\includegraphics[width=\linewidth]{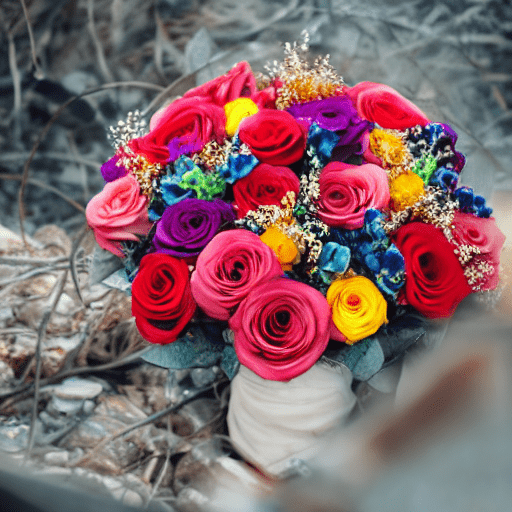} 
\end{minipage}
\vspace{2mm}
\\
\vspace{1mm}
\hspace{3mm} \textit{the tower commands attention with its majestic height}  \hspace{17mm}  \textit{a bouquet of vibrant roses exuding elegance}
\vspace{1mm}
\\
\begin{minipage}{0.24\textwidth}
\includegraphics[width=\linewidth]{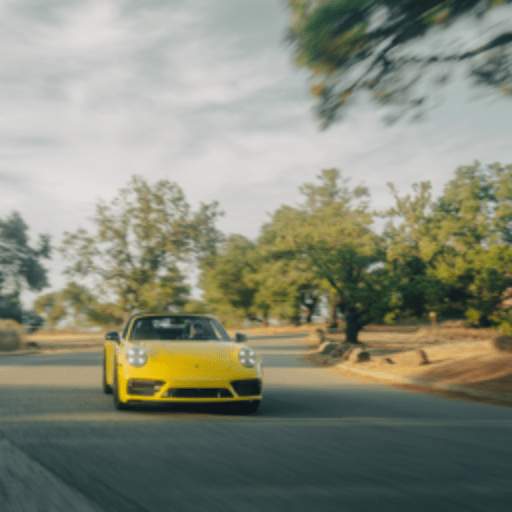}
\end{minipage}\hfill
\begin{minipage}{0.24\textwidth}
\includegraphics[width=\linewidth]{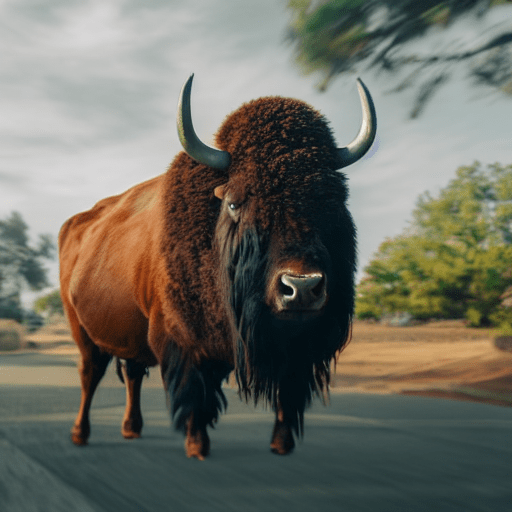}   
\end{minipage}\hfill
\begin{minipage}{0.24\textwidth}
\includegraphics[width=\linewidth]{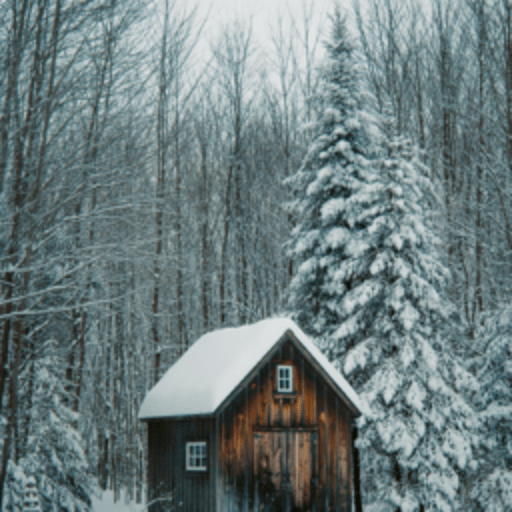} 
\end{minipage}\hfill
\begin{minipage}{0.24\textwidth}
\includegraphics[width=\linewidth]{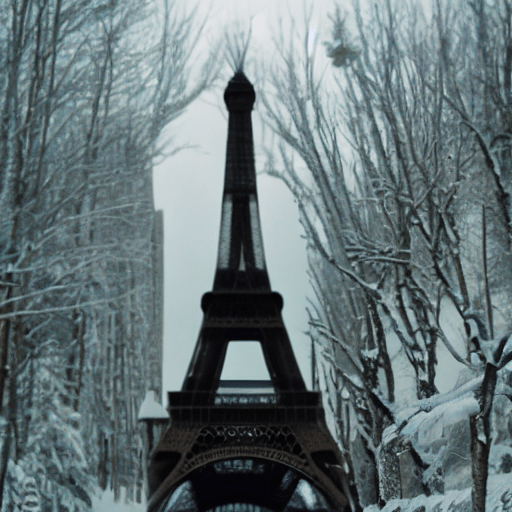} 
\end{minipage}
\vspace{2mm}
\\
\vspace{1mm}
\hspace{3mm} \textit{a sturdy brown bison with a thick fur coat}  \hspace{2mm}  \textit{the iconic Eiffel Tower adorned with a picturesque layer of glistening snow}
\vspace{1mm}
\\
\begin{minipage}{0.24\textwidth}
\includegraphics[width=\linewidth]{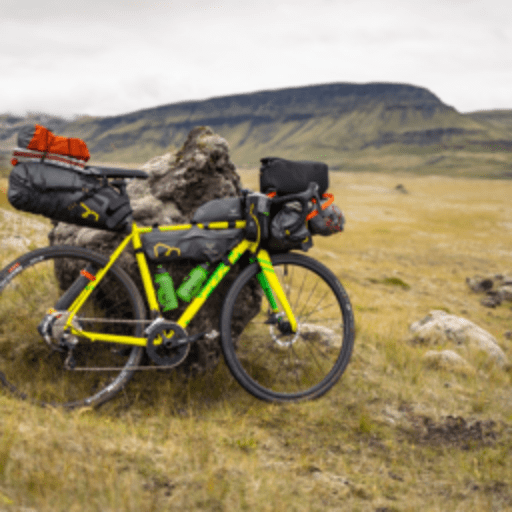}
\end{minipage}\hfill
\begin{minipage}{0.24\textwidth}
\includegraphics[width=\linewidth]{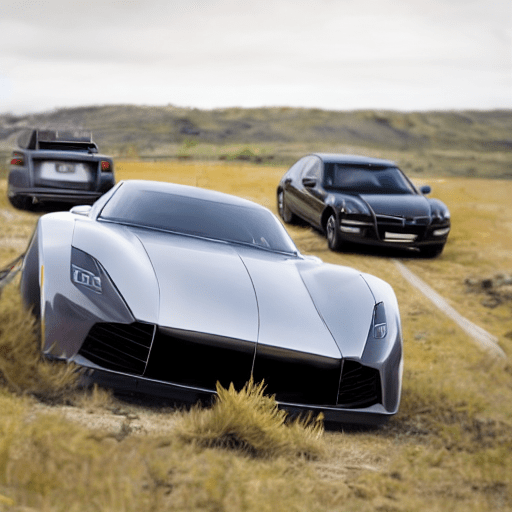}   
\end{minipage}\hfill
\begin{minipage}{0.24\textwidth}
\includegraphics[width=\linewidth]{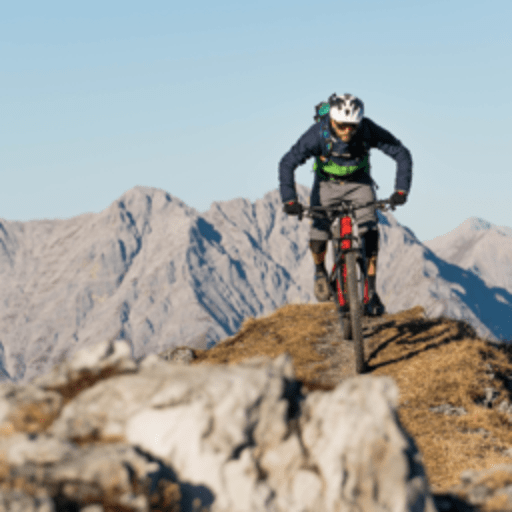} 
\end{minipage}\hfill
\begin{minipage}{0.24\textwidth}
\includegraphics[width=\linewidth]{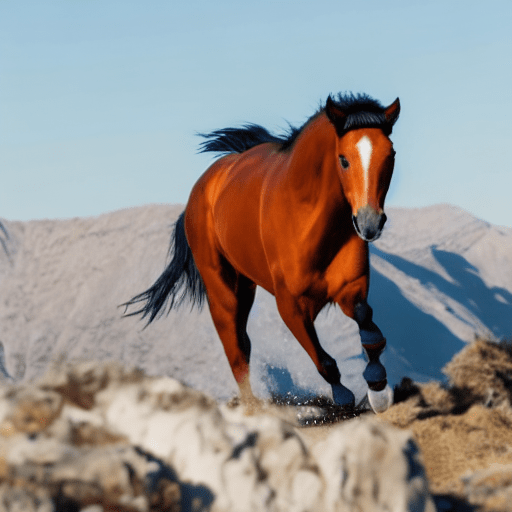} 
\end{minipage}
\vspace{2mm}
\\
\vspace{1mm}
\hspace{6mm} \textit{beautiful cars with sleek lines and polished exterior}  \hspace{11mm}  \textit{a spirited brown horse charges through the terrains}
\vspace{1mm}
\\
\begin{minipage}{0.24\textwidth}
\includegraphics[width=\linewidth]{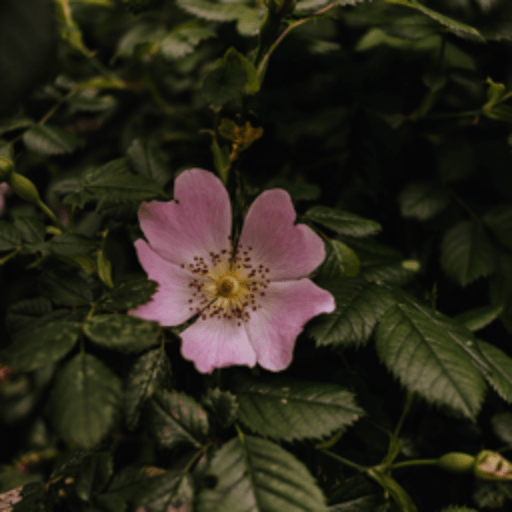}
\end{minipage}\hfill
\begin{minipage}{0.24\textwidth} 
\includegraphics[width=\linewidth]{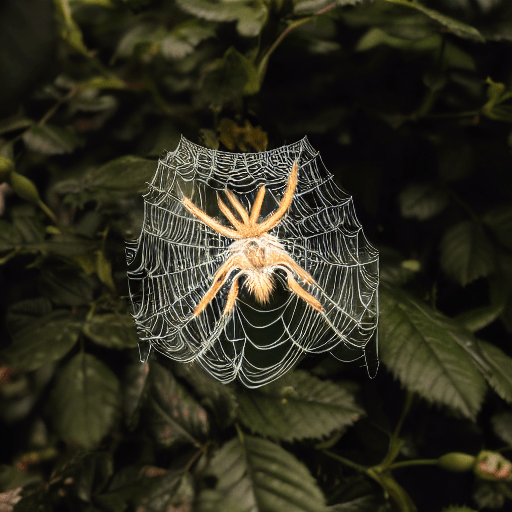}   
\end{minipage}\hfill
\begin{minipage}{0.24\textwidth}
\includegraphics[width=\linewidth]{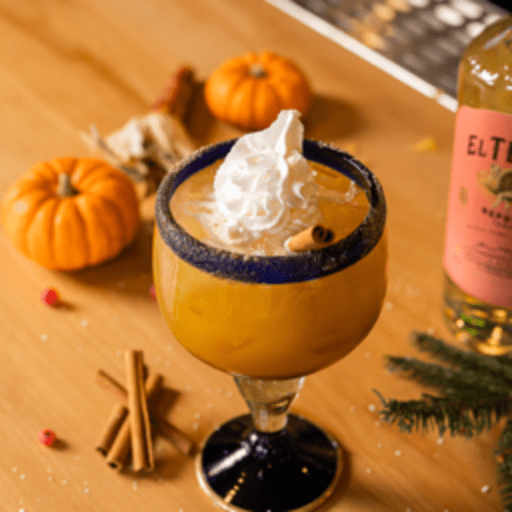} 
\end{minipage}\hfill
\begin{minipage}{0.24\textwidth}
\includegraphics[width=\linewidth]{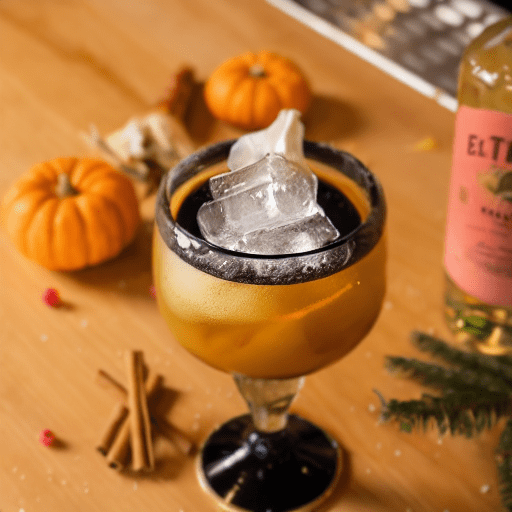} 
\end{minipage}
\vspace{2mm}
\\
\vspace{1mm}
 \textit{a delicate spider weaves an intricate web} \hspace{3mm}  \textit{a delightful cup holds the harmonious blend of ice wine and velvety ice cream}

\captionof{figure}{\textbf{Text-driven image editing results.} Given an input image and a language description, our method can generate realistic and relevant images without the need for user-specified regions for editing. It performs local image editing while preserving the image context.}
\label{figure5_2}
\end{figure*}

\end{document}